\documentclass{article} %
\newif\ificlrpreprintbuild
\iclrpreprintbuildtrue

\usepackage{iclr2027_conference,times}

\usepackage{amsmath,amsfonts,bm}

\def\eqref#1{equation~\ref{#1}}

\def\1{\bm{1}}

\DeclareMathAlphabet{\mathsfit}{\encodingdefault}{\sfdefault}{m}{sl}
\SetMathAlphabet{\mathsfit}{bold}{\encodingdefault}{\sfdefault}{bx}{n}

\usepackage{amsthm}
\usepackage{amssymb}
\ificlrpreprintbuild\else
  \usepackage{hyperref}
\fi
\usepackage{url}
\usepackage{algorithm}
\usepackage{algpseudocode}
\usepackage{graphicx}
\graphicspath{{images/}}
\usepackage{booktabs}
\usepackage{placeins}
\usepackage{float}
\usepackage{tikz}
\usetikzlibrary{arrows.meta,positioning,calc,fit,backgrounds}
\ificlrpreprintbuild
  \usepackage{hyperref}
\fi
\newtheorem{theorem}{Theorem}

\newtheoremstyle{resultbreak}
  {\topsep}
  {\topsep}
  {\itshape}
  {}
  {\bfseries}
  {.}
  {\newline}
  {\thmname{#1}\thmnumber{ #2}%
   \thmnote{\space\fontseries{bx}\selectfont(#3)}}
\theoremstyle{resultbreak}
\newtheorem{resulttheorem}[theorem]{Theorem}
\newtheorem{resultlemma}[theorem]{Lemma}
\newcommand{\resultstatementheading}[1]{%
  \noindent{\bfseries #1}\par\noindent}
\theoremstyle{definition}
\newtheorem{definition}[theorem]{Definition}
\newtheoremstyle{appendixdefinitionbreak}
  {\topsep}
  {\topsep}
  {\normalfont}
  {}
  {\bfseries}
  {.}
  {\newline}
  {\thmname{#1}\thmnumber{ #2}%
   \thmnote{\space\bfseries(#3)}}
\theoremstyle{appendixdefinitionbreak}
\newtheorem{appendixdefinition}[theorem]{Definition}
\newtheorem{appendixassumption}[theorem]{Assumption}

\title{Distortion-Free Semantic Watermarks for Black-Box Generative Models}

\ificlrpreprintbuild
  \hypersetup{
    hidelinks,
    pdftitle={Distortion-Free Semantic Watermarks for Black-Box Generative Models},
    pdfauthor={Aviv Yossef; Or Patashnik; Or Zamir}
  }
\fi

\author{Aviv Yossef \\
Tel Aviv University
\And
Or Patashnik \\
Tel Aviv University
\And
Or Zamir \\
Tel Aviv University
}

\newcommand{\blfootnote}[1]{%
  \begingroup
  \renewcommand{\thefootnote}{}%
  \footnotetext{#1}%
  \endgroup
}

\ificlrpreprintbuild
  \iclrpreprintcopy
\fi
\begin{document}

\blfootnote{Code is available at \url{https://github.com/avivyossef/distortion-free-semantic-watermark}.}

\maketitle

\begin{abstract}
We present a provably distortion-free semantic watermarking algorithm that requires only black-box sampling access to a generative model and black-box access to a semantic embedding. 
A watermark is called distortion-free if, averaged over its secret detection key, it leaves the generator's output distribution unchanged; this provides a formal guarantee that watermarking does not reduce output quality. 
Conventional such watermarks are often brittle to transformations. 
This is especially problematic for images, where benign operations such as compression, cropping,  or resizing substantially change the entire pixel representation. 
To this end, semantic watermarks introduce the requirement that watermark detection should be robust to transformations that preserve content. 
Previous methods combining these properties rely heavily on specific white-box properties of the underlying model: either to autoregressive text generation or to diffusion-specific access, such as control of the initial noise or diffusion inversion. 
In contrast, our construction is fully generic: It only interacts with the generative model and the semantic embedding, a function that determines the notion of ``preserving content'' under which detection should be robust, via sample and query access. 
We then instantiate this framework for image generation. We characterize the two competing properties required of the semantic embedding: it should remain stable under transformations that preserve an image’s content, but should distinguish independent generations from the same prompt. 
We train an image encoder directly for this tradeoff. 
The resulting encoder improves separation while maintaining stability over off-the-shelf alternatives and enables strong aggregate detection while remaining robust to various transformations.
We demonstrate the effectiveness of our watermark scheme on both open-source and closed models, which is possible only due to restriction to black-box access.

\end{abstract}

\vspace{-10pt}
\section{Introduction}

Generative models can produce large amounts of text and images for which the original source is difficult to determine. Watermarking addresses this problem by embedding a signal that can later be detected using a secret key. Such a signal can be used to reliably attribute content to a particular model or service.

A watermark should not noticeably degrade the model's output. 
\cite{christ2024undetectable} introduced a theoretical framework to guarantee the preservation of quality via the notion of cryptographic undetectability: without a (secret) watermark detection key, no efficient observer should be able to distinguish watermarked outputs from ordinary model outputs; in particular, no benchmark may show any degradation. 
A weaker variant of the above notion is distortion-freeness, under which the marginal distribution of a (single) watermarked output, averaged over the secret key, is identical to the original output distribution. 
This approach was subsequently adopted in practical systems. DeepMind introduced a configurable distortion-free version of SynthID-Text and deployed it in Gemini~\citep{dathathri2024scalable}. Anthropic later adopted a version of SynthID-Text for Claude~\citep{anthropic2026claude}. 
The resulting deployments illustrate an important advantage of distortion-free watermarking: the quality-preservation guarantee circumvents reevaluation for every benchmark or class of prompts.

Distribution preservation, however, does not guarantee another ideal property of watermarking: robustness to edits. 
A watermark may be detectable in the original output but disappear after the output is slightly modified. 
The problem with editing is highly relevant for images. 
Even mundane operations such as JPEG compression, cropping, resizing, or applying a filter can change essentially the entire pixel representation of the image. 
A watermark tied directly to that pixel representation may therefore disappear even though the content of the image remains the same. 
This motivates semantic watermarking, in which detection should be robust to transformations that preserve the content represented by the image. 

Watermarking the semantic meaning of a model's output is also arguably far more robust to watermark removal attempts. Usually, some of the semantic components of an output are requested in the prompt, but some of them are randomly chosen by the model during generation.
An adversary that only sees the output does not know which semantic components were inherent to the prompt and which were not, so they should not be able to remove the model's semantic choices without possibly degrading the intended meaning of the output substantially. 
This comes in sharp contrast to syntactic attacks, such as paraphrasing of sorts, that completely change the syntax without hurting the intended meaning of the output.

Existing semantic watermarking methods typically rely on specific ``white-box'' properties of the underlying generation process. 
For text, this is less restrictive because many models share an autoregressive token-by-token interface. 
Image watermarks, on the other hand, often require control over a particular diffusion latent space or noise distribution, as well as a model-specific inversion procedure. 
Such access is unavailable through closed image-generation APIs and any change in the generation architecture may render the watermark scheme inapplicable.

We introduce the first watermarking framework that is simultaneously distortion-free, semantically robust, and uses only black-box access with respect to both the generative model and the semantic embedding, making it applicable to general generative models. 
We access the generative model only as an oracle that returns independent samples and the semantic embedding only as an oracle that maps a completed output to a vector, without assuming how either oracle is implemented.

Our algorithm samples several candidate outputs, maps them to semantic codes depending on their semantic embeddings, and selects one using a secret keyed function. 
By accounting for the number of candidates that received each code, we ensure that the selected output has exactly the generator's original marginal distribution when averaged over the key and the algorithm's randomness. 
This holds for every generator and semantic embedding. 
If an edit preserves the semantic representation, the watermark is likely to survive (Appendix~\ref{app:semantic_robustness}). 
We prove these guarantees for any two oracles, without requiring a particular embedding architecture or model architecture.
This generality comes at a cost: single-output detection has a non-negligible false-positive rate and limited true-positive rate. This limitation is inherent to black-box output selection: selecting among \(N\) independent candidates gives single-output detection rates satisfying \(\mathrm{TPR}\leq N\,\mathrm{FPR}\) (Appendix~\ref{app:single-image-sampling-limit}). We therefore focus on settings where multiple outputs can be attributed to the same source, such as images uploaded by one social-media profile or assignments submitted by one student. Aggregating the watermark detection scores across this batch of outputs enables high-confidence detection.

To apply our framework to images, we need a semantic embedding that distinguishes different images generated from the same prompt, while remaining stable under edits that preserve an image's content. Standard semantic encoders may intentionally treat such independent generations from the same prompt as similar because they depict the same high-level concept.
For watermarking, however, we need to observe the variation among the generated outputs that remains distinguishable after editing. Otherwise, selecting one output rather than another cannot introduce a detectable semantic watermark. 
This requires randomness in the meaning of the generated content itself, beyond randomness in its pixel representation; we call this \emph{semantic entropy} and derive a lower bound on our algorithm's detection probability in terms of this entropy.
We train an image encoder specifically to distinguish outputs generated by the same model from the same prompt using different random seeds, while keeping each image close to edited versions of it.
We compare the resulting encoder with existing image embeddings, evaluating both its stability under content-preserving transformations and its ability to distinguish independent generations across several image-generation models.

We evaluate detection from individual images and batches, robustness to common transformations, image quality, and same-prompt diversity. With \(32\) candidates, we obtain an AUROC of \(0.920\) for one image and \(1.000\) for ten.
Even when reusing a fixed key, our method achieves the lowest FID relative to unwatermarked generations among the evaluated methods, while retaining essentially the same Vendi diversity.
We also apply our framework to OpenAI's closed GPT Image API, using only sampling access, and obtain an AUROC of \(0.978\) for batches of five images with eight candidates per output.

Our main contributions are therefore:
\begin{enumerate}
    \item We give a black-box, distortion-free semantic watermarking framework for general generative models, requiring only sampling access to the generator and a semantic-embedding oracle.
    \item We characterize how the semantic encoder affects detection. We use semantic entropy to quantify the encoder's ability to distinguish independent generations from the same prompt, and prove a lower bound on the true-positive rate that increases with semantic entropy.
    \item For diffusion-image generation, we train an encoder and evaluate image quality, same-prompt diversity, detection, and robustness to content-preserving transformations.
\end{enumerate}

\section{Related works}

Early language-model watermarks embedded statistical signals during autoregressive generation. \citet{kirchenbauer2023watermark} promoted a keyed subset of tokens during decoding. Later work strengthened the distributional guarantees: \citet{christ2024undetectable} formalized cryptographic undetectability, \citet{vaikuntanathan2026undetectableconversations} introduced bundle sampling that exactly preserves the output distribution, \citet{kuditipudi2024robust} used Gumbel-based sampling, and \citet{bahri2026blackbox} obtained distortion-free watermarking using only sampled text, and observed that shared meanings among candidates may weaken the watermark. MUSE selects among samples from black-box tabular generators but is not distortion-free~\citep{fang2026muse}.

Preserving the output distribution does not ensure robustness to edits. Robust text schemes withstand substitutions and paraphrasing, including a construction based on pseudorandom error-correcting codes~\citep{kuditipudi2024robust,christ2024prc}. Semantic methods make the signal depend on meaning: SemStamp uses locality-sensitive hashing, k-SemStamp uses clustering, and PMark considers multiple channel constraints~\citep{hou2024semstamp,hou2024ksemstamp,huo2026pmark}. These methods exploit autoregressive decisions; ours scores outputs, aggregating evidence across independent generations.

Image watermarking has largely relied on diffusion internals. Tree-Ring inserts Fourier patterns into the initial noise, while Gaussian Shading maps watermark information to Gaussian latents~\citep{wen2023treering,yang2024gaussianshading}. Gunn et al.'s image construction uses pseudorandom error-correcting codes for multi-query undetectability and distinguishes Gaussian Shading's single-image guarantee from fixed-key reuse, where they report reduced same-prompt diversity~\citep{gunn2025undetectable}. T2SMark adds per-generation randomness to relax fixed-codeword restrictions~\citep{yang2025t2smark}. These methods require access to the initial noise, which is typically unavailable through image-generation APIs, and diffusion inversion for detection. NoisePrints avoids inversion but requires the generation seed~\citep{goren2026noiseprints}. We instead select among completed outputs.

Images also make the motivation for semantic watermarking clear: a watermark should survive pixel-level changes that preserve content, but not changes to the content itself. Reprompting attacks expose the latter risk by generating different content that retains the original watermark~\citep{muller2025forgery}. SEAL encodes caption-derived information in the initial noise~\citep{arabi2025seal}, although captions may omit appearance or composition. SemBind learns prompt-consistent codes~\citep{zhang2026sembind}, while SLICE separates semantic factors~\citep{gao2026slice}. We instead use direct image representations that remain stable under content-preserving edits yet distinguish independent same-prompt generations, without modifying or inverting generation.

\vspace{-8pt}
\section{Preliminaries}
\vspace{-6pt}
\label{sec:preliminaries}

In this section, we introduce the notation used by our method, with additional formal definitions and discussion provided in Appendix~\ref{app:formal_preliminaries}.
Let
\(G:\mathcal{P}\times\mathcal{S}\to\mathcal{X}\) be a conditional generator,
where \(p\in\mathcal{P}\) is a prompt, \(s\in\mathcal{S}=\{0,1\}^{k}\) is a
seed, and \(\mathcal{X}\) is the output space. For a fixed prompt, we write
\(X=G(p,S)\) for \(S\sim\operatorname{Unif}(\mathcal{S})\), with distribution
\(P_G(\cdot\mid p)\). A semantic encoder
\(E:\mathcal{X}\to\mathbb{R}^{d}\) maps an output to a semantic
representation. For a nonzero representation, let
\[
z(x)=\frac{E(x)}{\|E(x)\|_2},
\qquad
d_E(x,x')=1-\langle z(x),z(x')\rangle.
\]
We use an \(M\)-bit random-hyperplane locality-sensitive hash
\(H_{\mathcal R}:\mathbb{R}^{d}\to\mathcal{C}\), where
\(\mathcal{C}=\{0,1\}^{M}\), which partitions the representation space into
regions identified by binary codes. Here, \(\mathcal R\) denotes the
sampled hyperplane normals. The binary representation also provides a
natural notion of proximity between hash regions: for
\(c,c'\in\mathcal{C}\), the Hamming distance \(d_{\mathrm H}(c,c')\) is the
number of bits on which they differ. The parameter \(r_{\mathrm{ham}}\)
denotes the maximum number of boundary-uncertain bits considered during
detection (Section~\ref{sec:method}). For a key \(\kappa\in\mathcal{K}\),
where \(\mathcal{K}\) is the secret-key space, a keyed pseudorandom function
\(F_\kappa:\mathcal{C}\to[0,1]\) assigns each hash code a secret score.

\vspace{-8pt}
\section{Semantic Encoders for Distortion-Free Watermarking}
\vspace{-6pt}
\label{sec:semantic}

Distortion-freeness imposes an inherent entropy requirement. If a prompt deterministically produces one output, any distortion-free scheme must return it, making watermarked and nonwatermarked outputs indistinguishable. More generally, Christ, Gunn, and Zamir showed that the detection power of distortion-free watermarking is necessarily tied to generation entropy~\citep{christ2024undetectable}.

Semantic robustness creates a stronger limitation: a generator may have high output entropy while producing outputs equivalent under the chosen semantics. For example, the same scene viewed from slightly different viewpoints.
A robust semantic representation should map them to one semantic region, leaving no variation from which to introduce a detectable signal. We therefore require sufficient variation in the semantic representations of outputs sampled from the same model and prompt. We refer to this variation as \emph{semantic entropy}. The definition below quantifies this variation at a chosen semantic resolution.

The semantic embedding determines both detection power and robustness. Detection power grows with the semantic entropy of the generated outputs (Theorem~\ref{thm:semantic-entropy-tpr}), and robustness depends on whether an edit keeps an image close to its original representation in the semantic space (Appendix~\ref{app:semantic_robustness}). Our framework is not specific to images or to a particular embedding architecture. For our image instantiation, however, the embedding must satisfy two complementary requirements: an image and its content-preserving transformations should remain close in the embedding space, while independent same-prompt generations should remain distinguishable.

Black-box encoder access also lets users tailor what counts as a
content-preserving transformation. For example, a layout-based representation
may favor robustness to style changes, whereas a style-based representation
may favor robustness to spatial transformations.
Appendix~\ref{app:encoder_dependent_robustness} demonstrates these contrasting
behaviors across three style edits and three spatial transformations,
without changing the watermarking algorithm.

However, the chosen representation must also distinguish independent same-prompt generations. Most semantic encoders are designed to separate images associated with different concepts or prompts, rather than different generations of the same prompt. Yet changing the random seed can substantially alter content and composition even when the prompt remains fixed~\citep{xu2025goodseed}. We therefore train the encoder to capture these instance-level differences.

Figures~\ref{fig:semantic_encoder} and~\ref{fig:semantic_encoder_training_pairs} in Appendix~\ref{app:semantic_encoder_training} illustrate the representation requirements and training pairs: transformations of one image should preserve its representation, whereas an independent same-prompt generation may receive a different representation. Existing representations are not explicitly optimized for both requirements: CLIP emphasizes image--text alignment, DINOv2 invariance across augmented views, and TPIPS perceptual similarity. We therefore fine-tune DINOv2 with a contrastive objective that treats content-preserving transformations as positives and different-seed generations from the same prompt as negatives. Appendix~\ref{app:semantic_encoder_training} provides architecture and training details.

Unlike exact-output entropy, which treats every distinct image as a different
outcome, semantic entropy measures how diverse independent same-prompt
generations are in the encoder space. For a fixed semantic radius
\(\varepsilon\), semantic entropy is high when two independent generations
are unlikely to lie within distance \(\varepsilon\) of one another.

\begin{definition}[Semantic entropy]
\label{def:semantic_entropy}
Fix \(\varepsilon\in[0,2]\), and let
\(X,X'\overset{\mathrm{i.i.d.}}{\sim}P_G(\cdot\mid p)\) be two independent
generations for prompt \(p\). We define
\[
H_{\mathrm{sem}}^{E,\varepsilon}(G\mid p)
=
-\log
\Pr_{X,X'}\!\left[d_E(X,X')\leq\varepsilon\right].
\]
\end{definition}

The choice of \(\varepsilon\) determines what counts as meaningfully different.
If \(\varepsilon\) is very small, semantic entropy can be high simply because
two generations are rarely almost identical, even when they remain
semantically very similar. A larger \(\varepsilon\) gives a stronger test:
high semantic entropy then means that independent generations remain separated
even at a meaningful semantic radius.

Because enumerating all seeds is impractical, given a prompt \(p\), we estimate
semantic entropy from \(n\) independent samples \(X_1,\ldots,X_n\sim P_G(\cdot\mid p)\)
and report \(\widehat H_{\mathrm{sem}}^{E,\varepsilon}\). We estimate the probability
inside the logarithm by the fraction of pairs whose distance is at
most \(\varepsilon\):
\[
\widehat P_{\varepsilon}
=
\binom n2^{-1}
\sum_{1\leq i<j\leq n}
\mathbf 1\!\left\{d_E(X_i,X_j)\leq\varepsilon\right\},
\qquad
\widehat H_{\mathrm{sem}}^{E,\varepsilon}
=
-\log \widehat P_{\varepsilon}.
\]

Appendix~\ref{app:semantic_entropy} reports how estimated semantic entropy
varies with \(\varepsilon\). Across the evaluated semantic radii
\(0.3\leq\varepsilon\leq1\), our encoder achieves the highest estimated
semantic entropy among the evaluated encoders for every tested generator.

Semantic entropy captures the representation-level variability available for watermarking. The next section shows how it yields a detectable signal while preserving the generator's output distribution.

\section{Method}
\label{sec:method}

We watermark a black-box generator by selecting which of its outputs to
return. Given a prompt, we independently sample \(N\) candidate images, embed
each image using the semantic encoder \(E\), and map its normalized embedding
\(z(x)\) to a semantic region \(c=H_{\mathcal R}(z(x))\) using a fixed LSH
function \(H_{\mathcal R}\). For a chosen code length \(M\), our scheme
samples \(M\) independent uniform unit hyperplane normals \(\mathcal R\)
once, independently of the secret key, and reuses the resulting
\(H_{\mathcal R}\) for generation and detection. We assign each semantic region a pseudorandom
score \(F_\kappa(c)\), determined by a secret key \(\kappa\) that we reuse
across generations and during detection. We then use these scores to select
the output image. By assigning the same score to images in the same semantic
region, we aim to make the watermark robust to non-semantic transformations.
In particular, when a transformation preserves an image's semantic region,
it also preserves its score and therefore the detector's decision.

The pseudorandom scoring function allows us to construct a distortion-free
selection rule. When all \(N\) candidates belong to distinct semantic
regions, each region has an equal probability of receiving the highest score
over the random secret key, under the ideal-PRF assumption
(Assumption~\ref{ass:ideal-prf}). Thus, selecting the highest-scoring
candidate gives each candidate probability \(1/N\). As we prove in
Appendix~\ref{app:distortion_free_proof}, this uniform-selection property is
sufficient to guarantee distortion-freeness after averaging over the secret
key. However, multiple candidates may share the same semantic region,
breaking this simple argument. We first address this problem with representative sampling (Algorithm~\ref{alg:sampling}), an intuitive construction inspired by bundle sampling that selects one representative uniformly from each semantic region and
compensates for the remaining candidates through an additional
random-selection step. We describe the full construction in
Appendix~\ref{app:algorithmic_specification}.

\begin{figure}[H]
\centering
\setlength{\abovecaptionskip}{4pt}
\resizebox{0.94\linewidth}{!}{\input{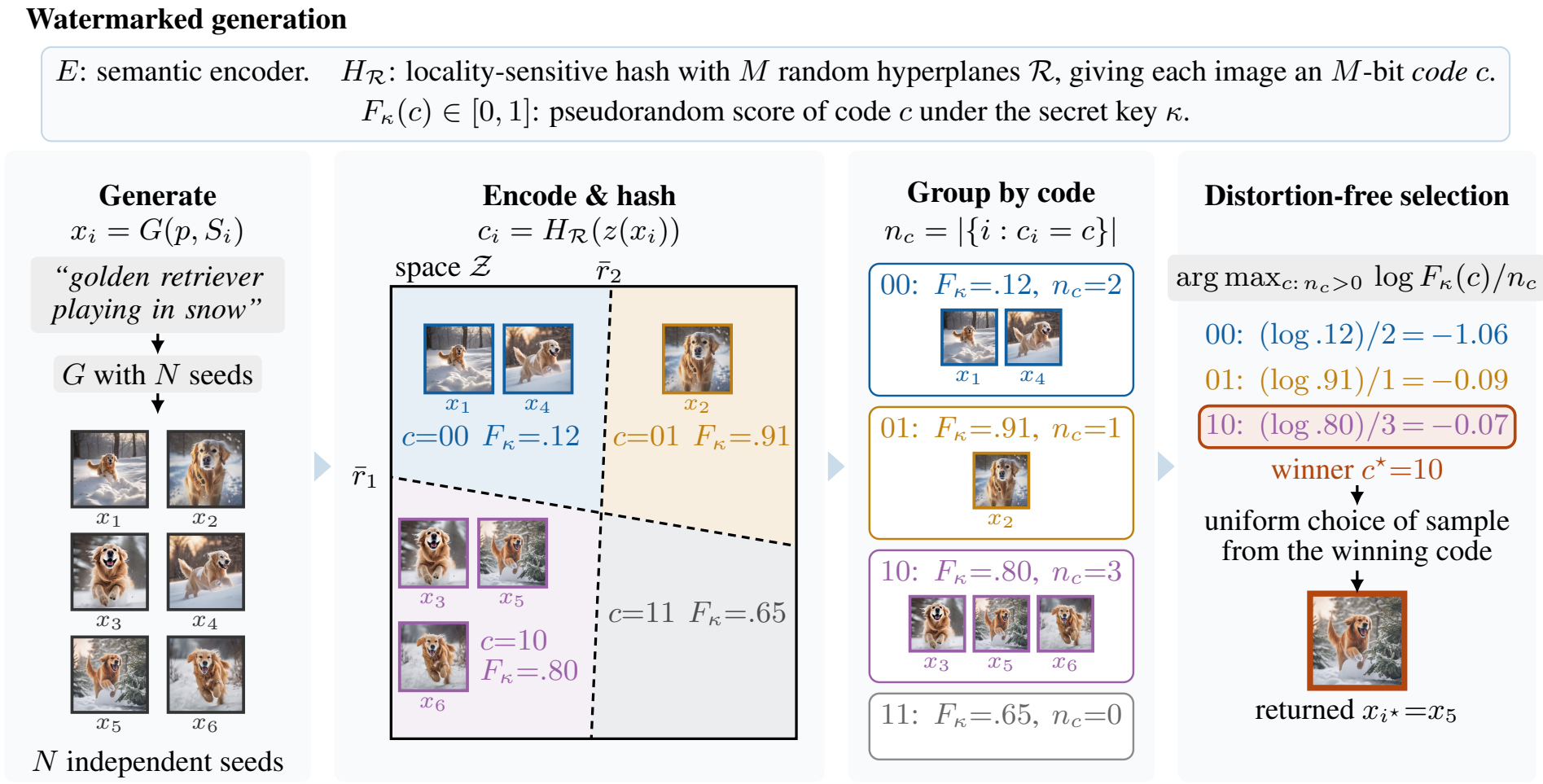}}
\caption{\textbf{Method overview.}
Candidates are encoded, hashed, and grouped by semantic code; each occupied
code \(c\) has a keyed score \(F_\kappa(c)\) and \(n_c\) members. Gumbel
selection (Algorithm~\ref{alg:gumbel_sampling}) returns a uniformly random
member of the code maximizing \(\log F_\kappa(c)/n_c\). Averaged over keys,
code \(c\) wins with probability \(n_c/N\), so each candidate is returned
with probability \(1/N\) and the output follows the generator's
distribution (Theorem~\ref{thm:distortion-free}). With a fixed key, the
preference for particular codes supplies the signal the detector tests.
Other distortion-free rules, such as representative sampling in
Appendix~\ref{app:algorithmic_specification}, can replace the selection
stage.}
\label{fig:method}
\vspace{-4pt}
\end{figure}

Representative sampling is not the only distortion-free selection rule. We
therefore compare several sampling rules in
Appendix~\ref{app:alternative_sampling_rules} and find that Gumbel-based
selection~\citep{kuditipudi2024robust} achieves the strongest empirical
detection performance among the methods we evaluate. We consequently
replace representative sampling with the Gumbel-based sampler in our final
generation algorithm. Algorithm~\ref{alg:gumbel_sampling} and
Figure~\ref{fig:method} illustrate this procedure, and
Appendix~\ref{app:distortion_free_proof} proves the distortion-freeness of
both sampling rules.

\begin{algorithm}[H]
\caption{Watermarked generation with Gumbel selection}
\label{alg:blackbox_watermark}
\label{alg:gumbel_sampling}
\begin{algorithmic}[1]
\Statex \textbf{Input:} Prompt \(p\) and number of candidates \(N\)
\Statex \textbf{Parameters:} Generator \(G\), encoder \(E\), and secret key \(\kappa\)
\For{\(i=1,\ldots,N\)}
    \State Draw \(S_i\sim\operatorname{Unif}(\mathcal S)\) independently; set \(x_i\gets G(p,S_i)\)
    \State Set \(c_i\gets H_{\mathcal R}(z(x_i))\)
\EndFor
\State \(\mathcal C_N\gets\{c_i:i\in[N]\}\)
\State Form \(I_c\gets\{i:c_i=c\}\) and \(n_c\gets|I_c|\) for each \(c\in\mathcal C_N\)
\State \(c^\star\gets\arg\max_{c\in\mathcal C_N}\log F_\kappa(c)/n_c\), breaking ties uniformly
\State Draw \(i^\star\sim\operatorname{Unif}(I_{c^\star})\)
\State \Return \(x_{i^\star}\)
\end{algorithmic}
\end{algorithm}

For detection, we require access only to the semantic encoder and the keyed
scoring function, without access to the original generator. Our basic
detector (Algorithm~\ref{alg:exact_code_detection}) computes the semantic
score of an image and declares it watermarked if the score is at least
\(\tau\). For batch detection, we average the image scores before applying
the threshold.

\begin{algorithm}[H]
\caption{Exact-code detection from one or more images}
\label{alg:exact_code_detection}
\begin{algorithmic}[1]
\Statex \textbf{Input:} Images \(y_1,\ldots,y_B\), where \(B\geq1\)
\Statex \textbf{Parameters:} Encoder \(E\), secret key \(\kappa\), and threshold \(\tau\)
\For{\(j=1,\ldots,B\)}
    \State \(c_j\gets H_{\mathcal R}(z(y_j))\)
    \State \(s_j\gets F_\kappa(c_j)\)
\EndFor
\State \Return \(\mathbf 1\{B^{-1}\sum_{j=1}^{B}s_j\geq\tau\}\)
\end{algorithmic}
\end{algorithm}

However, non-semantic transformations may change the semantic region of an
image near a region boundary, potentially changing its score. To address
this limitation, we introduce a boundary-aware detector that also considers
nearby semantic regions when computing the detection score. We use this
detector in all our experiments and provide its full description in
Appendix~\ref{app:algorithmic_specification}.

Having introduced our generation and detection algorithms, we next establish
their theoretical guarantees. In Section~\ref{sec:analysis}, we prove that
our generation algorithm is distortion-free and analyze how the detector's
performance depends on the semantic entropy of the generator.

\section{Theoretical Guarantees}
\label{sec:analysis}

We now establish the theoretical guarantees of our watermarking scheme.
First, we prove that our generation algorithm is distortion-free for any
black-box generative model and semantic encoder, after averaging over the
secret key. Second, we establish the detector's false-positive rate and prove
that sufficiently high semantic entropy guarantees strong detection for an
appropriate choice of parameters. Finally, we establish guarantees for batch
detection and robustness to transformations that approximately preserve the
semantic representation.

\stepcounter{theorem}
\newcounter{uniformselectionnumber}
\setcounter{uniformselectionnumber}{\value{theorem}}

\begin{samepage}
\begin{resulttheorem}[Distortion-free generation]
\label{thm:distortion-free}
Let \(G\) be any conditional generative model and \(E\) any encoder for
which Algorithm~\ref{alg:gumbel_sampling} is well-defined. Under the ideal-PRF
assumption (Assumption~\ref{ass:ideal-prf}), the watermarked generator defined
by Algorithm~\ref{alg:gumbel_sampling} is distortion-free
(Definition~\ref{def:distortion-free}): for every prompt \(p\),
\[
\mathbb E_{K\sim\operatorname{Unif}(\mathcal K)}
\left[P_{G_K^{\mathrm{wm}}}(\cdot\mid p)\right]
=P_G(\cdot\mid p).
\]
\end{resulttheorem}
\end{samepage}

We prove Theorem~\ref{thm:distortion-free} in
Appendix~\ref{app:distortion_free_proof}. This ensures that, regardless of the semantic distribution of the outputs, the watermarking generator's output is also distributed according to the original model’s distribution when we average over the random choice of the key.

We next establish a lower bound on the detection advantage of our
watermarking scheme and analyze how it depends on the semantic entropy of
the generator \(G\) and encoder \(E\) for a given prompt \(p\), as well as on
three configurable parameters:
\begin{itemize}
\item Number of candidates \(N\): The number of black-box generator
calls per watermarked output. Larger \(N\) increases the potential detection
advantage, at a generation cost that scales linearly with \(N\).
\item Number of hash bits \(M\): Increasing \(M\) reduces hash
collisions, improving the detection bound but potentially decreasing
robustness, since small changes in the semantic representation are more
likely to change the hash code.
\item Detection threshold \(\tau\): Controls the tradeoff between
false-positive and true-positive rates.
\end{itemize}
The guarantee also depends on the semantic entropy
\(H_{\mathrm{sem}}^{E,\varepsilon}(G\mid p)\), defined in
Section~\ref{sec:semantic}. Higher semantic entropy strengthens our
detection guarantee.

\begin{resulttheorem}[Semantic entropy and detection]
\label{thm:semantic-entropy-tpr}
Fix a prompt \(p\) and a semantic radius \(\varepsilon\in[0,2]\). Generate
\(N\geq2\) independent candidates and apply Algorithm~\ref{alg:gumbel_sampling}. Under the ideal-PRF assumption,
for single-image detection, the exact-code detector
(Algorithm~\ref{alg:exact_code_detection}) at threshold \(\tau\in[0,1]\) satisfies
\[
\operatorname{FPR}(\tau)
=1-\tau
\]
and
\[
\operatorname{TPR}^{(N)}(\tau)
\geq
1-\tau
+\frac{1-q_{M,\varepsilon}(p)}
{1+(N-1)q_{M,\varepsilon}(p)}
\left(\tau-\tau^N\right),
\]
where
\[
q_{M,\varepsilon}(p)
=e^{-H_{\mathrm{sem}}^{E,\varepsilon}(G\mid p)}
+\left(1-e^{-H_{\mathrm{sem}}^{E,\varepsilon}(G\mid p)}\right)
\left(1-\frac{\arccos(1-\varepsilon)}{\pi}\right)^M.
\]
Both rates average over the secret key and random hyperplanes for the fixed
prompt \(p\).
\end{resulttheorem}

Theorem~\ref{thm:semantic-entropy-tpr} shows that higher semantic entropy
strengthens the guaranteed detection advantage without increasing the
false-positive rate. For a suitable choice of parameters, sufficiently high
semantic entropy therefore guarantees a substantial separation between
watermarked and unwatermarked outputs. These guarantees average over the
secret key and random hyperplanes. We provide the proof in
Appendix~\ref{app:semantic_entropy_tpr}.

Finally, we prove in Appendix~\ref{app:batch_detection} that batch detection
reduces both error probabilities exponentially with the number of
independently generated outputs, provided their expected clean and
watermarked scores lie on opposite sides of the threshold for a fixed key.
In Appendix~\ref{app:semantic_robustness}, we establish that transformations
preserving semantic similarity are likely to preserve the hash code and
hence the watermark score.

We next evaluate our method empirically. We first examine image quality and
diversity under fixed-key reuse, since our distortion-free guarantee
averages over the secret key (Section~\ref{sec:exp_quality_diversity}). We
then evaluate single-image and batch detection and robustness to
content-preserving transformations (Section~\ref{sec:exp_detection}), and
compare semantic entropy across different encoders
(Section~\ref{sec:encoder_analysis}).

\section{Experiments and Results}
\label{sec:experiments}

\vspace{-6pt}
\paragraph{Experimental setup.} 
We conduct the watermarking experiments using Stable Diffusion~2.1~\citep{stabilityai2022sd21, rombach2022high} and COCO-2017 prompts~\citep{lin2014coco,chen2015cococaptions} held out from semantic encoder training.
Our black-box method selects among complete candidate images. Because sampling-based watermarks primarily target language while image methods typically access the generation process, no directly comparable image-watermarking baseline shares the same sampling-only access model. We therefore compare with Tree-Ring~\citep{wen2023treering}, Gaussian Shading~\citep{yang2024gaussianshading}, and SEAL~\citep{arabi2025seal}, three closely related methods for watermarking diffusion-generated images. We evaluate all methods on a common generation backbone using each method's native generation and detection procedures.
For representation analysis, we compare our encoder with DINOv2~\citep{oquab2024dinov2}, CLIP~\citep{radford2021clip}, DINOv3~\citep{simeoni2025dinov3}, and TPIPS~\citep{wang2026tpips} on independent generations from SD2.1 and two FLUX variants~\citep{blackforestlabs2024flux} and content-preserving perturbations of real images.

In addition to the main experiments, which test watermarking quality,
detectability, and robustness, Appendix~\ref{app:additional_results}
presents several extensions. We demonstrate closed-API use with GPT Image
through black-box sampling (Appendix~\ref{app:closed_api_generation}) and
examine our method in a different modality, audio
(Appendix~\ref{app:audio_watermark}). We also evaluate detection in mixed
batches (Appendix~\ref{app:mixed_source_batches}), test whether edits that
replace the main object remove the watermark signal
(Appendix~\ref{app:content_change_detection}), and examine how the choice of
semantic encoder affects robustness
(Appendix~\ref{app:encoder_dependent_robustness}).

\vspace{-4pt}
\subsection{Generation Quality and Diversity}
\vspace{-4pt}
\label{sec:exp_quality_diversity}

Our distortion-free guarantee matches the original generator distribution after averaging over secret keys; it does not require each fixed key to preserve that distribution. Because deployment typically reuses one key across many generations, we empirically evaluate fixed-key quality and diversity. Prior work shows that fixed-key reuse can reduce same-prompt diversity for some initial-noise watermarking methods~\citep{gunn2025undetectable,yang2025t2smark}. We therefore measure fixed-key distributional deviation and same-prompt diversity; Appendix~\ref{app:quality_alignment_table} reports prompt alignment.

We evaluate Stable Diffusion~2.1 Base at \(512 \times 512\). We measure distributional deviation by computing FID between each watermarked population and its matched unwatermarked population, using 10,000 generations from 1,000 COCO-2017 prompts at 10 generations per prompt. To assess same-prompt diversity, we generate 60 images for each of 60 prompts and measure the relative Vendi Score in a frozen DINOv2-base feature space, independent of our semantic encoder. Each configuration uses an unwatermarked reference set with the same model, prompts, and sampling settings.

\begin{table}[!ht]
\centering
\small
\setlength{\tabcolsep}{10pt}
\caption{\textbf{Fixed-key generation quality and diversity.} FID between 10,000 matched watermarked and clean generations from 1,000 COCO-2017 prompts; Relative FID divides by the clean-vs-clean reference of 4.64. Diversity change is $100(\mathrm{Vendi}_{\rm wm}/\mathrm{Vendi}_{\rm clean}-1)$ in DINOv2-base space over 60 prompts with 60 generations each.}
\label{tab:experiment3_quality_diversity}
\begin{tabular}{@{}lccc@{}}
\toprule
Method & FID $\downarrow$ & Relative FID $\downarrow$ & DINOv2-base diversity change \\
\midrule
Clean vs.\ clean & 4.64 & 1.000 & -- \\
\midrule
Ours \((N,M)=(8,4)\)  & 4.75 & 1.024 & $-0.5\%$ \\
Ours \((N,M)=(16,8)\) & 5.04 & 1.086 & $-1.2\%$ \\
Ours \((N,M)=(32,12)\) & 5.22 & 1.125 & $-1.0\%$ \\
\midrule
Tree-Ring             & 5.61 & 1.209 & $+0.4\%$ \\
Gaussian Shading      & 5.42 & 1.168 & $-9.1\%$ \\
SEAL                  & 6.88 & 1.483 & $-13.2\%$ \\
\bottomrule
\end{tabular}
\end{table}
\FloatBarrier

Table~\ref{tab:experiment3_quality_diversity} reports both quantities. Across our three configurations, matched-population FID ranges from 4.75 to 5.22, averaged across 10 key selections, compared with 5.42--6.88 for the baselines. Relative to the clean-versus-clean reference of 4.64, these ranges are 1.024--1.125 and 1.168--1.483, respectively. Same-prompt diversity remains within \(1.3\%\) of clean generation for our method and is unchanged for Tree-Ring, whereas Gaussian Shading reduces it by \(9.1\%\) and SEAL by \(13.2\%\). Together, these results show that, among the evaluated methods, our method has the smallest measured distributional shift under fixed-key reuse while retaining near-clean same-prompt diversity. Appendix~\ref{app:quality_alignment_table} reports complementary FID against real COCO-2017 images and image-prompt alignment results.

\subsection{Watermark Detection and Robustness}
\label{sec:exp_detection}

We evaluate detection on clean outputs and after common content-preserving transformations. %

\paragraph{Clean images and image sets.}
Using 1,000 held-out COCO-2017 prompts, we evaluate matched watermarked and unwatermarked generations for $(N,M)\in\{(8,4),(16,8),(32,12)\}$ and set sizes $B\in\{1,5,10\}$. We report AUROC and TPR at empirical FPRs of $1\%$, $5\%$, and $20\%$. Tree-Ring, Gaussian Shading, and SEAL provide native single-image references; Appendix~\ref{app:robustness_tables} reports the ROC curves and aggregation protocol.

Table~\ref{tab:experiment3_detection} shows that aggregation improves detection: for $(N,M)=(32,12)$, TPR@20\% rises from $0.920$ for one image to $0.999$ for five and $1.000$ for ten. While native baselines are stronger for single-image detection because they modify the generation process, our black-box method instead selects among $N$ complete outputs and preserves its distortion-free guarantee. Consequently, it faces a theoretical single-image ceiling, $\mathrm{TPR}\leq N\,\mathrm{FPR}$, inherent to black-box sampling (Appendix~\ref{app:extended_limitations}).

We also test the framework through OpenAI's closed GPT Image API, using eight candidates per output. With only sampling access to the generator, we obtain an AUROC of \(0.815\) for individual images and \(0.978\) for batches of five (Appendix~\ref{app:closed_api_generation}).

\begin{table}[!ht]
\centering
\small
\setlength{\tabcolsep}{3.5pt}
\caption{Clean detection on 1,000 matched prompts. TPR@20\% is the true-positive rate read at 20\% FPR from the mean ROC curve. A dash means that only single-image detection
was evaluated for that native baseline.}
\label{tab:experiment3_detection}
\begin{tabular}{lcccccc}
\toprule
& \multicolumn{2}{c}{$B=1$} & \multicolumn{2}{c}{$B=5$} &
  \multicolumn{2}{c}{$B=10$} \\
\cmidrule(lr){2-3}\cmidrule(lr){4-5}\cmidrule(lr){6-7}
Method & AUROC & TPR@20\% & AUROC & TPR@20\% & AUROC & TPR@20\% \\
\midrule
Ours $(N=8,M=4)$           & 0.802 & 0.642 & 0.938 & 0.905 & 0.960 & 0.941 \\
Ours $(N=16,M=8)$          & 0.879 & 0.815 & 0.991 & 0.994 & 0.998 & 0.998 \\
Ours $(N=32,M=12)$          & 0.920 & 0.920 & 0.998 & 0.999 & 1.000 & 1.000 \\
\midrule
Tree-Ring                  & 1.000 & 1.000 & -- & -- & -- & -- \\
Gaussian Shading           & 1.000 & 1.000 & -- & -- & -- & -- \\
SEAL                       & 1.000 & 1.000 & -- & -- & -- & -- \\
\bottomrule
\end{tabular}
\end{table}

\paragraph{Robustness under common transformations.}
To approximate deployment conditions, we evaluate images after transformations commonly encountered when they are stored, shared, or edited: appearance changes, compression, geometric transformations, and localized edits produced by a generative image-editing model. For the configurations $(N,M)\in\{(8,4),(16,8),(32,12)\}$, we compare with the baseline methods in the single-image setting. We report both detection after transformation and retention, defined as the fraction of images detected before transformation that remain detected afterward. This distinction separates robustness from the initial detection strength of each method. Appendix~\ref{app:robustness_protocols} details transformations, detector configurations, calibration, and settings.

\begin{figure}[t]
    \centering
    \resizebox{\linewidth}{!}{\input{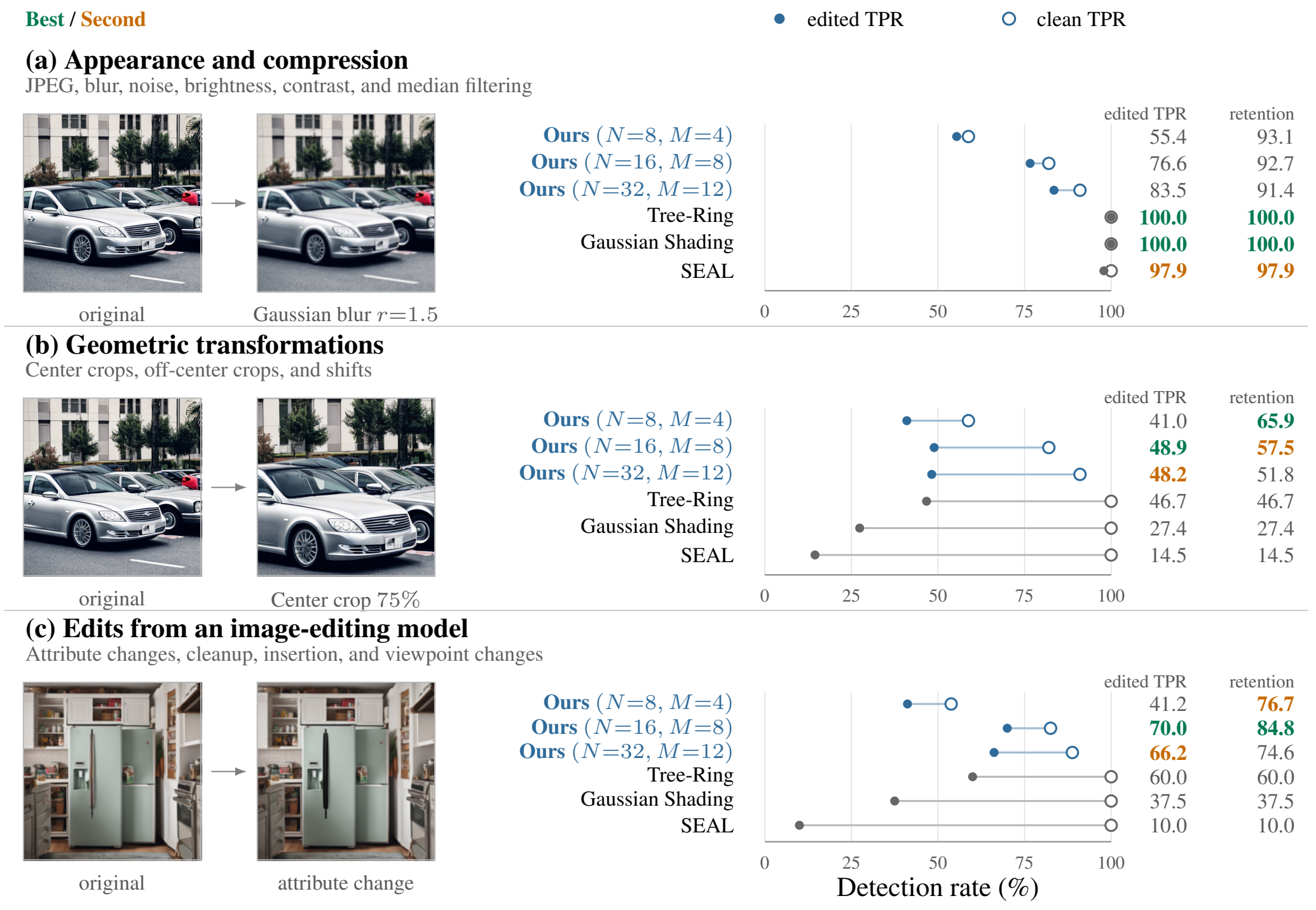}}
    \caption{Robustness at 20\% nominal clean FPR. Filled and open markers show edited and clean detection; retention is the fraction of initial detections preserved after editing. Panels (a,b) cover appearance/compression and geometric transformations; panel (c) covers 80 generative edits. Colors mark the best and second-best values. Appendix~\ref{app:robustness_protocols} gives detector configurations, per-setting results, and FPRs.}
    \label{fig:experiment3_robustness_categories}
\end{figure}

Figure~\ref{fig:experiment3_robustness_categories} shows different patterns across transformation types. The baselines are stronger under appearance changes and compression, whereas all three of our configurations retain more initial detections under geometric transformations. For $(N,M)=(16,8)$ and $(32,12)$, detection after geometric transformation reaches $48.9\%$ and $48.2\%$, above Tree-Ring ($46.7\%$), Gaussian Shading, and SEAL. Thus, detection after transformation and retention provide complementary views of robustness. Tables~\ref{tab:robustness_single_lowfpr} and~\ref{tab:robustness_lowfpr_batches} in Appendix~\ref{app:robustness_protocols} report TPR and retention at $1\%$ and $5\%$ nominal clean FPR for single images and batches of edited images.

For the 80 localized edits made with a generative image editor, $(N,M)=(16,8)$ achieves the highest TPR after editing ($70.0\%$), followed by $(32,12)$ ($66.2\%$), both exceeding Tree-Ring ($60.0\%$), Gaussian Shading ($37.5\%$), and SEAL ($10.0\%$). $(16,8)$ also has the highest retention ($84.8\%$). In contrast, $(32,12)$ detects the most images before editing ($88.8\%$) but keeps only $74.6\%$ of those detections. This matches the tradeoff in Section~\ref{sec:analysis}: more hash bits strengthen detection but make the code more sensitive to edits.

\subsection{Semantic Representation Analysis}
\label{sec:encoder_analysis}

Our watermarking method relies on a semantic representation with two complementary properties: content-preserving transformations should produce similar representations, while independent generations should remain distinguishable. We analyze whether the learned encoder satisfies these properties, its transfer across generators, and how its geometry relates to watermark retention.

We compare our encoder with DINOv2-base, CLIP ViT-B/32, DINOv3 L/16, and
TPIPS-Embed. To evaluate same-prompt diversity, we generate $16$ images with different
random seeds for each of $200$ COCO-2017 prompts held out from encoder
training, using three generators: SD2.1 and two FLUX variants. Because the
encoder was trained only on SDXL~\citep{podell2023sdxl} images, this also
tests transfer to unseen generators. Appendix~\ref{app:semantic_entropy}
gives full protocol details.

We quantify these properties using mean pairwise cosine similarity, which
measures how closely representations are clustered. For content-preserving
transformations, higher cosine similarity indicates greater invariance. For
independent same-prompt generations, lower cosine similarity indicates greater
diversity in the encoder's representation. We also report semantic entropy
at $\varepsilon=0.5$, counting pairs with cosine similarity at least $0.5$
as close.

Table~\ref{tab:encoder_comparison} compares how the encoders represent
independent generations from the same prompt. Our encoder achieves the lowest
mean cosine similarity and highest estimated semantic entropy at
$\varepsilon=0.5$ for all three evaluated generators, none of which were used
for training. Figure~\ref{fig:semantic_entropy_epsilon} in
Appendix~\ref{app:semantic_entropy} shows how the entropy comparison changes
with $\varepsilon$.

\begin{table}[H]
\centering
\small
\setlength{\tabcolsep}{3pt}
\renewcommand{\arraystretch}{0.95}
\setlength{\abovecaptionskip}{3pt}
\setlength{\belowcaptionskip}{3pt}
\caption{\textbf{Same-prompt diversity in encoder space.} Mean pairwise cosine
similarity and estimated semantic entropy at $\varepsilon=0.5$, across 16
independent generations from the same prompt. Lower cosine similarity and
higher estimated semantic entropy indicate greater diversity in the encoder's
representation. Appendix~\ref{app:semantic_entropy} provides the complete
protocol and stability results under content-preserving transformations.}
\label{tab:encoder_comparison}
\begin{tabular}{@{}l*{10}{r}@{}}
\toprule
 & \multicolumn{2}{c}{DINOv2} & \multicolumn{2}{c}{CLIP} & \multicolumn{2}{c}{DINOv3} & \multicolumn{2}{c}{TPIPS} & \multicolumn{2}{c}{Ours} \\
\cmidrule(lr){2-3}\cmidrule(lr){4-5}\cmidrule(lr){6-7}\cmidrule(lr){8-9}\cmidrule(lr){10-11}
 & Cos.$\downarrow$ & $\widehat H\uparrow$ & Cos.$\downarrow$ & $\widehat H\uparrow$ & Cos.$\downarrow$ & $\widehat H\uparrow$ & Cos.$\downarrow$ & $\widehat H\uparrow$ & Cos.$\downarrow$ & $\widehat H\uparrow$ \\
\midrule
FLUX distilled & 0.808 & 0.033 & 0.902 & 0.000 & 0.783 & 0.054 & 0.826 & 0.000 & \textbf{0.431} & \textbf{0.739} \\
FLUX base & 0.730 & 0.093 & 0.873 & 0.001 & 0.720 & 0.108 & 0.762 & 0.001 & \textbf{0.297} & \textbf{1.104} \\
SD2.1 & 0.634 & 0.252 & 0.863 & 0.001 & 0.651 & 0.217 & 0.645 & 0.039 & \textbf{0.222} & \textbf{1.345} \\
\bottomrule
\end{tabular}

\end{table}

Appendix~\ref{app:semantic_entropy} reports two further encoder analyses.
First, it measures stability under content-preserving transformations; here,
our encoder and DINOv3 are strongest overall: ours under JPEG compression
and resizing, and DINOv3 under Gaussian blur and mild crop-and-color
transformations. Second, it examines the discrete hash codes that the
watermark actually uses. For independent generations of the same prompt, our
encoder produces more distinct codes, and the proof of
Theorem~\ref{thm:semantic-entropy-tpr} shows that more occupied codes increase
detection probability. Under content-preserving transformations, the opposite
is desirable, and our encoder keeps relatively few distinct codes across all
four transformation families, so an edited image tends to keep its original
code. Appendix~\ref{app:encoder_detection_ablation}
isolates the encoder's effect in the complete watermarking pipeline, where ours
achieves the highest mean AUROC in all three evaluated configurations.

Encoder similarity closely tracks watermark retention. Across the $20$
transformation settings, the Spearman correlation between encoder similarity
and retention ranges from $0.988$ to $0.998$ for our configurations, compared
with $0.747$ to $0.779$ for the baseline watermarks. This is consistent with our geometric
analysis: higher cosine similarity makes the original and transformed images
more likely to receive the same LSH code and clean keyed score
(Lemma~\ref{lem:semantic-score-preservation} in
Appendix~\ref{app:semantic_robustness}).
Figure~\ref{fig:robustness_geometry_diagnostic}
in Appendix~\ref{app:additional_results} provides the complete visualization.
Consequently, an encoder that better matches human judgments of whether an
edited image still shows the same scene, while still separating independent
generations, could directly improve detection and robustness.
Appendix~\ref{app:encoder_dependent_robustness} further demonstrates that
replacing the encoder changes which transformations preserve watermark
detection, allowing robustness to be tailored to application-specific image
properties.

\section{Organization of the Appendices}

We organize the appendices to provide the technical foundations, proofs, and additional experimental evidence supporting our results. Appendix~\ref{app:formal_preliminaries} introduces the formal definitions and assumptions underlying our framework, while Appendix~\ref{app:algorithmic_specification} provides the full algorithmic specifications. We then establish our theoretical guarantees. In Appendix~\ref{app:distortion_free_proof}, we prove that our generation algorithm is distortion-free for any black-box generator and semantic encoder. In Appendix~\ref{app:semantic_entropy_tpr}, we show how increasing semantic entropy strengthens the detection guarantee. In Appendix~\ref{app:batch_detection}, we prove that aggregating independently generated outputs improves detection reliability by reducing detection errors exponentially. In Appendix~\ref{app:semantic_robustness}, we establish robustness guarantees for transformations that approximately preserve semantic representations.

To instantiate our framework for images, we train a semantic encoder that separates independent generations from the same prompt while remaining stable under content-preserving transformations. Appendix~\ref{app:semantic_encoder_training} describes the training procedure, and Appendix~\ref{app:semantic_entropy} evaluates the resulting representation. We show that our encoder achieves higher estimated semantic entropy than the evaluated alternatives across multiple image generators, while maintaining stability under common image transformations.

We further investigate the effectiveness and generality of our method through additional experiments. Appendix~\ref{app:detection_robustness_overview} provides the complete detection and robustness evaluation protocols used in experiments, additional results at low false-positive rates, and examples of image transformations. In Appendix~\ref{app:additional_results}, we examine several design choices and extensions. We relate encoder similarity to watermark retention (Appendix~\ref{app:encoder_similarity_retention}). To motivate our algorithmic and encoder choices, we compare our trained encoder against off-the-shelf alternatives and Gumbel sampling against other distortion-free selection rules (Appendices~\ref{app:encoder_detection_ablation} and~\ref{app:alternative_sampling_rules}). We also compare how different embedding discretizers spread candidates across codes and preserve those codes under edits (Appendix~\ref{app:binarization_ablation}). We show that our encoder improves watermark detection and that Gumbel sampling achieves stronger detection than the alternative sampling rules we evaluate. We further demonstrate the flexibility of our black-box framework in two settings. First, by changing the semantic encoder, we can target different notions of robustness without modifying the watermarking algorithm (Appendix~\ref{app:encoder_dependent_robustness}). Second, we demonstrate that our method works with closed-API image generators without accessing their internals. In particular, our GPT Image experiment achieves an AUROC of 0.978 when aggregating five images (Appendix~\ref{app:closed_api_generation}). We also watermark text-to-audio outputs (Appendix~\ref{app:audio_watermark}) and evaluate detection in batches that mix images from different keys or include unwatermarked images (Appendix~\ref{app:mixed_source_batches}). Finally, we evaluate generation quality and prompt alignment under fixed-key reuse, finding the smallest measured distributional shift among the evaluated methods while maintaining nearly unchanged prompt alignment (Section~\ref{sec:exp_quality_diversity} and Appendix~\ref{app:quality_alignment_table}).

Appendix~\ref{app:content_change_detection} compares detection on watermarked
and clean images after edits that replace the main subject, scene, or action.

Finally, Appendix~\ref{app:extended_limitations} discusses the limitations of our approach. In particular, we establish an inherent limitation of black-box sampling: when selecting one of $N$ unmodified candidate outputs, the single-image true-positive rate cannot exceed $N$ times the false-positive rate (Appendix~\ref{app:single-image-sampling-limit}). We also discuss computational costs, fixed-key reuse, adaptive attacks, and extensions beyond images.

\section{Limitations and Future Work}
\label{sec:limitations}

Our approach has three main limitations, each pointing to future work:

\begin{itemize}
\item \textbf{Single-image detection.} Our current image implementation provides
weaker clean single-image detection than methods that directly modify the
diffusion process (Table~\ref{tab:experiment3_detection}). This tradeoff
matters most when every image must be detected independently and internal
model access is available. Our method instead prioritizes black-box
applicability, semantic binding, and distribution preservation, while
aggregation substantially strengthens detection
(Table~\ref{tab:experiment3_detection}). More generally, any method that
selects one of $N$ independent, unmodified generator outputs satisfies
\(\mathrm{TPR}\leq N\,\mathrm{FPR}\) for single-image detection
(Theorem~\ref{thm:single-image-sampling-limit} in
Appendix~\ref{app:single-image-sampling-limit}). Thus, achieving constant TPR
at exponentially small FPR requires exponentially many generator queries. This
$O(N)$ ceiling belongs to black-box sampling, not our construction.

\item \textbf{Semantic encoder.} A better semantic encoder could improve
detection within this limit and strengthen robustness. It must keep
content-preserving transformations close while separating independent
generations; ours improves this balance over the evaluated alternatives
(Appendix~\ref{app:encoder_detection_ablation}).

\item \textbf{Computational cost.} Our strongest configuration generates 32
complete candidates per released image, which may be impractical for
large-scale or latency-sensitive applications. Early-selection methods may
reduce this cost by using intermediate denoiser activations to predict quality
or reusing low-resolution semantic features across denoising
steps~\citep{guo2026earlyquality,habibian2024clockwork,parmar2026scaling}. If
these signals predict the final semantic code, unpromising candidates could be
discarded early.
\end{itemize}

In Appendix~\ref{app:extended_limitations} we discuss these limitations at length as well as further open problems.

\section{Conclusion}

We introduced a black-box, distortion-free semantic watermarking framework
requiring only sampling access to a generator and query access to a semantic
encoder. Our analysis connects detectability to semantic entropy, while
experiments demonstrate aggregate detection and robustness to common
transformations without compromising generation quality or diversity. These
results establish semantic rejection sampling as a general black-box foundation
for generative watermarking. Stronger encoders and more efficient sampling
remain directions for future work.

\section{AI Use Statement, Reproducibility and Acknowledgments}

Generative AI assistants were used to support software development and
analysis workflows, literature discovery, and the editing and formatting of
manuscript text. They were also used as a source of feedback while refining
the presentation of the method, experimental protocol, and theoretical
exposition. We independently designed the study, ran and checked the
experiments, verified code and mathematical statements, and take full
responsibility for the paper's content. The images produced by the generative
models evaluated in this work are experimental outputs generated under the
stated protocols, rather than content produced by these assistants.

The main paper specifies the sampling-based watermarking method, its
theoretical assumptions, and the experimental setup.
Our code is available at
\url{https://github.com/avivyossef/distortion-free-semantic-watermark}.

The research of OZ and AY is supported by grants from the AISI Alignment Project and the Israeli Ministry of Science. 

\bibliography{iclr2027_conference}\bibliographystyle{iclr2027_conference}
\appendix
\section{Formal Preliminaries and Definitions}
\label{app:formal_preliminaries}

\subsection{Generative Models and Sampling}
\begin{appendixdefinition}[Conditional generative model]
Let \(\mathcal{P}\) be the set of all possible prompts, let
\(\mathcal{S} = \{0,1\}^k\) be the set of all possible seeds, and let
\(\mathcal{X}\) be the output space. A conditional generative model is a
mapping
\[
    G \colon \mathcal{P} \times \mathcal{S} \to \mathcal{X}.
\]
\end{appendixdefinition}

We model \(\mathcal{S}\) as a finite set of seeds, as is the case in
practical implementations, whereas the prompt space \(\mathcal{P}\) may
be infinite. For any fixed prompt \(p \in \mathcal{P}\) and seed
\(s \in \mathcal{S}\), the output \(G(p,s)\) is deterministic.

\begin{appendixdefinition}[I.i.d. sample from a conditional generative model]
Fix a prompt \(p \in \mathcal{P}\). Let
\[
    S_1,\ldots,S_n
    \overset{\mathrm{i.i.d.}}{\sim}
    \operatorname{Unif}(\mathcal{S}),
    \qquad
    X_i = G(p,S_i).
\]
Any realization \((x_1,\ldots,x_n)\) of
\((X_1,\ldots,X_n)\) is called an i.i.d.\ sample from \(G\)
conditioned on the prompt \(p\).
\end{appendixdefinition}

\begin{appendixdefinition}[Encoder and semantic encoder]
Let \(\mathcal{Z} = \mathbb{R}^d\) be a \(d\)-dimensional representation
space. An encoder is a mapping
\[
    E \colon \mathcal{X} \to \mathcal{Z}.
\]
For \(x \in \mathcal{X}\), we call \(E(x)\) the representation of \(x\).
We call \(E\) a \emph{semantic encoder} if its representations capture
the semantic content of the outputs.
\end{appendixdefinition}
The precise properties required of a semantic encoder in our framework
are introduced in Section~\ref{sec:semantic}.

\subsection{Watermarking Generative Models}
\begin{appendixdefinition}[Keyed watermarking scheme]
Let \(G\) be a conditional generative model and let \(\mathcal{K}\) be
a finite key space. A keyed watermarking scheme for \(G\) is a pair
\[
    \mathsf{W}
    =
    \bigl(G^{\mathrm{wm}},\mathsf{Detect}\bigr),
\]
where, for every key \(\kappa \in \mathcal{K}\):
\begin{enumerate}
    \item The \emph{watermarked generator} is a possibly randomized mapping
    \[
        G_{\kappa}^{\mathrm{wm}}
        \colon \mathcal{P}\times\mathcal{S}^{N}\to\mathcal{X}.
    \]
    Given a prompt and \(N\) seeds, it produces a watermarked output.
    \item For every batch size \(B\geq 1\), the \emph{detector} is a mapping
    \[
        \mathsf{Detect}_{\kappa,\tau}^{(B)}
        \colon \mathcal{X}^{B}\to\{0,1\}.
    \]
    Given a batch of images \(x_{1:B}=(x_1,\ldots,x_B)\) and a fixed detection
    threshold \(\tau\in[0,1]\), it outputs \(1\) when the batch is detected as
    watermarked under the key \(\kappa\), and \(0\) otherwise. The case \(B=1\)
    corresponds to single-image detection.
\end{enumerate}
\end{appendixdefinition}

\begin{appendixdefinition}[Black-box watermarking scheme]
A watermarking scheme \(\mathsf{W}=(G^{\mathrm{wm}},\mathsf{Detect})\) is
\emph{black-box with respect to \(G\)} if \(G^{\mathrm{wm}}\) and
\(\mathsf{Detect}\) may interact with \(G\) only by evaluating \(G(p,s)\),
and otherwise have no access to its internal implementation.
\end{appendixdefinition}

\begin{appendixdefinition}[Distortion-free watermarking]
\label{def:distortion-free}
Let \(P_G(\cdot\mid p)\) denote the distribution of \(G(p,S)\), where
\(S\sim\operatorname{Unif}(\mathcal{S})\). For each key
\(\kappa\in\mathcal{K}\), let
\(P_{G_{\kappa}^{\mathrm{wm}}}(\cdot\mid p)\) denote the output
distribution of the watermarked generator for prompt \(p\).
The watermarking scheme is \emph{distortion-free} if
\[
    \mathbb{E}_{K\sim\operatorname{Unif}(\mathcal{K})}
    \left[P_{G_K^{\mathrm{wm}}}(\cdot\mid p)\right]
    = P_G(\cdot\mid p)
\]
for every prompt \(p\in\mathcal{P}\).
\end{appendixdefinition}

This is a key-averaged, single-output guarantee: it requires equality after
averaging over a uniformly sampled key \(K\), but not for every fixed key.

\subsection{Hashing and Pseudorandom Functions}
\begin{appendixdefinition}[Hash function]
A hash function is a mapping \(h\colon\mathcal{Z}\to\mathcal{C}\), where
\(\mathcal{C}\) is a finite code space. Two representations \(z,z'\in
\mathcal{Z}\) \emph{collide} under \(h\) if \(h(z)=h(z')\).
\end{appendixdefinition}

\begin{appendixdefinition}[Locality-sensitive hash family]
Let \((\mathcal{Z},d)\) be a metric space. A distribution \(\mathcal{H}\)
over hash functions \(h\colon\mathcal{Z}\to\mathcal{C}\) is
\((r,cr,p_1,p_2)\)-locality-sensitive if \(c>1\), \(p_1>p_2\), and for every
\(z,z'\in\mathcal{Z}\),
\[
    d(z,z')\leq r \Longrightarrow
    \Pr_{h\sim\mathcal{H}}[h(z)=h(z')]\geq p_1,
\]
whereas
\[
    d(z,z')\geq cr \Longrightarrow
    \Pr_{h\sim\mathcal{H}}[h(z)=h(z')]\leq p_2.
\]
\end{appendixdefinition}

We use the standard notion of locality-sensitive hashing
\citep{indyk1998approximate}. In our method, \(\mathcal{H}\) is instantiated
with the \(M\)-bit random-hyperplane LSH construction for angular distance,
equivalently cosine similarity, following~\citet{charikar2002similarity}.

\begin{appendixdefinition}[Random-hyperplane locality-sensitive hash]
Let \(\mathcal{R}=(\bar r_1,\ldots,\bar r_M)\) consist of \(M\) unit
hyperplane normals sampled independently and uniformly from the unit sphere. For a normalized representation \(z\in\mathbb{R}^d\),
the random-hyperplane LSH is the code
\[
    H_{\mathcal{R}}(z)
    =
    \bigl(
        \mathbf{1}\{\langle \bar r_1,z\rangle\geq 0\},
        \ldots,
        \mathbf{1}\{\langle \bar r_M,z\rangle\geq 0\}
    \bigr)
    \in\{0,1\}^M.
\]
Thus, each bit indicates on which side of its corresponding hyperplane the
representation lies.
\end{appendixdefinition}

\begin{appendixdefinition}[Hamming distance and Hamming ball]
Following~\citet{hamming1950error}, the Hamming distance between two codes
\(c,c'\in\{0,1\}^M\) is the number of bit positions on which they differ:
\[
    d_{\mathrm H}(c,c')
    =
    \sum_{j=1}^{M}\mathbf{1}\{c_j\neq c'_j\}.
\]
For an integer \(r\in\{0,\ldots,M\}\), the radius-\(r\) Hamming ball centered
at \(c\) is
\[
    \mathcal{B}_{\mathrm H}(c,r)
    =
    \bigl\{c'\in\{0,1\}^M:d_{\mathrm H}(c,c')\leq r\bigr\}.
\]
\end{appendixdefinition}

\begin{appendixdefinition}[Boundary-restricted Hamming neighborhood]
For a normalized representation \(z\) and boundary threshold \(\delta\), let
\[
    J_\delta(z)
    =
    \bigl\{j\in[M]:|\langle\bar r_j,z\rangle|\leq\delta\bigr\}
\]
be the set of hash bits whose corresponding hyperplanes lie within the
boundary margin. Let \(\operatorname{Flip}(c,S)\) denote the code obtained by
flipping the bits of \(c\) indexed by \(S\). For an integer
\(r\in\{0,\ldots,M\}\), define
\[
    \mathcal N_{\delta,r}(z)
    =
    \left\{
        \operatorname{Flip}(H_{\mathcal R}(z),S):
        S\subseteq J_\delta(z),\ |S|\leq r
    \right\}.
\]
Every code in \(\mathcal N_{\delta,r}(z)\) lies within Hamming distance \(r\)
of \(H_{\mathcal R}(z)\). Unlike the complete Hamming ball, however, this
neighborhood permits changes only to bits associated with nearby hyperplane
boundaries.
\end{appendixdefinition}

We use a pseudorandom function (PRF) to assign a secret score to each hash
code, following~\citet{goldreich1986construct}.

\begin{appendixdefinition}[Keyed scoring function]
Let \(\mathcal K\) be the key space and let \(\mathcal C=\{0,1\}^M\) be the
code space. A keyed scoring family is a collection of functions
\[
    \{F_\kappa:\mathcal C\to[0,1]\}_{\kappa\in\mathcal K}.
\]
For a secret key \(\kappa\), the value \(F_\kappa(c)\) is the normalized secret
score assigned to code \(c\).
\end{appendixdefinition}

\begin{appendixassumption}[Ideal-PRF scoring model]
\label{ass:ideal-prf}
Let \(K\sim\operatorname{Unif}(\mathcal K)\). For every collection of distinct
codes \(c_1,\ldots,c_L\), the scores
\[
    F_K(c_1),\ldots,F_K(c_L)
\]
are independent and uniformly distributed over \([0,1]\). Equivalently, in
the theoretical analysis, \(F_K\) is modeled as a uniformly random function
from \(\mathcal C\) to \([0,1]\). In particular, distinct codes receive
independent scores, and score ties occur with probability zero.
\end{appendixassumption}

A practical implementation can obtain normalized scores from the finite
bit-string output of a cryptographic PRF. The exact equalities in our
theoretical results refer to the ideal model above; a secure PRF provides the
corresponding computational interpretation.

\section{Algorithmic Specification}
\label{app:algorithmic_specification}

In Section~\ref{sec:method}, we introduced our watermarking algorithm using
Gumbel selection (Figure~\ref{fig:method}). Here, we first introduce a simpler
sampling rule to explain how we preserve distortion-freeness when candidates
share semantic codes. We then explain why we replace this rule with Gumbel
selection in our final algorithm. Finally, we describe the boundary-aware
detector we use in our experiments.

When all candidates have distinct semantic codes, we can simply return the
image with the highest keyed score. Under the ideal-PRF assumption
(Assumption~\ref{ass:ideal-prf}), each candidate has probability \(1/N\) of
receiving the highest score after averaging over the secret key. However,
when multiple candidates share a code, they also share its score. Selecting
the highest-scoring code and then choosing uniformly among its candidates
would favor candidates belonging to smaller groups.

We first address this problem with a simpler construction inspired by
bundle sampling~\citep{vaikuntanathan2026undetectableconversations}, which
we call \emph{representative sampling}. We choose one representative
uniformly from each group of candidates sharing a code and let the
representatives compete according to their keyed scores. We then compensate
for the remaining candidates through an additional uniform-selection branch.

More precisely, let \(I_c=\{i\in[N]:c_i=c\}\) denote the group of candidates
sharing code \(c\). We choose one representative from each group to form
\(U\), and let \(D=[N]\setminus U\). With probability \(|U|/N\), we return
the highest-scoring representative. Otherwise, we select uniformly from
\(D\). Algorithm~\ref{alg:sampling} specifies the complete procedure. In
Appendix~\ref{app:distortion_free_proof}, we prove that this rule selects
every candidate with probability \(1/N\) after averaging over the secret key
and sampler randomness.

\begin{algorithm}[H]
\caption{Representative sampling}
\label{alg:sampling}
\begin{algorithmic}[1]
\Statex \textbf{Input:} Prompt \(p\) and number of candidates \(N\)
\Statex \textbf{Parameters:} \(G\), \(E\), and secret key \(\kappa\)
\State Sample independent seeds \(S_1,\ldots,S_N\)
\hfill\textit{Candidate seeds}
\For{\(i=1,\ldots,N\)}
    \State \(x_i\gets G(p,S_i)\) \hfill\textit{Generate candidate}
    \State \(c_i\gets H_{\mathcal{R}}(z(x_i))\)
    \hfill\textit{Assign its LSH code}
\EndFor
\State \(\mathcal C_N\gets\{c_i:i\in[N]\}\) \hfill\textit{Occupied codes}
\State \(U\gets\emptyset\)
\ForAll{\(c\in\mathcal C_N\)}
    \State \(I_c\gets\{i\in[N]:c_i=c\}\)
    \hfill\textit{Candidates sharing code \(c\)}
    \State Sample \(R_c\sim\operatorname{Uniform}(I_c)\)
    \hfill\textit{Choose one representative}
    \State \(U\gets U\cup\{R_c\}\)
\EndFor
\State \(D\gets[N]\setminus U\) \hfill\textit{Remaining candidates}
\State Sample \(Z\sim\operatorname{Bernoulli}(|U|/N)\)
\hfill\textit{Preserve uniform marginal selection}
\If{\(Z=1\)}
    \State \(A\gets\arg\max_{i\in U}F_\kappa(c_i)\)
    \hfill\textit{Highest keyed score}
    \State \(i^\star\sim\operatorname{Uniform}(A)\)
    \hfill\textit{Break ties uniformly}
\Else
    \State \(i^\star\sim\operatorname{Uniform}(D)\)
    \hfill\textit{Select a remaining candidate}
\EndIf
\State \Return \(x_{i^\star}\)
\end{algorithmic}
\end{algorithm}

Representative sampling provides an intuitive distortion-free construction,
but its uniform-selection branch ignores the keyed scores of the remaining
candidates. We therefore replace it with Gumbel
selection~\citep{kuditipudi2024robust}, which accounts for code multiplicities
while using keyed scores throughout code selection. Specifically, we select
the code maximizing \(\log F_\kappa(c)/n_c\), where \(n_c=|I_c|\), and then
sample uniformly from its candidates, as specified in
Algorithm~\ref{alg:gumbel_sampling}. Both rules coincide when all candidates
have distinct codes. In Appendix~\ref{app:alternative_sampling_rules}, we
compare the two rules and tournament sampling on identical candidate pools
and show that Gumbel selection achieves stronger detection performance. We
therefore use Gumbel selection in our final algorithm and experiments. In
Appendix~\ref{app:distortion_free_proof}, we prove that both sampling rules
preserve the original generator distribution.

\paragraph{Boundary-aware detection.}
We now describe how we improve robustness to content-preserving
transformations. Such transformations may move an image's semantic
representation across a nearby hyperplane and change its hash code, even
when the image retains its semantic content. To address this problem, we
identify hash bits whose representations lie close to their corresponding
hyperplane boundaries. We then evaluate nearby codes obtained by flipping
up to \(r_{\mathrm{ham}}\) of these bits and use the highest keyed score for
detection.

When \(r_{\mathrm{ham}}=0\), we recover the exact-code detector from
Algorithm~\ref{alg:exact_code_detection}. In all our experiments, including
clean-image detection, we use \(r_{\mathrm{ham}}=1\) and calibrate detection
thresholds on clean images. By restricting the search to near-boundary bits,
we avoid evaluating the complete Hamming ball. We formally define the
boundary-restricted neighborhood in Appendix~\ref{app:formal_preliminaries}
and specify the complete detection procedure for both individual images and
batches in Algorithm~\ref{alg:detection}.

\begin{algorithm}[H]
\caption{Boundary-aware detection from one or more images}
\label{alg:detection}
\label{alg:score}
\begin{algorithmic}[1]
\Statex \textbf{Input:} Images \(x_1,\ldots,x_B\), where \(B\geq1\)
\Statex \textbf{Parameters:} Encoder \(E\), boundary threshold \(\delta\),
Hamming radius \(r_{\mathrm{ham}}\), secret key \(\kappa\), and detection
threshold \(\tau\)
\Statex \textbf{Notation:} \(\operatorname{Flip}(c,S)\) denotes code \(c\)
with the bits indexed by \(S\) flipped.
\For{\(b=1,\ldots,B\)}
    \State \(z_b\gets E(x_b)/\|E(x_b)\|_2\)
    \State \(c_b\gets H_{\mathcal R}(z_b)\)
    \State \(J_\delta(z_b)\gets
        \{j\in[M]:|\langle\bar r_j,z_b\rangle|\leq\delta\}\)
    \State \(\mathcal N_{\delta,r_{\mathrm{ham}}}(z_b)\gets
        \{\operatorname{Flip}(c_b,S):S\subseteq J_\delta(z_b),
        \ |S|\leq r_{\mathrm{ham}}\}\)
    \State \(s_b\gets\max_{c'\in\mathcal N_{\delta,r_{\mathrm{ham}}}(z_b)}
        F_\kappa(c')\)
\EndFor
\State \Return \(\mathbf 1\left\{\frac{1}{B}\sum_{b=1}^{B}s_b\geq\tau\right\}\)
\end{algorithmic}
\end{algorithm}

We use Algorithm~\ref{alg:detection} for both single-image
and batch detection in our experiments. For \(B=1\), we evaluate an individual
image; for larger batches, we average the boundary-aware scores across
independently generated outputs. In Appendix~\ref{app:batch_detection}, we
prove that batch detection reduces both error probabilities exponentially
with batch size when the expected clean and watermarked scores lie on
opposite sides of the threshold for a fixed key.

\section{Proof of Distortion-Free Generation}
\label{app:distortion_free_proof}

We now prove the distortion-free guarantee stated in Theorem~\ref{thm:distortion-free}. Our goal is to show that our watermarking algorithm preserves the original generator's distribution after averaging over the secret key, regardless of the generator or semantic encoder. We first prove that representative sampling (Algorithm~\ref{alg:sampling}, Appendix~\ref{app:algorithmic_specification}) and Gumbel selection (Algorithm~\ref{alg:gumbel_sampling}, Section~\ref{sec:method}) both select every candidate with probability \(1/N\), even when candidates share semantic codes. We then show that this uniform-selection property guarantees distortion-free generation.

\subsection{Conditional Uniform Candidate Selection}
\label{app:conditional_uniform_selection}

We begin with representative sampling (Algorithm~\ref{alg:sampling}), which handles candidates sharing semantic codes by choosing one representative from each group and compensating for the remaining candidates through a uniform draw.

\phantomsection
\makeatletter
\begingroup
\edef\@currentlabel{\arabic{uniformselectionnumber}}
\label{lem:uniform-selection}
\endgroup
\makeatother
\resultstatementheading{Lemma~\ref*{lem:uniform-selection}
(Conditional uniform candidate selection).}

Let \(X_1,\ldots,X_N\overset{\mathrm{i.i.d.}}{\sim}P_G(\cdot\mid p)\), and let \(I^\star\in[N]\) be the index selected by Algorithm~\ref{alg:sampling}. Under the ideal-PRF assumption (Assumption~\ref{ass:ideal-prf}), for every fixed candidate batch \(x_{1:N}\) and every \(i\in[N]\),

\[
\mathbb E_{K\sim\mathrm{Unif}(\mathcal K)}
\left[
\Pr(I^\star=i\mid X_{1:N}=x_{1:N},K)
\right]
=\frac1N.
\]

\begin{proof}
Fix a candidate batch \(x_{1:N}\) and its codes \(c_{1:N}\). Let \(L=|U|\) denote the number of occupied codes, and let \(n_i=|I_{c_i}|\) denote the number of candidates sharing candidate \(i\)'s code.

Candidate \(i\) becomes the representative of its group with probability \(1/n_i\). We choose the representative branch with probability \(L/N\). Under the ideal-PRF assumption, the \(L\) distinct occupied codes receive independent, uniformly distributed scores, so each code wins with probability \(1/L\). The representative branch therefore selects candidate \(i\) with probability

\[
\frac{1}{n_i}\cdot\frac{L}{N}\cdot\frac{1}{L}
=\frac{1}{Nn_i}.
\]

If \(L<N\), candidate \(i\) belongs to \(D\) with probability \((n_i-1)/n_i\). We choose the complementary branch with probability \((N-L)/N\) and select uniformly among the \(N-L\) remaining candidates. This branch selects candidate \(i\) with probability

\[
\frac{n_i-1}{n_i}\cdot
\frac{N-L}{N}\cdot\frac{1}{N-L}
=\frac{n_i-1}{Nn_i}.
\]

Adding both contributions, we obtain

\[
\Pr(I^\star=i\mid X_{1:N}=x_{1:N})
=
\frac{1}{Nn_i}+
\frac{n_i-1}{Nn_i}
=\frac1N,
\]

where we average over the secret key and sampler randomness.

If \(L=N\), every candidate has a distinct code, so the representative branch alone selects each candidate with probability \(1/N\).
\end{proof}

We next establish the same uniform-selection property for Gumbel selection (Algorithm~\ref{alg:gumbel_sampling}), which we use in our final watermarking algorithm. Unlike representative sampling, Gumbel selection accounts for code multiplicities while using keyed scores throughout code selection.

Fix a candidate batch and let \(I_c=\{i:c_i=c\}\), with \(n_c=|I_c|\). Under the ideal-PRF assumption, the scores \(F_K(c)\) are independent and uniformly distributed over \([0,1]\) for distinct codes. Consequently, the random variables

\[
T_c=\frac{-\log F_K(c)}{n_c}
\]

are independent exponential random variables with rates \(n_c\). By the exponential-minimum sampling identity~\citep{kuditipudi2024robust}, the code minimizing \(T_c\), equivalently maximizing \(\log F_K(c)/n_c\), wins with probability

\[
\Pr(c^\star=c)=
\frac{n_c}{\sum_{c'\in \mathcal C_N}n_{c'}}
=\frac{n_c}{N}.
\]

We then select uniformly among the \(n_c\) candidates sharing the winning code. Therefore, each candidate \(i\) has selection probability

\[
\Pr(I^\star=i\mid X_{1:N}=x_{1:N})
=
\frac{n_{c_i}}N\cdot\frac1{n_{c_i}}
=\frac1N,
\]

where we again average over the secret key and sampler randomness.

We have thus established conditional uniform candidate selection for both representative sampling and Gumbel selection, even when multiple candidates share the same semantic code.

\subsection{Distortion-Free Generation}

We now prove Theorem~\ref{thm:distortion-free} using the conditional uniform-selection property established in Section~\ref{app:conditional_uniform_selection}. We first apply it to Gumbel selection (Algorithm~\ref{alg:gumbel_sampling}) and then conclude that the same guarantee holds for representative sampling.

\noindent\begin{minipage}{\linewidth}
\resultstatementheading{Theorem~\ref{thm:distortion-free} (Distortion-free generation).}

Let \(G\) be any conditional generative model and \(E\) any encoder for
which Algorithm~\ref{alg:gumbel_sampling} is well-defined. Under the ideal-PRF
assumption (Assumption~\ref{ass:ideal-prf}), the watermarked generator defined
by Algorithm~\ref{alg:gumbel_sampling} is distortion-free
(Definition~\ref{def:distortion-free}): for every prompt \(p\),
\[
\mathbb E_{K\sim\operatorname{Unif}(\mathcal K)}
\left[P_{G_K^{\mathrm{wm}}}(\cdot\mid p)\right]
=P_G(\cdot\mid p).
\]
\end{minipage}

\begin{proof}
Fix a prompt \(p\) and draw \(X_1,\ldots,X_N\) independently from \(P_G(\cdot\mid p)\). Let \(I^\star\) be the index selected by Algorithm~\ref{alg:gumbel_sampling}, and set \(X^{\mathrm{wm}}=X_{I^\star}\). Fix any measurable set of outputs \(A\). By conditioning on the candidate batch and applying the uniform-selection property of Algorithm~\ref{alg:gumbel_sampling} proved in Section~\ref{app:conditional_uniform_selection}, we obtain

\[
\begin{aligned}
\mathbb E_{K\sim\operatorname{Unif}(\mathcal K)}
\left[P_{G_K^{\mathrm{wm}}}(A\mid p)\right]
&=
\Pr(X^{\mathrm{wm}}\in A\mid p)\\
&=
\mathbb E_{X_{1:N}}\left[
\sum_{i=1}^N
\mathbf 1\{X_i\in A\}
\Pr(I^\star=i\mid X_{1:N})
\right]\\
&=
\mathbb E_{X_{1:N}}\left[
\frac1N\sum_{i=1}^N\mathbf 1\{X_i\in A\}
\right]\\
&=
\frac1N\sum_{i=1}^N\Pr(X_i\in A\mid p)\\
&=P_G(A\mid p).
\end{aligned}
\]

Therefore, the watermarked output follows the original generator distribution after averaging over the secret key.
\end{proof}

We also prove distortion-freeness for representative sampling (Algorithm~\ref{alg:sampling}) by applying the same argument to its uniform-selection property established in Section~\ref{app:conditional_uniform_selection}.

\section{High Semantic Entropy and Detection}
\label{app:semantic_entropy_tpr}

We want to understand when high semantic entropy leads to strong detection
in our construction. Detection benefits from candidates occupying distinct
codes, since each occupied code supplies another keyed score. We therefore
use semantic entropy to bound code agreement, turn this into a lower bound on
expected code occupancy, and then derive the detection guarantee.

\paragraph{Detection rates.}
We evaluate detection for a fixed prompt \(p\), declaring an output watermarked
when its exact-code score is at least \(\tau\). The false-positive rate
\(\operatorname{FPR}(\tau)\) measures how often an independent unwatermarked
sample \(X\sim P_G(\cdot\mid p)\) is declared watermarked. The true-positive
rate \(\operatorname{TPR}^{(N)}(\tau)\) measures how often the detector correctly
identifies a watermarked output from Algorithm~\ref{alg:gumbel_sampling}, using
\(N\) candidates for the same prompt. We average both rates over the secret key, random hyperplanes, and
generation and sampler randomness, using the same key and hyperplanes for
watermarked generation and detection. Because the unwatermarked sample is
independent of the key, its score is uniform under the ideal-PRF assumption.
Therefore,
\[
\operatorname{FPR}(\tau)=\Pr(U\geq\tau)=1-\tau,
\qquad U\sim\operatorname{Unif}([0,1]).
\]
The boundary-aware detector used in our experiments also checks nearby codes
and therefore requires separate false-positive calibration. These averaged
rates do not give a guarantee for every fixed key or hyperplane draw.

\begin{resultlemma}[High semantic entropy limits hash agreement]
\label{lem:semantic-entropy-hash-agreement}
Fix a prompt \(p\) and a semantic radius \(\varepsilon\in[0,2]\). For two
independent generations and an \(M\)-bit random-hyperplane LSH, the
probability that their codes agree is at most
\[
q_{M,\varepsilon}(p)
= e^{-H_{\mathrm{sem}}^{E,\varepsilon}(G\mid p)}
+\left(1-e^{-H_{\mathrm{sem}}^{E,\varepsilon}(G\mid p)}\right)
\left(1-\frac{\arccos(1-\varepsilon)}{\pi}\right)^M.
\]
\end{resultlemma}

\begin{proof}
Let \(A=\{d_E(X,X')\leq\varepsilon\}\), and let
\(B=\{H_{\mathcal R}(z(X))=H_{\mathcal R}(z(X'))\}\).
The probability is over two independent generations and the independent random
hyperplanes. By Definition~\ref{def:semantic_entropy},
\[
\Pr(A)=e^{-H_{\mathrm{sem}}^{E,\varepsilon}(G\mid p)},
\qquad
\Pr(A^c)=1-e^{-H_{\mathrm{sem}}^{E,\varepsilon}(G\mid p)}.
\]
For a fixed pair at distance \(d\), the random-hyperplane identity gives
\[
\Pr_{\mathcal R}(B\mid X,X')=
\left(1-\frac{\arccos(1-d)}{\pi}\right)^M
\quad\text{where }d=d_E(X,X').
\]
This expression decreases with \(d\). Thus \(\Pr(B\mid A)\leq1\) and
\(\Pr(B\mid A^c)\leq(1-\arccos(1-\varepsilon)/\pi)^M\).
The law of total probability now gives
\[
\begin{aligned}
\Pr(B)
&=\Pr(B\cap A)+\Pr(B\cap A^c)\\
&=\Pr(B\mid A)\Pr(A)+\Pr(B\mid A^c)\Pr(A^c)\\
&\leq\Pr(A)+\left(1-\frac{\arccos(1-\varepsilon)}{\pi}\right)^M\Pr(A^c)\\
&=e^{-H_{\mathrm{sem}}^{E,\varepsilon}(G\mid p)}
+\left(1-e^{-H_{\mathrm{sem}}^{E,\varepsilon}(G\mid p)}\right)
\left(1-\frac{\arccos(1-\varepsilon)}{\pi}\right)^M\\
&=q_{M,\varepsilon}(p).
\end{aligned}
\]
If a conditioning event has probability zero, its term is omitted.
\end{proof}

\begin{resultlemma}[Expected number of occupied codes]
\label{lem:expected-occupied-codes}
Fix a prompt \(p\), a semantic radius \(\varepsilon\in[0,2]\), and
\(N\geq2\). Let \(X_1,\ldots,X_N\) be independent generations from
\(P_G(\cdot\mid p)\), and let \(L\) be the number of distinct codes
\(H_{\mathcal R}(z(X_i))\) under an \(M\)-bit random-hyperplane LSH. Then
\[
\mathbb E[L]\geq\frac{N}{1+(N-1)q_{M,\varepsilon}(p)},
\]
where \(q_{M,\varepsilon}(p)\) is defined in
Lemma~\ref{lem:semantic-entropy-hash-agreement}. The expectation is over the
candidate generations and random hyperplanes.
\end{resultlemma}

\begin{proof}
Write \(q=q_{M,\varepsilon}(p)\) and \(C_i=H_{\mathcal R}(z(X_i))\), and
let \(J\) count pairs of
candidates whose codes agree. If the occupied-code multiplicities are
\(n_1,\ldots,n_L\), then
\[
J=\sum_{i<j}\mathbf 1\{C_i=C_j\}
=\sum_{j=1}^{L}\binom{n_j}{2},
\qquad
\sum_{j=1}^{L}n_j^2=N+2J.
\]
Lemma~\ref{lem:semantic-entropy-hash-agreement} and linearity of expectation give
\[
\mathbb E[J]=\sum_{i<j}\Pr(C_i=C_j)\leq\binom N2q.
\]
By Cauchy--Schwarz,
\[
N^2=\left(\sum_{j=1}^{L}n_j\right)^2
\leq L\sum_{j=1}^{L}n_j^2=L(N+2J).
\]
Taking expectations and applying Jensen's inequality to the reciprocal,
\[
\begin{aligned}
\mathbb E[L]
&\geq\mathbb E\!\left[\frac{N^2}{N+2J}\right]
\geq\frac{N^2}{N+2\mathbb E[J]}\\
&\geq\frac{N}{1+(N-1)q}.
\end{aligned}
\]
\end{proof}

To turn code occupancy into a detection guarantee, we next compute the score
distribution of the code returned by Gumbel selection. We fix the candidate
codes so that only the key remains random.

\begin{resultlemma}[Gumbel selection over occupied codes]
\label{lem:gumbel-race}
Fix candidate codes \(c_{1:N}\), let \(\mathcal C_N=\{c_i:i\in[N]\}\),
and let \(n_c=|\{i:c_i=c\}|\). Draw
\(K\sim\operatorname{Unif}(\mathcal K)\). Under the ideal-PRF assumption
(Assumption~\ref{ass:ideal-prf}), let
\(c^\star\) be the code selected by Algorithm~\ref{alg:gumbel_sampling}.
Then for every \(c\in\mathcal C_N\) and \(\tau\in[0,1]\),
\[
\Pr(c^\star=c)=\frac{n_c}{N},
\qquad
\Pr\!\left(F_K(c^\star)\geq\tau\mid c^\star=c\right)
=1-\tau^{N/n_c}.
\]
The probabilities average over the secret key.
\end{resultlemma}

\begin{proof}
Write \(U_c=F_K(c)\). Under the ideal-PRF assumption these scores are
independent and uniform on \([0,1]\). Gumbel selection maximizes
\(\log U_c/n_c\). Thus, for a fixed score \(U_c=u\in(0,1)\), code \(c\)
wins exactly when every other occupied code \(d\) satisfies
\[
\frac{\log U_d}{n_d}\leq\frac{\log u}{n_c},
\qquad\text{or equivalently}\qquad U_d\leq u^{n_d/n_c}.
\]
Ties have probability zero. Independence and uniformity therefore give
\[
\Pr(c^\star=c\mid U_c=u)
=\prod_{d\neq c}u^{n_d/n_c}
=u^{(N-n_c)/n_c}.
\]
Since \(U_c\) has uniform density, integrating over scores at least \(\tau\)
gives
\[
\begin{aligned}
\Pr(c^\star=c,\,U_c\geq\tau)
&=\int_\tau^1 u^{(N-n_c)/n_c}\,du\\
&=\frac{n_c}{N}\left(1-\tau^{N/n_c}\right).
\end{aligned}
\]
At \(\tau=0\), this gives \(\Pr(c^\star=c)=n_c/N\). Dividing the joint
probability by \(n_c/N\) then gives
\(\Pr(U_c\geq\tau\mid c^\star=c)=1-\tau^{N/n_c}\), as claimed.
This is the occupied-code form of the exponential-minimum identity
in~\citet[Lemma C.1]{kuditipudi2024robust}.
\end{proof}

\noindent\begin{minipage}{\linewidth}
\resultstatementheading{Theorem~\ref{thm:semantic-entropy-tpr}
(Semantic entropy and detection).}
Fix a prompt \(p\) and a semantic radius \(\varepsilon\in[0,2]\). Generate
\(N\geq2\) independent candidates and apply Algorithm~\ref{alg:gumbel_sampling}. Under the ideal-PRF assumption,
for single-image detection, the exact-code detector
(Algorithm~\ref{alg:exact_code_detection}) at threshold \(\tau\in[0,1]\) satisfies
\[
\operatorname{FPR}(\tau)=1-\tau
\]
and
\[
\operatorname{TPR}^{(N)}(\tau)
\geq
1-\tau
+\frac{1-q_{M,\varepsilon}(p)}
{1+(N-1)q_{M,\varepsilon}(p)}
\left(\tau-\tau^N\right),
\]
where
\[
q_{M,\varepsilon}(p)
=e^{-H_{\mathrm{sem}}^{E,\varepsilon}(G\mid p)}
+\left(1-e^{-H_{\mathrm{sem}}^{E,\varepsilon}(G\mid p)}\right)
\left(1-\frac{\arccos(1-\varepsilon)}{\pi}\right)^M.
\]
Both rates average over the secret key and random hyperplanes for the fixed
prompt \(p\); the definitions are given above.
\end{minipage}\par

\begin{proof}
Write \(q=q_{M,\varepsilon}(p)\in[0,1]\), and let \(L\) be the number
of occupied codes among the \(N\) candidates. For each candidate \(i\), let
\(n_i=|I_{c_i}|\) be the number of candidates sharing its code.

\noindent Detection conditioned on code multiplicities.\quad
Condition on the candidates and hyperplanes, and average over the independent
key and sampler randomness. By Lemma~\ref{lem:gumbel-race},
\[
\begin{aligned}
\Pr(\mathsf{Detect}=1\mid c_{1:N})
&=\sum_{c\in\mathcal C_N}\frac{n_c}{N}
\left(1-\tau^{N/n_c}\right)\\
&=1-\tau+\frac1N\sum_{i=1}^{N}
\left(\tau-\tau^{N/n_i}\right).
\end{aligned}
\]
The second equality counts each occupied code once for each of its members.

\noindent Combining the bounds.\quad
For \(0<\tau<1\), the function \(s\mapsto\tau-\tau^s\) is concave and
vanishes at \(s=1\). Its chord between \(s=1\) and \(s=N\) gives
\[
\tau-\tau^{N/n_i}
\geq\frac{N/n_i-1}{N-1}\left(\tau-\tau^N\right).
\]
Each occupied code contributes \(N\) to \(\sum_{i=1}^{N}N/n_i\), so this sum
equals \(NL\). Therefore,
\[
\Pr(\mathsf{Detect}=1\mid c_{1:N})
\geq1-\tau+\frac{L-1}{N-1}\left(\tau-\tau^N\right).
\]
Taking expectations over the candidates and hyperplanes and applying
Lemma~\ref{lem:expected-occupied-codes},
\[
\begin{aligned}
\operatorname{TPR}^{(N)}(\tau)
&\geq1-\tau+\frac{\mathbb E[L]-1}{N-1}
\left(\tau-\tau^N\right)\\
&\geq1-\tau+\frac{1}{N-1}
\left(\frac{N}{1+(N-1)q}-1\right)
\left(\tau-\tau^N\right)\\
&=1-\tau+\frac{1-q}{1+(N-1)q}
\left(\tau-\tau^N\right).
\end{aligned}
\]
An independent unwatermarked output has a uniform exact-code score after
averaging over the key, giving \(\operatorname{FPR}(\tau)=1-\tau\).
The endpoint cases \(\tau=0\) and \(\tau=1\) follow directly.
\end{proof}

\section{Batch-Detection Concentration}
\label{app:batch_detection}

We now analyze how aggregating multiple independently generated outputs
improves watermark detection. In Algorithm~\ref{alg:exact_code_detection},
we introduced batch detection by averaging the keyed scores of multiple
images and comparing the result with a threshold. We show that this
aggregation reduces both false-positive and false-negative probabilities
exponentially with the batch size, provided the expected clean and
watermarked scores lie on opposite sides of the threshold.

Unlike Theorem~\ref{thm:semantic-entropy-tpr}, which averages over the secret
key and random hyperplanes, we analyze detection with a fixed deployed key
and hash. Our result applies to both the exact-code detector
(Algorithm~\ref{alg:exact_code_detection}) and the boundary-aware detector
(Algorithm~\ref{alg:detection}), since both aggregate per-image scores in
\([0,1]\).

\begin{resultlemma}[Batch-detection concentration]
\label{lem:batch-detection-concentration}
Fix a secret key \(\kappa\) and hash \(H_{\mathcal R}\), and let \(\mu_0\)
and \(\mu_{\mathrm{wm}}\) denote the expected per-image scores for
unwatermarked and watermarked generations, respectively. Suppose that
\[
\mu_0<\tau<\mu_{\mathrm{wm}},
\]
where \(\tau\) is the batch-detection threshold. For a batch of \(B\)
independently generated images with the same expected score within each
population, the false-positive and false-negative probabilities satisfy
\[
\Pr(\mathrm{FP})
\le
\exp\left(-2B(\tau-\mu_0)^2\right)
\]
and
\[
\Pr(\mathrm{FN})
\le
\exp\left(-2B(\mu_{\mathrm{wm}}-\tau)^2\right).
\]
Thus, both error probabilities decrease exponentially with \(B\).
\end{resultlemma}

\begin{proof}
We first consider a batch of \(B\) independently generated unwatermarked
images \(Y_1,\ldots,Y_B\). Let
\[
S_b^0=\mathsf{Score}_{\kappa,\mathcal R}(Y_b),
\qquad
\overline S_B^0=\frac1B\sum_{b=1}^{B}S_b^0.
\]
For the fixed key and hash, the scores are independent, lie in \([0,1]\),
and have expectation \(\mu_0\). A false positive occurs when
\(\overline S_B^0\ge\tau\). Applying Hoeffding's inequality, we obtain
\[
\begin{aligned}
\Pr(\mathrm{FP})
&=\Pr(\overline S_B^0-\mu_0\ge\tau-\mu_0)\\
&\le\exp\left(-2B(\tau-\mu_0)^2\right).
\end{aligned}
\]

We next consider \(B\) independently generated watermarked images
\(X_1^{\mathrm{wm}},\ldots,X_B^{\mathrm{wm}}\). Define
\[
S_b^{\mathrm{wm}}=\mathsf{Score}_{\kappa,\mathcal R}(X_b^{\mathrm{wm}}),
\qquad
\overline S_B^{\mathrm{wm}}=\frac1B\sum_{b=1}^{B}S_b^{\mathrm{wm}}.
\]
For the same fixed key and hash, these scores are independent, lie in
\([0,1]\), and have expectation \(\mu_{\mathrm{wm}}\). A false negative
occurs when \(\overline S_B^{\mathrm{wm}}<\tau\). Applying Hoeffding's
inequality again, we obtain
\[
\begin{aligned}
\Pr(\mathrm{FN})
&=\Pr(\overline S_B^{\mathrm{wm}}<\tau)\\
&\le\Pr(\mu_{\mathrm{wm}}-\overline S_B^{\mathrm{wm}}
\ge\mu_{\mathrm{wm}}-\tau)\\
&\le\exp\left(-2B(\mu_{\mathrm{wm}}-\tau)^2\right).
\end{aligned}
\]
Since \(\mu_0<\tau<\mu_{\mathrm{wm}}\), both exponents are strictly negative
and decrease linearly with \(B\). We therefore obtain exponential decay of
both detection error probabilities.
\end{proof}

We emphasize that our fixed-key mean-separation assumption is necessary for
this result. The key-averaged guarantee in
Theorem~\ref{thm:semantic-entropy-tpr} does not imply that the expected
scores are separated for every deployed key and hash. Rather, we show that
whenever this separation holds, aggregating more independently generated
outputs exponentially improves detection reliability.

\section{Robustness Under Semantic Similarity}
\label{app:semantic_robustness}

We now analyze how transformations that preserve semantic similarity affect
our watermark. An edit may change many pixels while leaving the image's
semantic representation nearly unchanged. Since we assign watermark scores
to LSH codes, we want to understand when such transformations
preserve the original code and consequently its keyed score.

We show that the probability of preserving the hash code increases with the
cosine similarity between the original and transformed representations.
This provides a theoretical explanation for why our watermark can remain
detectable after content-preserving transformations.

\begin{resultlemma}[Hash-code preservation under semantic similarity]
\label{lem:semantic-score-preservation}
Fix two images \(x\) and \(x'\) with nonzero encoder representations, and let
\[
\rho=\langle z(x),z(x')\rangle.
\]
Let \(H_{\mathcal R}\) consist of \(M\) independently sampled uniform random
hyperplanes, and fix a secret key \(\kappa\). Then
\[
\Pr_{\mathcal R}\!\left[H_{\mathcal R}(z(x'))=H_{\mathcal R}(z(x))\right]
=
\left(1-\frac{\arccos(\rho)}{\pi}\right)^M.
\]
Whenever the codes coincide, the images receive identical exact-code keyed
scores. In particular, if the exact-code detector identifies \(x\) as
watermarked, our boundary-aware detector also identifies \(x'\) as
watermarked on this event.
\end{resultlemma}

\begin{proof}
For a single random hyperplane, the probability that \(z(x)\) and \(z(x')\)
receive the same bit is
\[
1-\frac{\arccos(\rho)}{\pi}.
\]
Since we sample the \(M\) hyperplanes independently, the probability that all
hash bits agree is
\[
\left(1-\frac{\arccos(\rho)}{\pi}\right)^M.
\]
When the codes agree, their keyed scores also agree:
\[
F_\kappa(H_{\mathcal R}(z(x')))=F_\kappa(H_{\mathcal R}(z(x))).
\]
Our boundary-aware detector always includes the observed hash code among the
codes it evaluates. Therefore, if \(F_\kappa(H_{\mathcal R}(z(x)))\ge\tau\),
then
\[
\mathsf{Score}_{\kappa,\mathcal R}(x')\ge F_\kappa(H_{\mathcal R}(z(x')))\ge\tau,
\]
and the boundary-aware detector identifies \(x'\) as watermarked.
\end{proof}

We thus establish a code-preservation guarantee that depends on semantic
similarity. The probability averages over random hyperplanes; we do not
claim the same guarantee for every fixed hash or against adaptive
transformations.

\section{Semantic Encoder Training}
\label{app:semantic_encoder_training}

\begin{figure}[H]
    \centering
    \includegraphics[width=\textwidth]{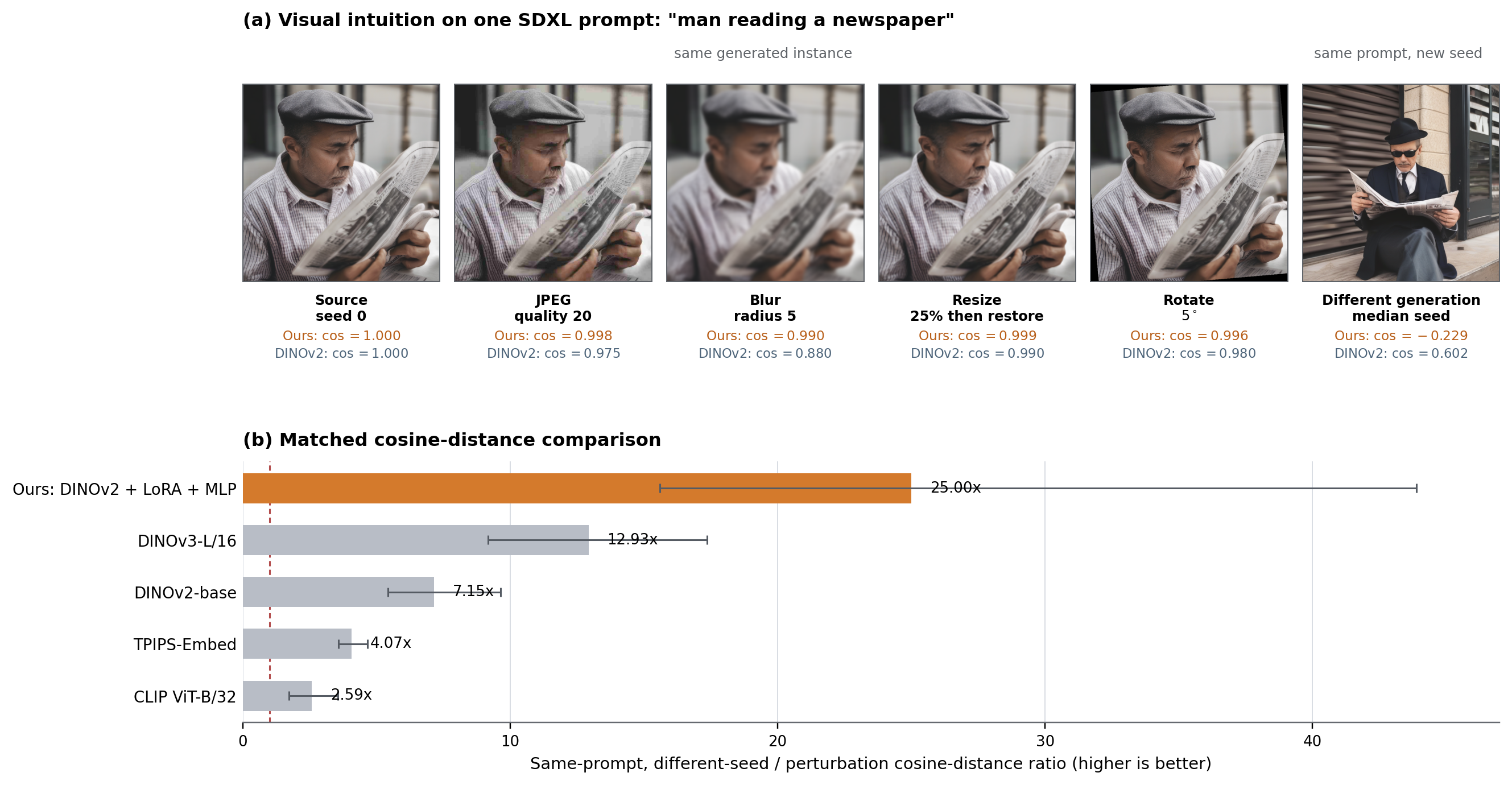}
    \caption{\textbf{The two requirements of a semantic encoder.}
    (a) An SDXL image, content-preserving transformations of that image, and an independent generation from the same prompt.
    (b) Mean cosine distance between different-seed generations divided by mean cosine distance under transformations. Higher values indicate stronger separation relative to transformation sensitivity.}
    \label{fig:semantic_encoder}
\end{figure}

\begin{figure}[H]
    \centering
    \includegraphics[width=\textwidth]{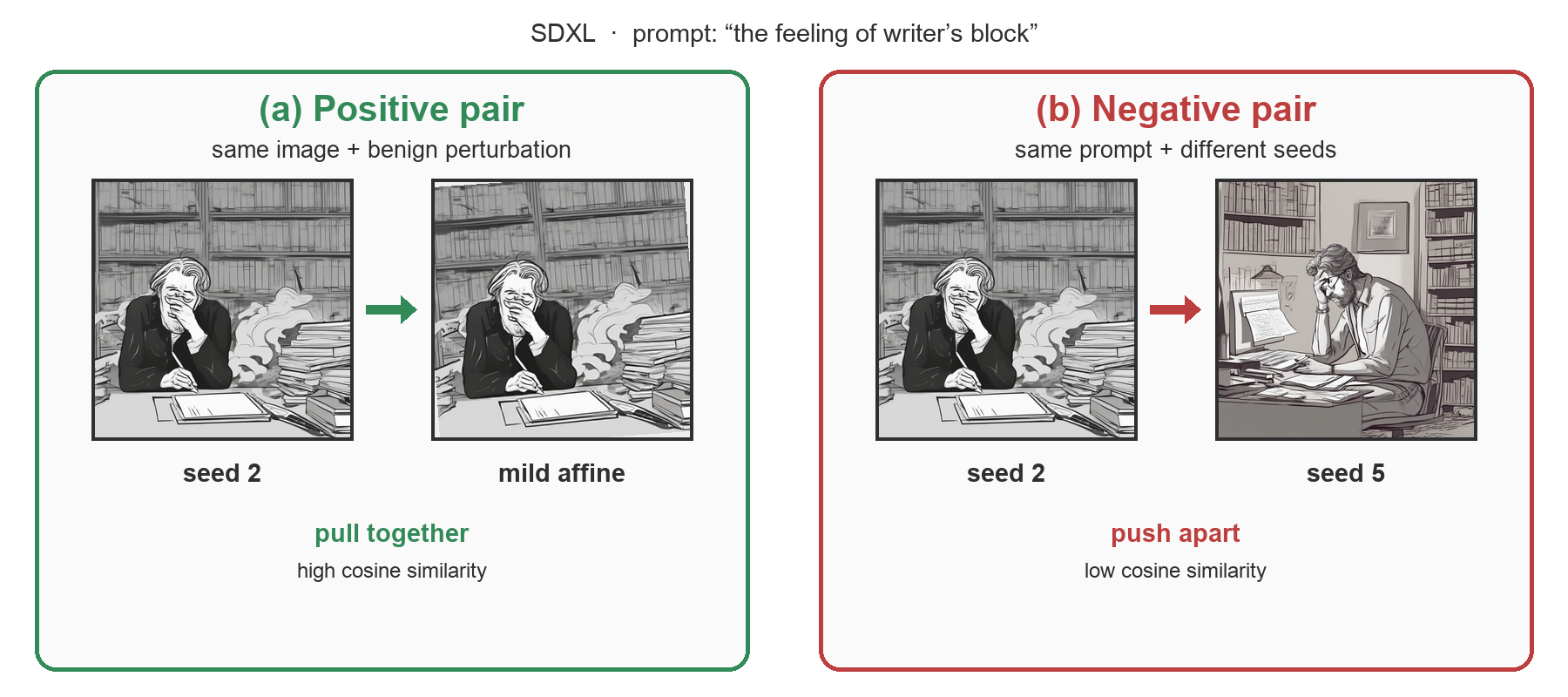}
    \caption{Training-pair construction using SDXL generations for the prompt \emph{the feeling of writer's block}. (a) An image and a
    semantics-preserving transformation form a positive pair and are pulled
    together. (b) Independent generations from the same prompt but different
    seeds \(2\) and \(5\) form a negative pair and are pushed apart.}
    \label{fig:semantic_encoder_training_pairs}
\end{figure}

We train the semantic encoder to satisfy two complementary requirements:
semantics-preserving transformations of the same image should remain close,
whereas independent generations from the same prompt should remain
distinguishable. Figure~\ref{fig:semantic_encoder} summarizes these
requirements, and Figure~\ref{fig:semantic_encoder_training_pairs} illustrates
the corresponding training pairs.

\paragraph{Architecture and objective.}
We initialize the encoder from DINOv2-base and freeze its pretrained
parameters. Rank-\(8\) LoRA adapters~\citep{hu2022lora} are applied to the query and value
projections of the last four transformer blocks. The CLS representation is
passed through a \(768\!\rightarrow\!512\!\rightarrow\!256\) MLP with GELU
activation and dropout \(0.1\), followed by \(\ell_2\) normalization.
For normalized embeddings \(z_i\) and \(z_j\), we use
\[
    \mathcal L_{+}=1-z_i^\top z_j,
    \qquad
    \mathcal L_{-}=\max\!\left(0,z_i^\top z_j\right).
\]
Positive pairs include Gaussian noise, Gaussian blur, JPEG compression,
resampling, photometric changes, and affine transformations. Negative pairs
contain generations obtained from different seeds of the same prompt.

\paragraph{Training details.}
We use \(100\) prompts evenly from DiffusionDB's 2M-text-only split and the
official COCO-2017 training captions. We divide them into \(80\) training and
\(20\) validation prompts; neither set is used in the reported evaluation
experiments. We use \(64\) seeds per prompt. Each optimizer update contains \(16\)
pairs of one type, with pair types balanced across updates. We train using
AdamW with learning rate \(10^{-4}\), mixed precision, gradient clipping at
\(1.0\), and learning-rate decay of \(0.95\) per epoch. We select the rank-\(8\)
LoRA model with the MLP head on the validation split. The checkpoint used in
Appendix~\ref{app:semantic_entropy} was selected after \(19{,}840\) optimizer
updates; the semantic-entropy results were not used for model selection.

\clearpage

\section{Encoder Evaluation: Details and Additional Results}
\label{app:semantic_entropy}

This appendix expands the semantic representation analysis in
Section~\ref{sec:encoder_analysis}. We provide the perturbation settings,
analyze code counts under both transformations and independent generation,
and conclude with a qualitative example. The main text gives the entropy
definition and encoder comparison.

\paragraph{Stability evaluation.}
We evaluate stability using 200 real source images and 16 semantics-preserving
transformations of each image and for every transformation family: JPEG compression,
Gaussian blur, resizing and resampling, and mild cropping with color adjustment. Within each
transformation family, severity increases gradually across the 16 variants:
JPEG quality ranges from \(35\) to \(95\), blur radius from \(0.25\) to
\(1.75\), resize scale from \(0.65\) to \(0.95\), and crop scale from
\(0.86\) to \(0.98\), with color factors from \(0.90\) to \(1.08\). Every
encoder is evaluated on the same source images and transformations. For each
family, Table~\ref{tab:encoder_perturbation_comparison} summarizes all 16
severity levels together: we compute mean pairwise cosine similarity across
the variants of each source image and then average the resulting per-image
values over the evaluation set.

\begin{table}[H]
\centering
\small
\setlength{\tabcolsep}{3pt}
\renewcommand{\arraystretch}{0.95}
\caption{\textbf{Encoder stability under content-preserving transformations.}
Mean pairwise cosine similarity across transformed versions of the same source
image. Higher values indicate greater invariance.}
\label{tab:encoder_perturbation_comparison}
\begin{tabular}{@{}l*{5}{r}@{}}
\toprule
 & DINOv2 & CLIP & DINOv3 & TPIPS & Ours \\
\midrule
JPEG compression & 0.974 & 0.959 & 0.976 & 0.956 & \textbf{0.985} \\
Gaussian blur & 0.922 & 0.932 & \textbf{0.974} & 0.895 & 0.963 \\
Resize--resample & 0.992 & 0.994 & 0.995 & 0.978 & \textbf{0.996} \\
Mild crop + color & 0.954 & 0.964 & \textbf{0.972} & 0.919 & 0.937 \\
\bottomrule
\end{tabular}

\end{table}

\begin{figure}[!htbp]
\centering
\includegraphics[width=\textwidth]{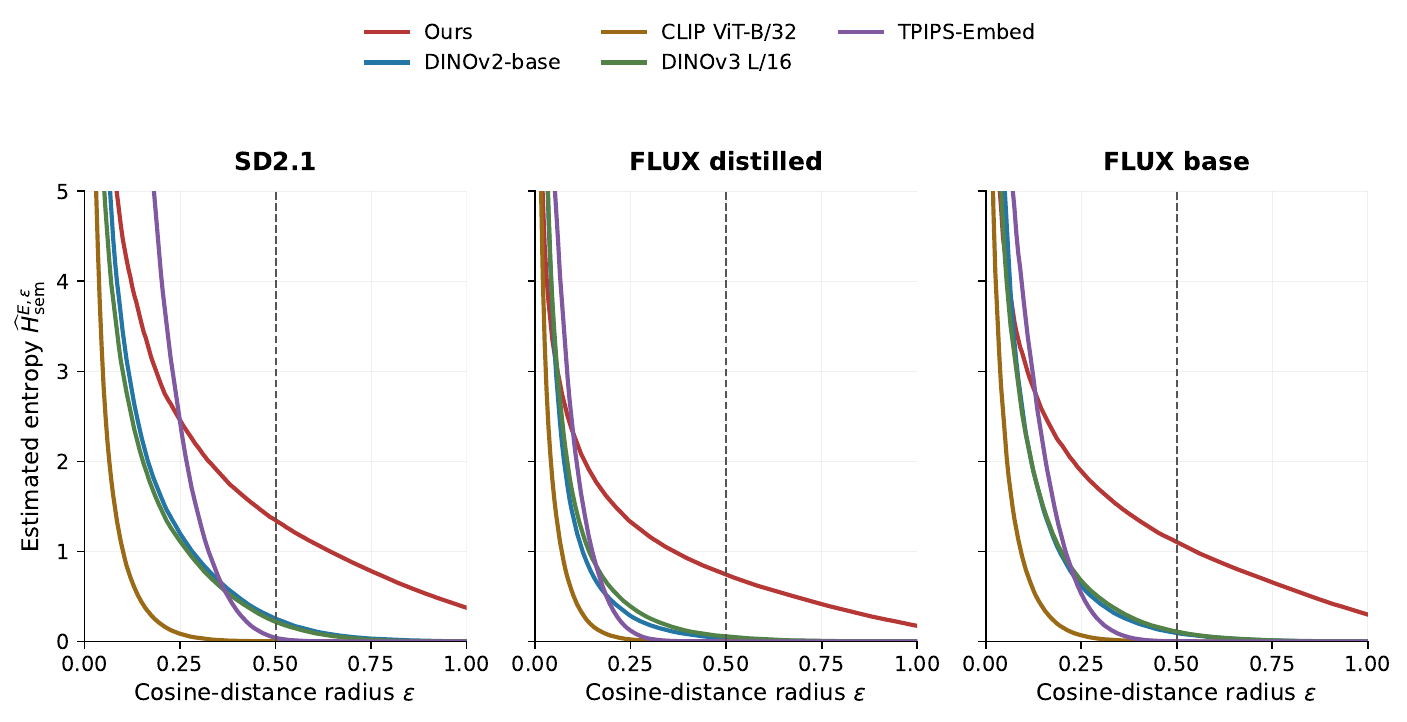}
\caption{\textbf{Semantic entropy as a function of \(\varepsilon\).}
Each panel reports \(\widehat H_{\mathrm{sem}}^{E,\varepsilon}\) for one
generator. Higher is better: at a fixed \(\varepsilon\), higher curves
indicate greater semantic diversity. Dashed lines mark the table radius,
$\varepsilon=0.5$. Panels show $\varepsilon\in[0,1]$ and entropy in $[0,5]$.}
\label{fig:semantic_entropy_epsilon}
\end{figure}

\paragraph{Sensitivity to \(\varepsilon\).}
Figure~\ref{fig:semantic_entropy_epsilon} reports semantic-entropy curves
using all \(\binom{16}{2}=120\) pairs per prompt and
200 held-out prompts per generator. For
Table~\ref{tab:encoder_comparison}, we use the same samples at
$\varepsilon=0.5$, counting pairs
with cosine similarity at least $0.5$ as close. Each reported entropy value
is computed from the overall fraction of close pairs across the 200 prompts.

\begin{figure}[h]
\centering
\includegraphics[width=\textwidth]{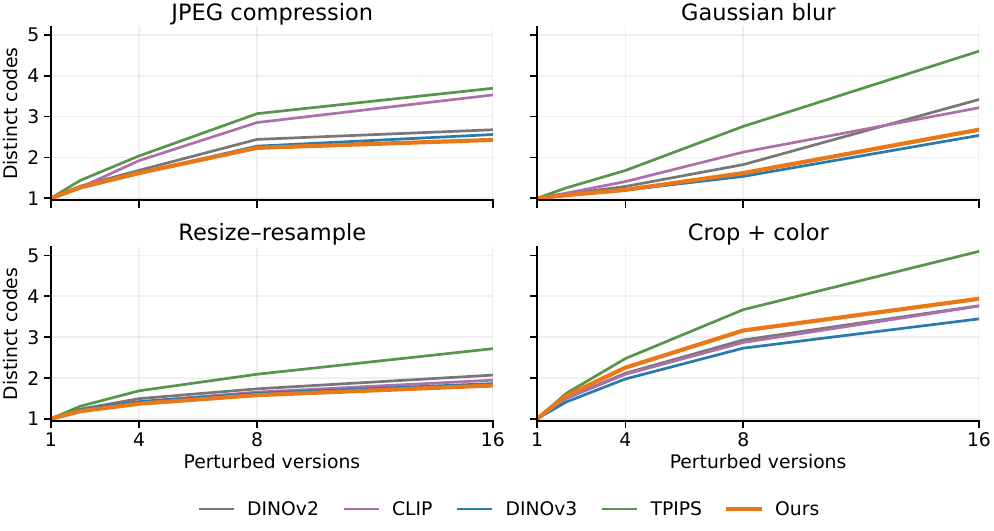}
\caption{\textbf{Distinct codes under semantics-preserving perturbations.}
Mean number of distinct 8-bit codes as additional perturbed versions are
included. Each panel averages 200 source-image groups over 200 random
hyperplane draws. Lower counts are desirable. Separate panels highlight the
different behavior of the four perturbation types.}
\label{fig:unique_codes_perturbations}
\end{figure}

\paragraph{Distinct codes under perturbations.}
We next examine whether semantics-preserving transformations retain the same
discrete hash code. For each transformation type, the analysis uses 200
source images and 16 perturbed variants of each image. We apply 8-bit
random-hyperplane hashing to the corresponding embeddings and average over
200 projections, using each projection consistently across all images and
encoders. Figure~\ref{fig:unique_codes_perturbations} reports the resulting
code counts. Fewer distinct codes indicate that transformed versions of the
same image more often retain its original code. This discrete analysis
complements the continuous results above.

\begin{figure}[h]
\centering
\includegraphics[width=\textwidth]{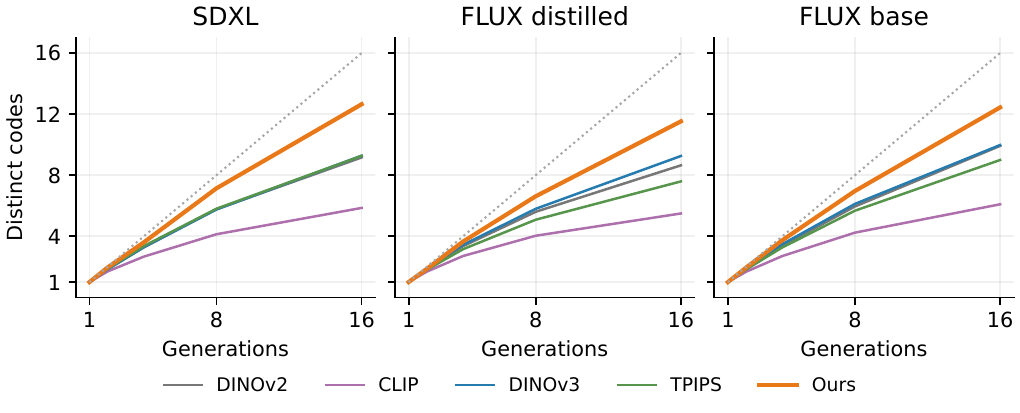}
\caption{\textbf{Distinct codes under independent generations.} Mean number
of distinct 8-bit codes as additional independent generations from the same
prompt are included. Each panel averages results across 200 prompts and 200
random-hyperplane projections. Higher counts are desirable because they
indicate fewer collisions. Separate panels show SDXL and the two FLUX
variants, making transfer across generators visible.}
\label{fig:unique_codes_generations}
\end{figure}

\paragraph{Distinct codes under independent generations.}
We perform the corresponding code-count analysis on independent generations
from the same prompt. For each of SDXL and the two FLUX variants, we evaluate
200 prompts with up to 16 independently generated images per prompt. All
encoders are evaluated on the same images. For every encoder, results are
averaged over 200 random-hyperplane projections, with each projection fixed
across all images. Figure~\ref{fig:unique_codes_generations} reports the
corresponding code counts. More distinct codes mean fewer repeated candidate
codes. The proof of Theorem~\ref{thm:semantic-entropy-tpr} relates code occupancy
to detection probability.

\begin{figure}[!ht]
\centering
\includegraphics[width=0.70\textwidth]{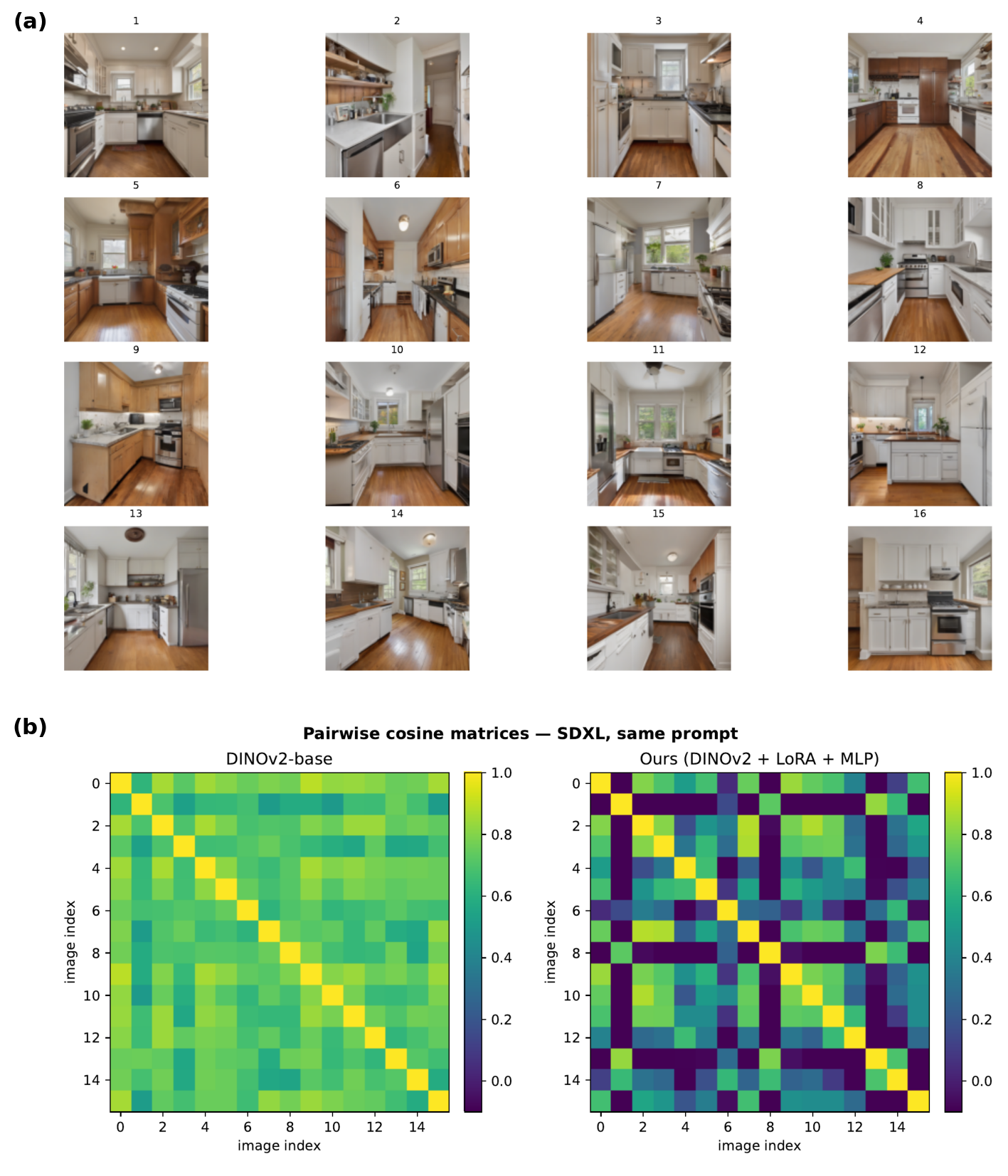}
\caption{\textbf{Different images from one prompt.} (a) Sixteen independent
SDXL generations. (b) Pairwise cosine similarities under DINOv2-base
and our encoder. DINOv2 assigns high similarity across the group, whereas
our encoder represents more of the differences between the images.
SDXL is the encoder's training domain; the prompts are held out.}
\label{fig:app_entropy_sdxl}
\end{figure}

\paragraph{Qualitative view of generation diversity.}
Figure~\ref{fig:app_entropy_sdxl} presents 16 SDXL images generated from one
prompt. Although the images share the same description, they differ in layout
and appearance. The cosine-similarity matrices show how DINOv2 and our encoder
represent these differences.

\clearpage

\section{Detection and Robustness Protocols}
\label{app:detection_robustness_overview}
\subsection{Evaluation Setup}
\label{app:robustness_tables}

\paragraph{Prompts and datasets.}
We evaluate clean detection using images generated from 1,000 captions from
the COCO-2017 validation set. For each prompt, we generate both watermarked
and unwatermarked images using the same model and generation settings. The
watermarked images form the positive population, while the unwatermarked
images form the negative population used to measure the false-positive rate.
These prompts are held out from semantic-encoder training; further details
about the encoder-training data are provided in
Appendix~\ref{app:semantic_encoder_training}.

\paragraph{Detection settings.}
The detector has three parameters: the boundary threshold $\delta$, the
Hamming radius $r_{\mathrm{ham}}$, and the score threshold $\tau$.
Boundary-aware detection is activated when an image representation lies
within margin $\delta$ of at least one of the $M$ random hyperplanes used to
construct its hash code. The activation rate is the fraction of clean
watermarked images for which this occurs. When activated, the detector
evaluates neighboring codes that differ in at most $r_{\mathrm{ham}}$ bits.
For each $(N,M)$ configuration, we select $\delta$ and $r_{\mathrm{ham}}$
through a parameter sweep on a separate development set drawn from the
COCO-2017 training set, considering clean detection, detection after editing,
and the corresponding false-positive rate. We use $r_{\mathrm{ham}}=1$
throughout and choose $\delta$ to obtain activation rates of $10\%$, $50\%$,
and $75\%$ for $(N,M)=(8,4)$, $(16,8)$, and $(32,12)$, respectively. The
sweep favors higher activation rates for larger $M$: longer codes introduce
more hyperplane boundaries that an edit may cross, while the larger candidate
budget provides a stronger watermark signal that can support more frequent
boundary correction. With $\delta$ and $r_{\mathrm{ham}}$ fixed, we vary
$\tau$ to obtain the required FPR. The selected settings are summarized in
Table~\ref{tab:detection_settings}.

\begin{table}[H]
\centering
\small
\caption{Selected detector settings. Because $\delta$ is calibrated separately
for each random-hyperplane frame, we report its target activation rate rather
than a single numeric value.}
\label{tab:detection_settings}
\begin{tabular}{ccc}
\toprule
$(N,M)$ & $\delta$ calibration target & $r_{\mathrm{ham}}$ \\
\midrule
$(8,4)$   & $10\%$ activation & 1 \\
$(16,8)$  & $50\%$ activation & 1 \\
$(32,12)$ & $75\%$ activation & 1 \\
\bottomrule
\end{tabular}
\end{table}

\begin{figure}[p]
    \centering
    \includegraphics[width=\linewidth]{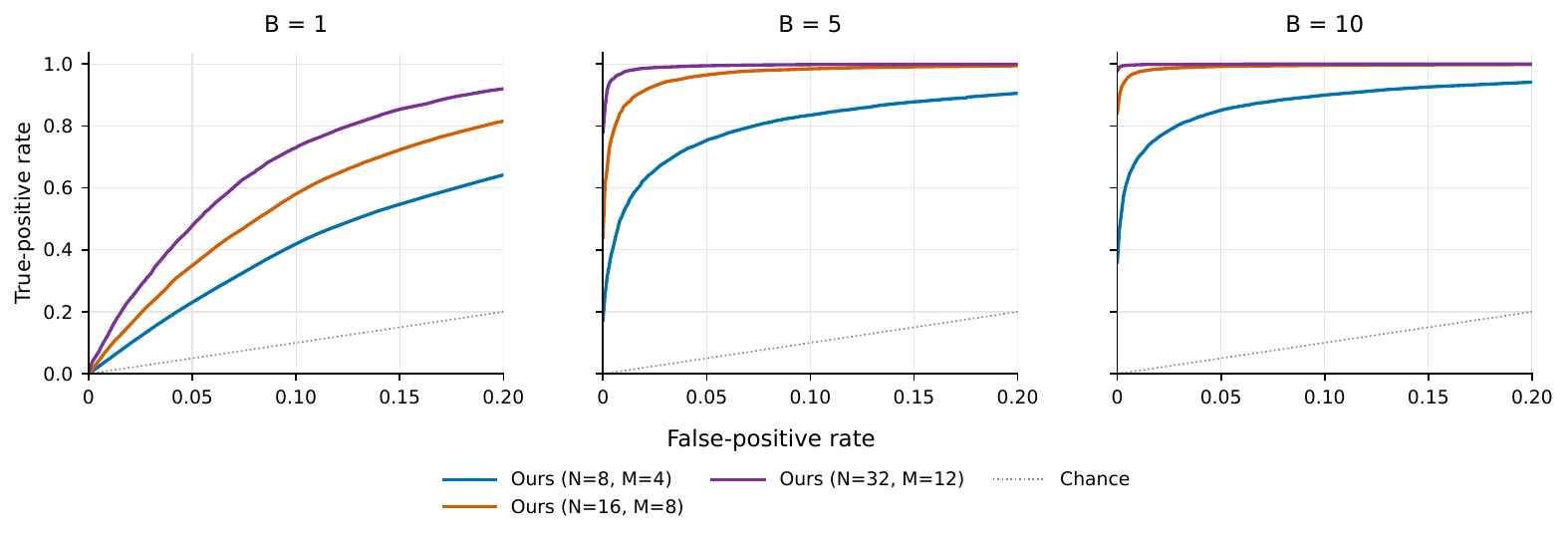}
    \caption{Clean detection for our method at $B\in\{1,5,10\}$ on a linear
    FPR axis. Curves are means over 25 hyperplane-frame and watermark-key
    combinations; Table~\ref{tab:clean_operating_points} gives numerical
    operating points.}
    \label{fig:experiment3_detection}
\end{figure}

\begin{figure}[p]
    \centering
    \includegraphics[width=\linewidth]{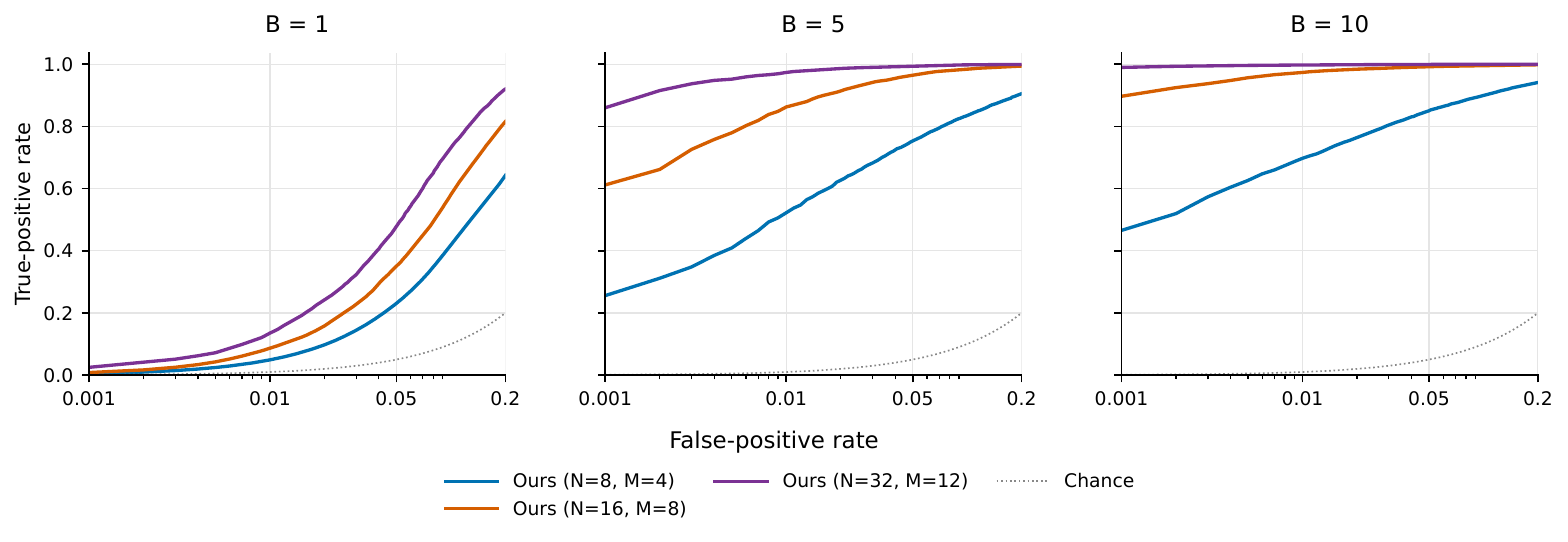}
    \caption{The same clean-detection curves on a logarithmic FPR axis,
    exposing aggregation gains at low FPR.}
    \label{fig:experiment3_detection_log}
\end{figure}

\paragraph{Clean-detection experiment.}
We evaluate detection from either a single image or a set of
$B\in\{5,10\}$ images generated using the same watermark key. For image sets,
we average the $B$ per-image detection scores and compare the resulting score
with $\tau$. Watermarked images form the positive population, and
unwatermarked images generated from the same prompts form the negative
population. For each configuration and set size, the false-positive rate is
the fraction of unwatermarked images or image sets whose score exceeds
$\tau$, while the true-positive rate is the corresponding fraction of
watermarked images or sets. We compute ROC curves from these held-out positive
and negative scores and report AUROC and TPR at empirical FPR targets, averaged
over 25 combinations of five random-hyperplane frames and five watermark keys.
Table~\ref{tab:experiment3_detection} summarizes the AUROC and TPR@20\%
results, and Figures~\ref{fig:experiment3_detection}
and~\ref{fig:experiment3_detection_log} show the corresponding TPR/FPR curves
on linear and logarithmic FPR axes.

Table~\ref{tab:clean_operating_points} reports the corresponding TPR at
$1\%$, $5\%$, and $20\%$ FPR for our configurations and the native
single-image baselines.

\begin{table}[p]
\centering
\small
\setlength{\tabcolsep}{6pt}
\caption{Clean-detection TPR (\%) at empirical FPR targets, averaged over 25
combinations of five hyperplane frames and five watermark keys. $B$ is the
number of images attributed to one source; baselines use their native
single-image detectors.}
\label{tab:clean_operating_points}
\begin{tabular}{llrrr}
\toprule
Method & $B$ & \multicolumn{3}{c}{TPR at FPR} \\
\cmidrule(lr){3-5}
 & & $1\%$ & $5\%$ & $20\%$ \\
\midrule
Ours $(N{=}8,M{=}4)$ & 1 & 4.9 & 23.1 & 64.2 \\
 & 5 & 52.3 & 75.3 & 90.5 \\
 & 10 & 69.7 & 85.0 & 94.1 \\
\addlinespace
Ours $(N{=}16,M{=}8)$ & 1 & 8.6 & 35.0 & 81.5 \\
 & 5 & 86.2 & 96.4 & 99.4 \\
 & 10 & 97.3 & 99.2 & 99.8 \\
\addlinespace
Ours $(N{=}32,M{=}12)$ & 1 & 13.5 & 47.8 & 92.0 \\
 & 5 & 97.3 & 99.4 & 99.9 \\
 & 10 & 99.7 & 99.9 & 100.0 \\
\midrule
Tree-Ring & 1 & 100.0 & 100.0 & 100.0 \\
Gaussian Shading & 1 & 100.0 & 100.0 & 100.0 \\
SEAL & 1 & 100.0 & 100.0 & 100.0 \\
\bottomrule
\end{tabular}
\end{table}

\FloatBarrier

\raggedbottom
\subsection{Robustness Protocols and Results}
\label{app:robustness_protocols}

\paragraph{Ordinary edits.}
We evaluate robustness on 200 prompts in the single-image setting ($B=1$).
Starting from clean watermarked and unwatermarked images, we apply the same
transformation to both populations. The detector parameters $\delta$,
$r_{\mathrm{ham}}$, and $\tau$ are fixed before applying the transformations
and are not recalibrated on edited images. For each transformation, we report
the edited TPR on watermarked images and the edited FPR on unwatermarked
images. We also report paired retention: the fraction of watermarked images
detected before transformation that remain detected afterward. This separates
robustness to an edit from the detector's initial clean-detection rate. The
evaluation includes 20 settings spanning compression, appearance changes,
noise and filtering, cropping, and translation.

DINOv2 is pretrained using augmentations including random resized crops,
color changes, and Gaussian blur, and therefore already maps many such
transformations to similar representations. Our fine-tuning preserves and
adapts this behavior using positive pairs constructed with noise, blur, JPEG
compression, resampling, photometric changes, and affine transformations,
while negative pairs separate independent generations of the same prompt.
The robustness evaluation includes transformations used during DINOv2
pretraining, transformations used during our fine-tuning, and transformations
used during neither. Figure~\ref{fig:robustness_per_edit_diagnostic} reports
the per-transformation results and marks this training exposure for each
transformation. This allows us to examine generalization to held-out images
and transformation settings, including transformation families not used
during training.

More generally, robustness depends on the geometry of the semantic
representation. For a target transformation, the method benefits from an
encoder that maps the original and transformed images to nearby
representations. Section~\ref{sec:encoder_analysis} compares DINOv2 with our
fine-tuned encoder and shows the relationship between encoder similarity and
watermark retention. For applications involving different transformations,
the semantic encoder can therefore be selected or trained to maintain high
cosine similarity under those transformations.

\begin{figure}[H]
\centering
\resizebox{\linewidth}{!}{\input{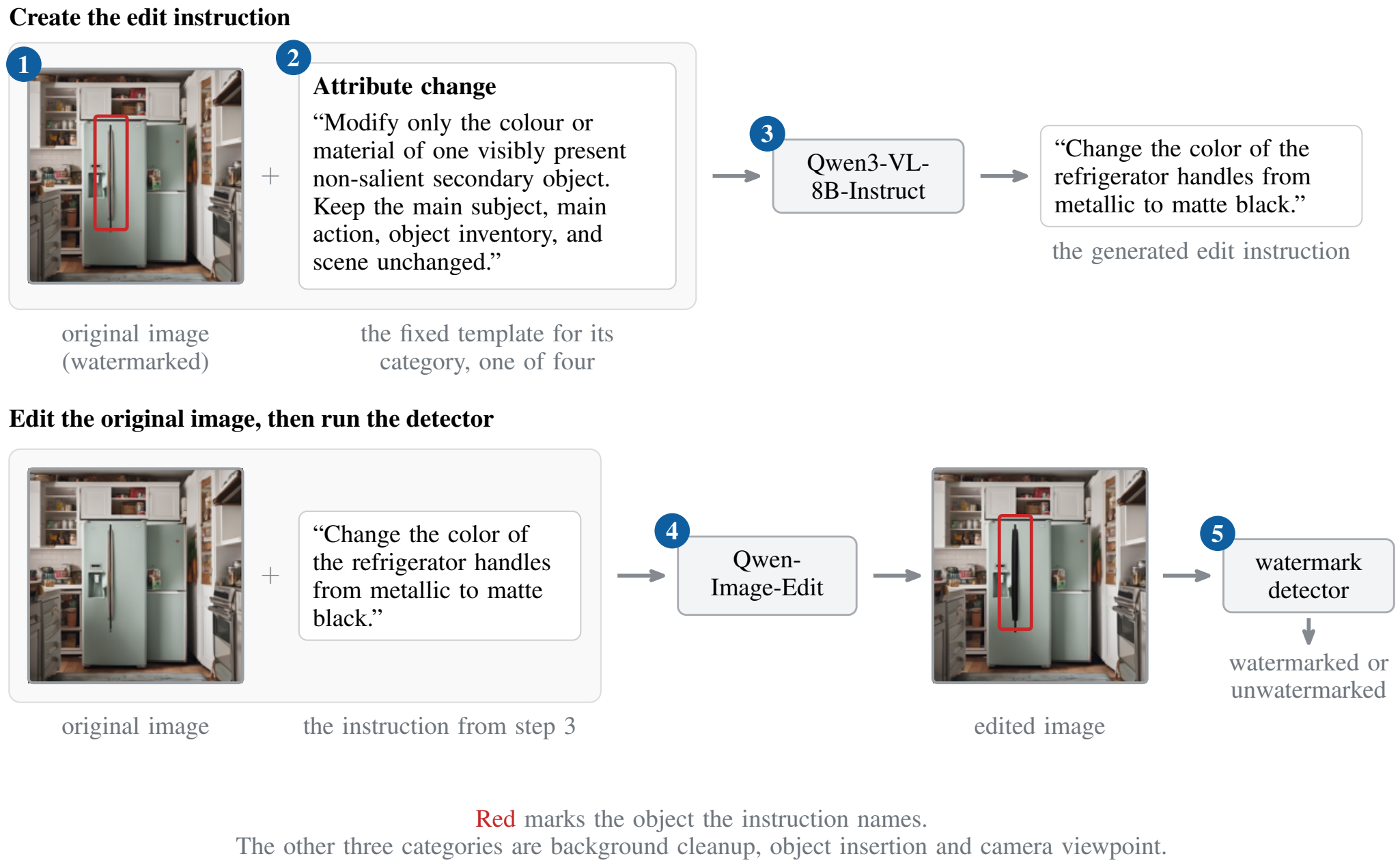}}
\caption{\textbf{Generative-edit evaluation protocol.}
For each source image, the assigned category prompt and image are given to
Qwen3-VL-8B-Instruct, which produces a validated image-specific instruction.
The original image and this instruction are then passed to Qwen-Image-Edit,
and the edited result is evaluated using detector parameters fixed before
editing. The same pipeline is applied to watermarked and corresponding
unwatermarked sources. The example changes the refrigerator handles from
metallic to matte black; the boxes identify the object named by the
instruction, not a guaranteed pixel-local change.}
\label{fig:ai_edit_protocol}
\end{figure}

For the non-semantic generative-edit evaluation, we use 80 watermarked images
and 80 corresponding unwatermarked images, with 20 examples assigned to each
of four edit categories: attribute change, background cleanup, object
insertion, and camera-viewpoint change. For each source image, we provide
Qwen3-VL-8B-Instruct~\citep{bai2025qwen3vl} with the original image and the fixed category-specific
prompt reported in Table~\ref{tab:ai_edit_prompts}. The vision model then
produces an image-specific editing instruction that follows the assigned
category while preserving the main subject and scene. Qwen-Image-Edit~\citep{wu2025qwenimage} receives
the original image together with this instruction and generates the edited
image. Figure~\ref{fig:ai_edit_protocol} shows this pipeline on one
attribute-change example. We manually review every edited image to verify that
the requested change is visible while the main subject and overall scene remain
preserved.
Representative examples from all four edit categories are shown in
Figure~\ref{fig:ai_edit_examples}.

\begin{table}[H]
\centering
\small
\caption{Fixed category-specific constraints provided to Qwen3-VL-8B-Instruct
when generating image-specific editing instructions.}
\label{tab:ai_edit_prompts}
\begin{tabular}{p{3.0cm}p{10.0cm}}
\toprule
Edit category & Fixed prompt constraint \\
\midrule
Attribute change & Modify only the colour or material of one visibly present,
non-salient secondary object. Preserve the main subject, main action, object
inventory, and scene. Do not target faces, text, logos, or watermarks. \\
Background cleanup & Remove only one tiny, non-salient background clutter
item. Preserve the main subject, main action, salient objects, and scene. Do not
target faces, text, logos, or watermarks. \\
Object insertion & Add one unrelated small realistic cat in the background.
The cat must be clearly visible but must not replace or occlude the main
subject. Do not target faces, text, logos, or watermarks. \\
Camera-viewpoint change & Change only the camera viewpoint to a clearly
different three-quarter or slightly elevated viewpoint. Preserve the same main
subject, objects, actions, scene, colours, and lighting. Do not target faces,
text, logos, or watermarks. \\
\bottomrule
\end{tabular}
\end{table}

All results average five hyperplane frames and five PRF keys. Paired retention
matches original and edited images within each method. For ordinary edits,
achieved FPR is measured on transformed negatives and each reported category is
an unweighted mean over its transformation settings. Table~\ref{tab:experiment3_robustness}
summarizes the ordinary and generative-edit results. Detailed
per-transformation TPR and retention results are reported in
Figure~\ref{fig:robustness_per_edit_diagnostic}.

\begin{table*}[t]
\centering
\scriptsize
\caption{Robustness at $20\%$ nominal clean FPR (all values in \%). Columns
report post-edit TPR, paired retention conditioned on original detection, and
post-edit FPR on unwatermarked images. Appearance/compression and geometric
results are unweighted means over 14 and six settings, respectively;
generative results summarize 80 edits. Detector settings are given in
Table~\ref{tab:detection_settings}.}
\label{tab:experiment3_robustness}
\setlength{\tabcolsep}{3.5pt}
\begin{tabular}{lccc@{\qquad}ccc@{\qquad}ccc}
\toprule
& \multicolumn{3}{c}{Appearance/compression}
& \multicolumn{3}{c}{Geometric}
& \multicolumn{3}{c}{Generative edits} \\
\cmidrule(lr){2-4}\cmidrule(lr){5-7}\cmidrule(lr){8-10}
Method & TPR & Ret. & FPR & TPR & Ret. & FPR & TPR & Ret. & FPR \\
\midrule
Ours $(N{=}8,M{=}4)$   &55.4&93.1&17.2&41.0&65.9&17.6&41.2&76.7&17.8 \\
Ours $(N{=}16,M{=}8)$  &76.6&92.7&19.6&48.9&57.5&20.2&70.0&84.8&19.2 \\
Ours $(N{=}32,M{=}12)$ &83.5&91.4&19.2&48.2&51.8&20.0&66.2&74.6&20.5 \\
Tree-Ring               &100.0&100.0&22.6&46.7&46.7&21.7&60.0&60.0&21.0 \\
Gaussian Shading        &100.0&100.0&21.6&27.4&27.4&23.7&37.5&37.5&22.0 \\
SEAL                     &97.9&97.9&18.7&14.5&14.5&19.2&10.0&10.0&19.5 \\
\bottomrule
\end{tabular}
\end{table*}

The boundary-aware settings were selected on the separate development set
described above. Table~\ref{tab:robustness_detector_ablation} evaluates their
effect on the held-out ordinary-edit evaluation set by comparing them with
exact-code detection on the same images within each $(N,M)$ pair.

\begin{table}[H]
\centering
\small
\setlength{\tabcolsep}{5pt}
\caption{Boundary-aware detector ablation on the held-out ordinary-edit
evaluation set at $20\%$ nominal clean FPR. Values are unweighted means over
the 20 reported transformation settings (14 appearance/compression and six
geometric). TPR and FPR are measured after transformation, and retention is
conditioned on original detection. Exact-code detection uses no boundary
expansion; boundary-aware detection uses the settings in
Table~\ref{tab:detection_settings}. Each pair uses the same images.}
\label{tab:robustness_detector_ablation}
\begin{tabular}{ccrrr}
\toprule
$(N,M)$ & Detector & TPR & Retention & FPR \\
\midrule
$(8,4)$ & Exact code & 50.54 & 82.70 & 16.75 \\
 & Boundary aware & 51.06 & 84.96 & 17.33 \\
\addlinespace
$(16,8)$ & Exact code & 65.44 & 72.64 & 19.13 \\
 & Boundary aware & 68.28 & 82.14 & 19.78 \\
\addlinespace
$(32,12)$ & Exact code & 65.63 & 66.70 & 18.53 \\
 & Boundary aware & \textbf{72.91} & 79.51 & 19.47 \\
\bottomrule
\end{tabular}
\end{table}

\clearpage
\flushbottom
\begin{table}[p]
\centering
\scriptsize
\caption{Robustness at $1\%$ and $5\%$ nominal clean FPR (all values in \%). Columns
report TPR before editing (clean), post-edit TPR, and paired retention conditioned on original detection.
Appearance/compression and geometric
results are unweighted means over 14 and six settings, respectively;
generative results summarize 80 edits. Detector settings are given in
Table~\ref{tab:detection_settings}.}
\label{tab:robustness_single_lowfpr}
\setlength{\tabcolsep}{3.5pt}
\begin{tabular}{llccc@{\qquad}ccc@{\qquad}ccc}
\toprule
& & \multicolumn{3}{c}{Appearance/compression}
& \multicolumn{3}{c}{Geometric}
& \multicolumn{3}{c}{Generative edits} \\
\cmidrule(lr){3-5}\cmidrule(lr){6-8}\cmidrule(lr){9-11}
FPR & Method & Clean & Edited & Ret. & Clean & Edited & Ret. & Clean & Edited & Ret. \\
\midrule
$1\%$ & Ours $(N{=}8,M{=}4)$ & 0.0 & 0.0 & -- & 0.0 & 0.0 & -- & 0.0 & 0.0 & -- \\
 & Ours $(N{=}16,M{=}8)$ & 7.8 & 7.1 & 87.2 & 7.8 & 3.8 & 42.0 & 8.8 & 5.0 & 57.1 \\
 & Ours $(N{=}32,M{=}12)$ & 14.6 & 13.0 & 86.6 & 14.6 & 5.9 & 34.2 & 13.8 & 7.5 & 54.5 \\
 & Tree-Ring & 100.0 & 100.0 & 100.0 & 100.0 & 19.5 & 19.5 & 100.0 & 17.5 & 17.5 \\
 & Gaussian Shading & 100.0 & 100.0 & 100.0 & 100.0 & 1.8 & 1.8 & 100.0 & 7.5 & 7.5 \\
 & SEAL & 100.0 & 96.0 & 96.0 & 100.0 & 1.2 & 1.2 & 100.0 & 0.0 & 0.0 \\
\midrule
$5\%$ & Ours $(N{=}8,M{=}4)$ & 8.4 & 7.9 & 90.0 & 8.4 & 5.3 & 54.0 & 12.5 & 12.5 & 100.0 \\
 & Ours $(N{=}16,M{=}8)$ & 32.6 & 29.9 & 90.4 & 32.6 & 17.3 & 48.5 & 31.2 & 27.5 & 84.0 \\
 & Ours $(N{=}32,M{=}12)$ & 49.2 & 44.3 & 89.0 & 49.2 & 21.7 & 40.8 & 55.0 & 37.5 & 68.2 \\
 & Tree-Ring & 100.0 & 100.0 & 100.0 & 100.0 & 26.6 & 26.6 & 100.0 & 30.0 & 30.0 \\
 & Gaussian Shading & 100.0 & 100.0 & 100.0 & 100.0 & 11.5 & 11.5 & 100.0 & 17.5 & 17.5 \\
 & SEAL & 100.0 & 96.0 & 96.0 & 100.0 & 1.2 & 1.2 & 100.0 & 0.0 & 0.0 \\
\bottomrule
\end{tabular}
\end{table}

\begin{table}[p]
\centering
\scriptsize
\caption{Batch detection after editing for our configurations (all values in
\%). Each batch contains $B$ images from different prompts that received the
same transformation. FPRs are nominal, with thresholds calibrated on
additional unedited negatives.}
\label{tab:robustness_lowfpr_batches}
\setlength{\tabcolsep}{4pt}
\begin{tabular}{llccc@{\qquad}ccc}
\toprule
\multicolumn{8}{c}{\textbf{\boldmath$B=5$}} \\
\midrule
& & \multicolumn{3}{c}{Appearance/compression} & \multicolumn{3}{c}{Geometric} \\
\cmidrule(lr){3-5}\cmidrule(lr){6-8}
FPR & Method & Clean & Edited & Ret. & Clean & Edited & Ret. \\
\midrule
$1\%$ & Ours $(N{=}8,M{=}4)$ & 62.1 & 52.2 & 82.7 & 62.1 & 22.1 & 33.6 \\
 & Ours $(N{=}16,M{=}8)$ & 92.3 & 77.8 & 83.9 & 92.3 & 23.4 & 25.1 \\
 & Ours $(N{=}32,M{=}12)$ & 98.9 & 82.5 & 83.3 & 98.9 & 21.6 & 21.8 \\
\midrule
$5\%$ & Ours $(N{=}8,M{=}4)$ & 85.8 & 77.8 & 90.1 & 85.8 & 45.2 & 51.4 \\
 & Ours $(N{=}16,M{=}8)$ & 99.0 & 90.2 & 91.1 & 99.0 & 41.5 & 41.8 \\
 & Ours $(N{=}32,M{=}12)$ & 100.0 & 90.3 & 90.4 & 100.0 & 36.7 & 36.7 \\
\midrule
$20\%$ & Ours $(N{=}8,M{=}4)$ & 97.0 & 93.7 & 96.3 & 97.0 & 72.5 & 74.2 \\
 & Ours $(N{=}16,M{=}8)$ & 100.0 & 97.0 & 97.1 & 100.0 & 65.9 & 65.9 \\
 & Ours $(N{=}32,M{=}12)$ & 100.0 & 96.5 & 96.5 & 100.0 & 60.1 & 60.1 \\
\midrule[\heavyrulewidth]
\addlinespace[4pt]
\multicolumn{8}{c}{\textbf{\boldmath$B=10$}} \\
\midrule
& & \multicolumn{3}{c}{Appearance/compression} & \multicolumn{3}{c}{Geometric} \\
\cmidrule(lr){3-5}\cmidrule(lr){6-8}
FPR & Method & Clean & Edited & Ret. & Clean & Edited & Ret. \\
\midrule
$1\%$ & Ours $(N{=}8,M{=}4)$ & 91.9 & 83.5 & 90.3 & 91.9 & 42.2 & 45.4 \\
 & Ours $(N{=}16,M{=}8)$ & 99.9 & 93.3 & 93.4 & 99.9 & 38.0 & 38.0 \\
 & Ours $(N{=}32,M{=}12)$ & 100.0 & 93.3 & 93.3 & 100.0 & 33.9 & 33.9 \\
\midrule
$5\%$ & Ours $(N{=}8,M{=}4)$ & 98.2 & 94.7 & 96.2 & 98.2 & 65.6 & 66.6 \\
 & Ours $(N{=}16,M{=}8)$ & 100.0 & 97.5 & 97.5 & 100.0 & 57.5 & 57.5 \\
 & Ours $(N{=}32,M{=}12)$ & 100.0 & 96.8 & 96.8 & 100.0 & 50.1 & 50.1 \\
\midrule
$20\%$ & Ours $(N{=}8,M{=}4)$ & 99.8 & 99.0 & 99.1 & 99.8 & 85.5 & 85.6 \\
 & Ours $(N{=}16,M{=}8)$ & 100.0 & 99.3 & 99.3 & 100.0 & 77.2 & 77.2 \\
 & Ours $(N{=}32,M{=}12)$ & 100.0 & 99.0 & 99.0 & 100.0 & 71.0 & 71.0 \\
\bottomrule
\end{tabular}
\end{table}

\clearpage
\begin{figure}[p]
\centering
\definecolor{metricbest}{HTML}{007A53}
\definecolor{metricsecond}{HTML}{C76A00}
\begingroup
\scriptsize
\setlength{\tabcolsep}{3pt}
\renewcommand{\arraystretch}{0.96}
\newcommand{\trainyes}{\(\checkmark\)}
\newcommand{\trainno}{\(\times\)}
\textcolor{metricbest}{\textbf{Best}} / \textcolor{metricsecond}{\textbf{Second}}
\par\smallskip
\textbf{Edited TPR (\%)}\par\smallskip
\begin{tabular}{lcclrrrrrr}
\toprule
Edit & \shortstack{DINOv2\\pretrain} & \shortstack{Ours\\fine-tune} & & \shortstack{Ours\\$(8,4)$} & \shortstack{Ours\\$(16,8)$} & \shortstack{Ours\\$(32,12)$} & Tree-Ring & \shortstack{Gaussian\\Shading} & SEAL \\
\midrule
\multicolumn{10}{l}{\textbf{Appearance and compression}} \\
JPEG Q90 & \trainno & \trainyes && 58.0 & 81.4 & 90.1 & \textcolor{metricbest}{\textbf{100.0}} & \textcolor{metricbest}{\textbf{100.0}} & \textcolor{metricsecond}{\textbf{99.5}} \\
JPEG Q75 & \trainno & \trainyes && 56.4 & 80.1 & 88.3 & \textcolor{metricbest}{\textbf{100.0}} & \textcolor{metricbest}{\textbf{100.0}} & \textcolor{metricsecond}{\textbf{99.5}} \\
\addlinespace[2pt]
Gaussian blur $r=0.5$ & \trainyes & \trainyes && 58.4 & 81.8 & \textcolor{metricsecond}{\textbf{90.6}} & \textcolor{metricbest}{\textbf{100.0}} & \textcolor{metricbest}{\textbf{100.0}} & \textcolor{metricbest}{\textbf{100.0}} \\
Gaussian blur $r=1.5$ & \trainyes & \trainyes && 54.7 & 76.4 & \textcolor{metricsecond}{\textbf{83.0}} & \textcolor{metricbest}{\textbf{100.0}} & \textcolor{metricbest}{\textbf{100.0}} & \textcolor{metricbest}{\textbf{100.0}} \\
\addlinespace[2pt]
Gaussian noise $\sigma=2$ & \trainno & \trainyes && 57.6 & 81.2 & \textcolor{metricsecond}{\textbf{89.5}} & \textcolor{metricbest}{\textbf{100.0}} & \textcolor{metricbest}{\textbf{100.0}} & \textcolor{metricbest}{\textbf{100.0}} \\
Gaussian noise $\sigma=6$ & \trainno & \trainyes && 54.4 & 76.2 & 82.7 & \textcolor{metricbest}{\textbf{100.0}} & \textcolor{metricbest}{\textbf{100.0}} & \textcolor{metricsecond}{\textbf{99.5}} \\
\addlinespace[2pt]
Brightness $\times0.8$ & \trainyes & \trainyes && 58.1 & 81.3 & \textcolor{metricsecond}{\textbf{89.9}} & \textcolor{metricbest}{\textbf{100.0}} & \textcolor{metricbest}{\textbf{100.0}} & \textcolor{metricbest}{\textbf{100.0}} \\
Brightness $\times1.2$ & \trainyes & \trainyes && 56.7 & 79.2 & \textcolor{metricsecond}{\textbf{86.4}} & \textcolor{metricbest}{\textbf{100.0}} & \textcolor{metricbest}{\textbf{100.0}} & \textcolor{metricbest}{\textbf{100.0}} \\
\addlinespace[2pt]
Contrast $\times0.8$ & \trainyes & \trainyes && 58.1 & 81.3 & \textcolor{metricsecond}{\textbf{90.0}} & \textcolor{metricbest}{\textbf{100.0}} & \textcolor{metricbest}{\textbf{100.0}} & \textcolor{metricbest}{\textbf{100.0}} \\
Contrast $\times1.2$ & \trainyes & \trainyes && 56.9 & 79.2 & \textcolor{metricsecond}{\textbf{87.0}} & \textcolor{metricbest}{\textbf{100.0}} & \textcolor{metricbest}{\textbf{100.0}} & \textcolor{metricbest}{\textbf{100.0}} \\
\addlinespace[2pt]
Median $3\times3$ & \trainno & \trainno && 57.0 & 79.9 & \textcolor{metricsecond}{\textbf{87.9}} & \textcolor{metricbest}{\textbf{100.0}} & \textcolor{metricbest}{\textbf{100.0}} & \textcolor{metricbest}{\textbf{100.0}} \\
Median $5\times5$ & \trainno & \trainno && 52.5 & 71.2 & \textcolor{metricsecond}{\textbf{76.3}} & \textcolor{metricbest}{\textbf{100.0}} & \textcolor{metricbest}{\textbf{100.0}} & \textcolor{metricbest}{\textbf{100.0}} \\
\addlinespace[2pt]
Salt-and-pepper $1\%$ & \trainno & \trainno && 50.0 & 65.1 & 68.7 & \textcolor{metricbest}{\textbf{100.0}} & \textcolor{metricbest}{\textbf{100.0}} & \textcolor{metricsecond}{\textbf{91.5}} \\
Salt-and-pepper $3\%$ & \trainno & \trainno && 46.6 & 57.7 & 58.9 & \textcolor{metricbest}{\textbf{100.0}} & \textcolor{metricbest}{\textbf{100.0}} & \textcolor{metricsecond}{\textbf{80.5}} \\
\midrule
\multicolumn{10}{l}{\textbf{Geometric transformations}} \\
Center crop $50\%$ & \trainyes & \trainno && \textcolor{metricsecond}{\textbf{32.4}} & \textcolor{metricbest}{\textbf{32.6}} & 29.4 & 18.5 & 20.5 & 14.0 \\
Center crop $75\%$ & \trainyes & \trainno && 41.4 & \textcolor{metricsecond}{\textbf{49.1}} & 48.3 & \textcolor{metricbest}{\textbf{54.5}} & 32.5 & 9.0 \\
\addlinespace[2pt]
Off-center crop $75\%$ & \trainyes & \trainno && \textcolor{metricsecond}{\textbf{35.1}} & 35.0 & 31.8 & \textcolor{metricbest}{\textbf{38.5}} & 23.5 & 7.0 \\
Off-center crop $90\%$ & \trainyes & \trainno && 43.9 & \textcolor{metricbest}{\textbf{54.9}} & \textcolor{metricsecond}{\textbf{54.3}} & 47.5 & 41.0 & 18.5 \\
\addlinespace[2pt]
Shift 32 px & \trainno & \trainyes && 43.4 & \textcolor{metricsecond}{\textbf{54.8}} & \textcolor{metricbest}{\textbf{55.1}} & 21.0 & 23.5 & 30.0 \\
Shift 8 px & \trainno & \trainyes && 49.6 & 67.1 & \textcolor{metricsecond}{\textbf{70.0}} & \textcolor{metricbest}{\textbf{100.0}} & 23.5 & 8.5 \\
\midrule
\multicolumn{10}{l}{\textbf{Edits from an image-editing model}} \\
Attribute change & \trainno & \trainno && 45.0 & \textcolor{metricsecond}{\textbf{65.0}} & \textcolor{metricbest}{\textbf{70.0}} & \textcolor{metricsecond}{\textbf{65.0}} & 25.0 & 10.0 \\
Background cleanup & \trainno & \trainno && 45.0 & \textcolor{metricsecond}{\textbf{80.0}} & \textcolor{metricbest}{\textbf{90.0}} & 65.0 & 50.0 & 15.0 \\
Cat insertion & \trainno & \trainno && 45.0 & \textcolor{metricbest}{\textbf{80.0}} & 65.0 & \textcolor{metricsecond}{\textbf{75.0}} & 60.0 & 10.0 \\
Camera-viewpoint change & \trainno & \trainno && 30.0 & \textcolor{metricbest}{\textbf{55.0}} & \textcolor{metricsecond}{\textbf{40.0}} & 35.0 & 15.0 & 5.0 \\
\bottomrule
\end{tabular}
\par\medskip
\textbf{Paired retention (\%)}\par\smallskip
\begin{tabular}{lcclrrrrrr}
\toprule
Edit & \shortstack{DINOv2\\pretrain} & \shortstack{Ours\\fine-tune} & & \shortstack{Ours\\$(8,4)$} & \shortstack{Ours\\$(16,8)$} & \shortstack{Ours\\$(32,12)$} & Tree-Ring & \shortstack{Gaussian\\Shading} & SEAL \\
\midrule
\multicolumn{10}{l}{\textbf{Appearance and compression}} \\
JPEG Q90 & \trainno & \trainyes && 98.2 & 99.1 & 98.9 & \textcolor{metricbest}{\textbf{100.0}} & \textcolor{metricbest}{\textbf{100.0}} & \textcolor{metricsecond}{\textbf{99.5}} \\
JPEG Q75 & \trainno & \trainyes && 95.4 & 97.2 & 96.7 & \textcolor{metricbest}{\textbf{100.0}} & \textcolor{metricbest}{\textbf{100.0}} & \textcolor{metricsecond}{\textbf{99.5}} \\
\addlinespace[2pt]
Gaussian blur $r=0.5$ & \trainyes & \trainyes && 99.3 & \textcolor{metricsecond}{\textbf{99.6}} & 99.3 & \textcolor{metricbest}{\textbf{100.0}} & \textcolor{metricbest}{\textbf{100.0}} & \textcolor{metricbest}{\textbf{100.0}} \\
Gaussian blur $r=1.5$ & \trainyes & \trainyes && 92.0 & \textcolor{metricsecond}{\textbf{92.4}} & 90.8 & \textcolor{metricbest}{\textbf{100.0}} & \textcolor{metricbest}{\textbf{100.0}} & \textcolor{metricbest}{\textbf{100.0}} \\
\addlinespace[2pt]
Gaussian noise $\sigma=2$ & \trainno & \trainyes && 97.3 & \textcolor{metricsecond}{\textbf{98.6}} & 98.0 & \textcolor{metricbest}{\textbf{100.0}} & \textcolor{metricbest}{\textbf{100.0}} & \textcolor{metricbest}{\textbf{100.0}} \\
Gaussian noise $\sigma=6$ & \trainno & \trainyes && 91.3 & 91.9 & 90.5 & \textcolor{metricbest}{\textbf{100.0}} & \textcolor{metricbest}{\textbf{100.0}} & \textcolor{metricsecond}{\textbf{99.5}} \\
\addlinespace[2pt]
Brightness $\times0.8$ & \trainyes & \trainyes && 98.4 & \textcolor{metricsecond}{\textbf{98.9}} & 98.6 & \textcolor{metricbest}{\textbf{100.0}} & \textcolor{metricbest}{\textbf{100.0}} & \textcolor{metricbest}{\textbf{100.0}} \\
Brightness $\times1.2$ & \trainyes & \trainyes && 95.4 & \textcolor{metricsecond}{\textbf{95.9}} & 94.6 & \textcolor{metricbest}{\textbf{100.0}} & \textcolor{metricbest}{\textbf{100.0}} & \textcolor{metricbest}{\textbf{100.0}} \\
\addlinespace[2pt]
Contrast $\times0.8$ & \trainyes & \trainyes && 98.3 & \textcolor{metricsecond}{\textbf{98.8}} & 98.6 & \textcolor{metricbest}{\textbf{100.0}} & \textcolor{metricbest}{\textbf{100.0}} & \textcolor{metricbest}{\textbf{100.0}} \\
Contrast $\times1.2$ & \trainyes & \trainyes && 95.7 & \textcolor{metricsecond}{\textbf{96.1}} & 95.2 & \textcolor{metricbest}{\textbf{100.0}} & \textcolor{metricbest}{\textbf{100.0}} & \textcolor{metricbest}{\textbf{100.0}} \\
\addlinespace[2pt]
Median $3\times3$ & \trainno & \trainno && 96.4 & \textcolor{metricsecond}{\textbf{96.9}} & 96.3 & \textcolor{metricbest}{\textbf{100.0}} & \textcolor{metricbest}{\textbf{100.0}} & \textcolor{metricbest}{\textbf{100.0}} \\
Median $5\times5$ & \trainno & \trainno && \textcolor{metricsecond}{\textbf{87.5}} & 85.6 & 83.2 & \textcolor{metricbest}{\textbf{100.0}} & \textcolor{metricbest}{\textbf{100.0}} & \textcolor{metricbest}{\textbf{100.0}} \\
\addlinespace[2pt]
Salt-and-pepper $1\%$ & \trainno & \trainno && 82.9 & 78.0 & 74.8 & \textcolor{metricbest}{\textbf{100.0}} & \textcolor{metricbest}{\textbf{100.0}} & \textcolor{metricsecond}{\textbf{91.5}} \\
Salt-and-pepper $3\%$ & \trainno & \trainno && 76.1 & 68.9 & 63.9 & \textcolor{metricbest}{\textbf{100.0}} & \textcolor{metricbest}{\textbf{100.0}} & \textcolor{metricsecond}{\textbf{80.5}} \\
\midrule
\multicolumn{10}{l}{\textbf{Geometric transformations}} \\
Center crop $50\%$ & \trainyes & \trainno && \textcolor{metricbest}{\textbf{49.1}} & \textcolor{metricsecond}{\textbf{36.7}} & 30.7 & 18.5 & 20.5 & 14.0 \\
Center crop $75\%$ & \trainyes & \trainno && \textcolor{metricbest}{\textbf{67.1}} & \textcolor{metricsecond}{\textbf{57.7}} & 52.0 & 54.5 & 32.5 & 9.0 \\
\addlinespace[2pt]
Off-center crop $75\%$ & \trainyes & \trainno && \textcolor{metricbest}{\textbf{54.1}} & \textcolor{metricsecond}{\textbf{39.5}} & 33.5 & 38.5 & 23.5 & 7.0 \\
Off-center crop $90\%$ & \trainyes & \trainno && \textcolor{metricbest}{\textbf{71.7}} & \textcolor{metricsecond}{\textbf{65.4}} & 58.7 & 47.5 & 41.0 & 18.5 \\
\addlinespace[2pt]
Shift 32 px & \trainno & \trainyes && \textcolor{metricbest}{\textbf{70.7}} & \textcolor{metricsecond}{\textbf{65.1}} & 59.6 & 21.0 & 23.5 & 30.0 \\
Shift 8 px & \trainno & \trainyes && \textcolor{metricsecond}{\textbf{82.4}} & 80.5 & 76.1 & \textcolor{metricbest}{\textbf{100.0}} & 23.5 & 8.5 \\
\midrule
\multicolumn{10}{l}{\textbf{Edits from an image-editing model}} \\
Attribute change & \trainno & \trainno && \textcolor{metricbest}{\textbf{90.0}} & \textcolor{metricsecond}{\textbf{86.7}} & 77.8 & 65.0 & 25.0 & 10.0 \\
Background cleanup & \trainno & \trainno && 90.0 & \textcolor{metricbest}{\textbf{100.0}} & \textcolor{metricsecond}{\textbf{94.7}} & 65.0 & 50.0 & 15.0 \\
Cat insertion & \trainno & \trainno && \textcolor{metricsecond}{\textbf{81.8}} & \textcolor{metricbest}{\textbf{94.1}} & 81.2 & 75.0 & 60.0 & 10.0 \\
Camera-viewpoint change & \trainno & \trainno && \textcolor{metricsecond}{\textbf{50.0}} & \textcolor{metricbest}{\textbf{61.1}} & 44.4 & 35.0 & 15.0 & 5.0 \\
\bottomrule
\end{tabular}
\endgroup

\caption{Per-edit detection and paired retention for the 20 selected
appearance/compression and geometric settings, followed by four
edit families produced by an image-editing model, at 20\% nominal clean FPR. The upper section reports edited
TPR and the lower section reports paired retention; all values are percentages.
The two left columns in both sections mark transformation-family exposure
during DINOv2 pretraining and our encoder fine-tuning.
Bold green and orange indicate the best and second-best values within each row;
ties at the displayed precision share a color. Detector configurations and evaluation protocols are given in
Appendix~\ref{app:robustness_tables}.}
\label{fig:robustness_per_edit_diagnostic}
\end{figure}

\begin{figure}[p]
\centering
\textbf{Ordinary-edit examples (1/3)}\par\bigskip
\begingroup
\small
\setlength{\tabcolsep}{5pt}
\begin{tabular}{@{}lccc@{}}
\toprule
Family & Original & Setting 1 & Setting 2 \\
\midrule
\multicolumn{4}{l}{\textbf{Appearance and compression}} \\[6pt]
\parbox[c]{2.1cm}{JPEG} &
\begin{minipage}[c]{3cm}\centering\includegraphics[width=3cm]{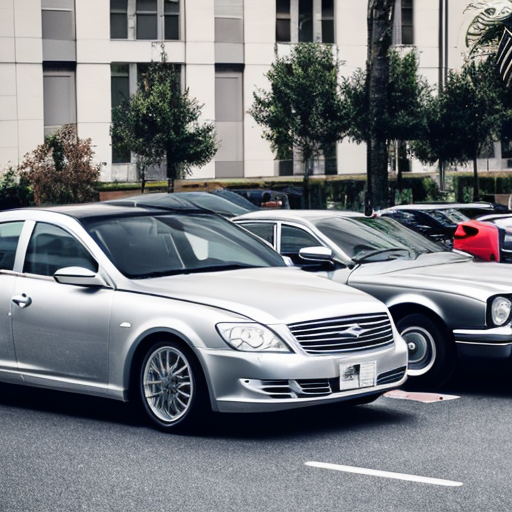}\par\vspace{10pt}\end{minipage} &
\begin{minipage}[c]{3cm}\centering\includegraphics[width=3cm]{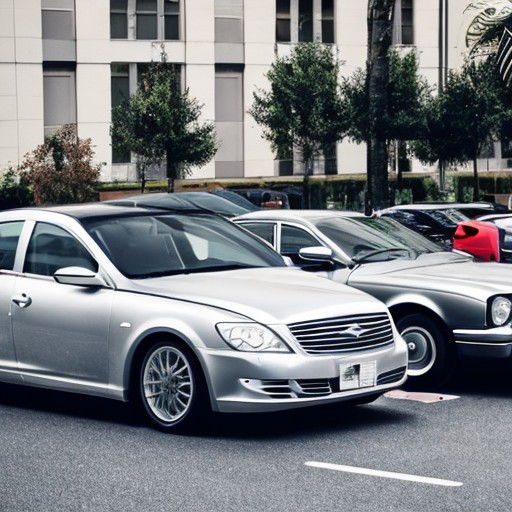}\par\scriptsize Q90\end{minipage} &
\begin{minipage}[c]{3cm}\centering\includegraphics[width=3cm]{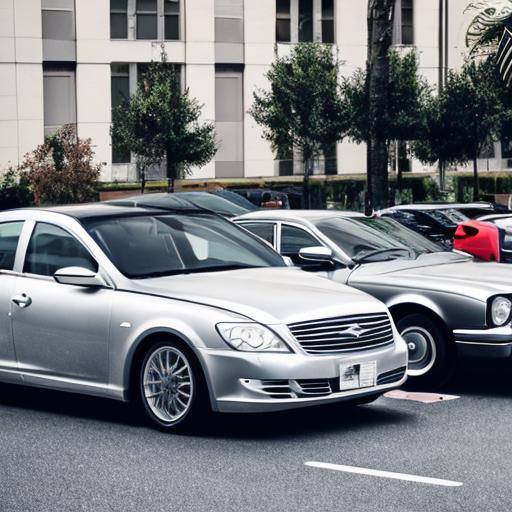}\par\scriptsize Q75\end{minipage} \\[8pt]
\specialrule{0.3pt}{0pt}{6pt}
\parbox[c]{2.1cm}{Gaussian blur} &
\begin{minipage}[c]{3cm}\centering\includegraphics[width=3cm]{robustness_example_selected_original.png}\par\vspace{10pt}\end{minipage} &
\begin{minipage}[c]{3cm}\centering\includegraphics[width=3cm]{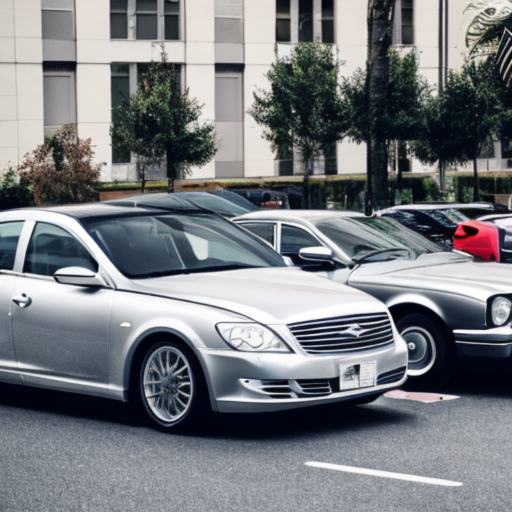}\par\scriptsize $r=0.5$\end{minipage} &
\begin{minipage}[c]{3cm}\centering\includegraphics[width=3cm]{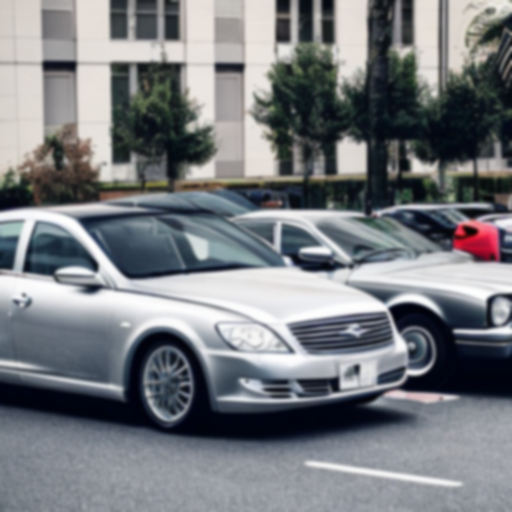}\par\scriptsize $r=1.5$\end{minipage} \\[8pt]
\specialrule{0.3pt}{0pt}{6pt}
\parbox[c]{2.1cm}{Gaussian noise} &
\begin{minipage}[c]{3cm}\centering\includegraphics[width=3cm]{robustness_example_selected_original.png}\par\vspace{10pt}\end{minipage} &
\begin{minipage}[c]{3cm}\centering\includegraphics[width=3cm]{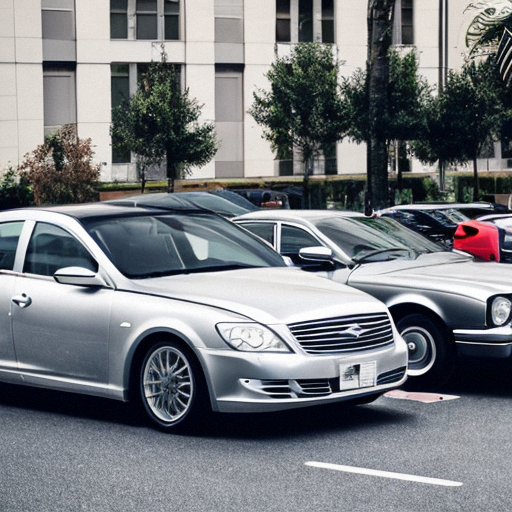}\par\scriptsize $\sigma=2$\end{minipage} &
\begin{minipage}[c]{3cm}\centering\includegraphics[width=3cm]{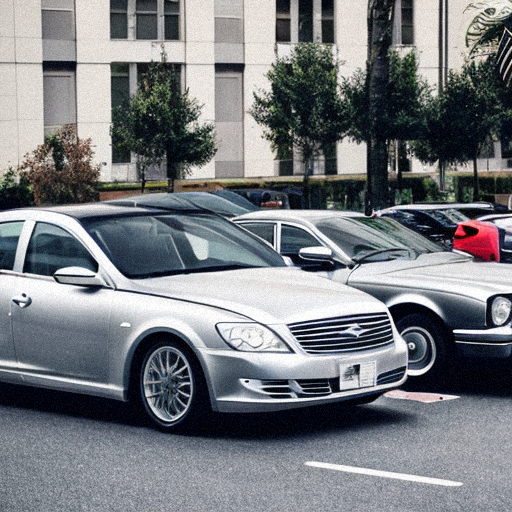}\par\scriptsize $\sigma=6$\end{minipage} \\[8pt]
\specialrule{0.3pt}{0pt}{6pt}
\parbox[c]{2.1cm}{Brightness} &
\begin{minipage}[c]{3cm}\centering\includegraphics[width=3cm]{robustness_example_selected_original.png}\par\vspace{10pt}\end{minipage} &
\begin{minipage}[c]{3cm}\centering\includegraphics[width=3cm]{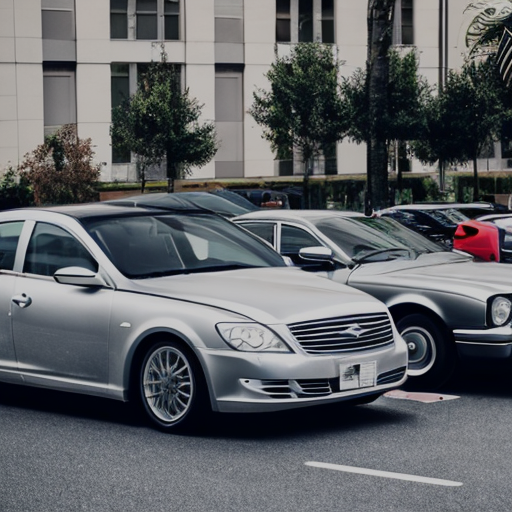}\par\scriptsize $\times0.8$\end{minipage} &
\begin{minipage}[c]{3cm}\centering\includegraphics[width=3cm]{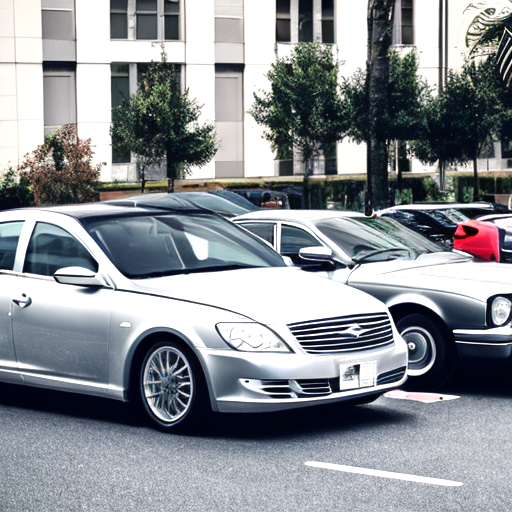}\par\scriptsize $\times1.2$\end{minipage} \\[8pt]
\specialrule{0.3pt}{0pt}{6pt}
\parbox[c]{2.1cm}{Contrast} &
\begin{minipage}[c]{3cm}\centering\includegraphics[width=3cm]{robustness_example_selected_original.png}\par\vspace{10pt}\end{minipage} &
\begin{minipage}[c]{3cm}\centering\includegraphics[width=3cm]{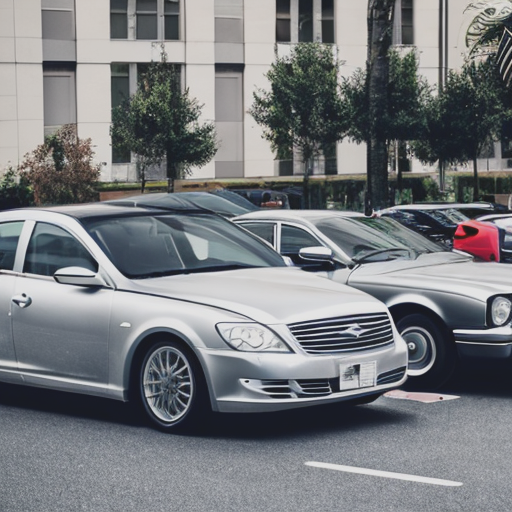}\par\scriptsize $\times0.8$\end{minipage} &
\begin{minipage}[c]{3cm}\centering\includegraphics[width=3cm]{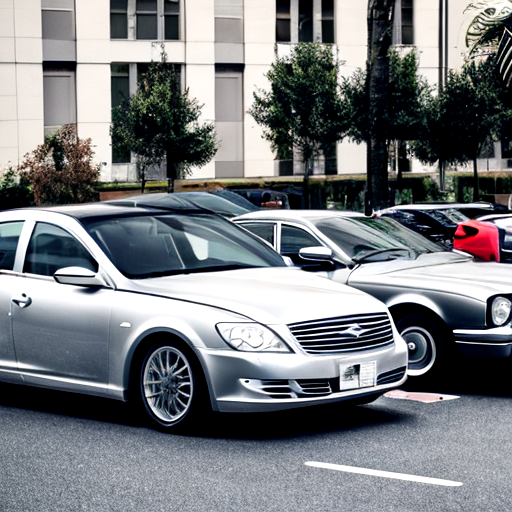}\par\scriptsize $\times1.2$\end{minipage} \\[8pt]
\bottomrule
\end{tabular}
\endgroup

\caption{Examples of ordinary edits, grouped by family across three pages:
appearance and compression, geometric transformations, and edits from an image-editing model.
Each row pairs the original image with its edited version or versions;
parameter values are shown below the images where applicable.}
\label{fig:ai_edit_examples}
\end{figure}

\begin{figure}[p]
\centering
\textbf{Ordinary-edit examples (2/3)}\par\bigskip
\begingroup
\small
\setlength{\tabcolsep}{5pt}
\begin{tabular}{@{}lccc@{}}
\toprule
Family & Original & Setting 1 & Setting 2 \\
\midrule
\multicolumn{4}{l}{\textbf{Appearance and compression}} \\[6pt]
\parbox[c]{2.1cm}{Median filter} &
\begin{minipage}[c]{3cm}\centering\includegraphics[width=3cm]{robustness_example_selected_original.png}\par\vspace{10pt}\end{minipage} &
\begin{minipage}[c]{3cm}\centering\includegraphics[width=3cm]{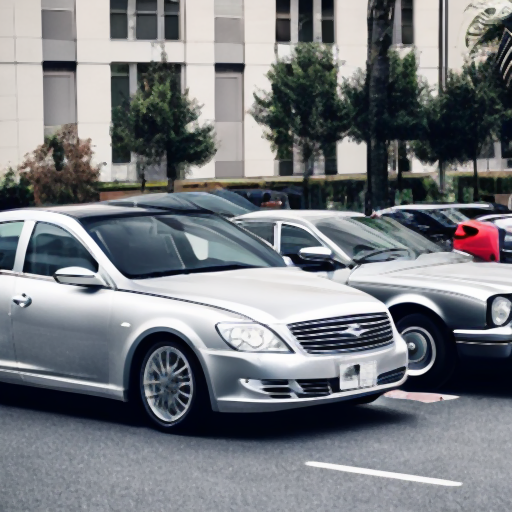}\par\scriptsize $3\times3$\end{minipage} &
\begin{minipage}[c]{3cm}\centering\includegraphics[width=3cm]{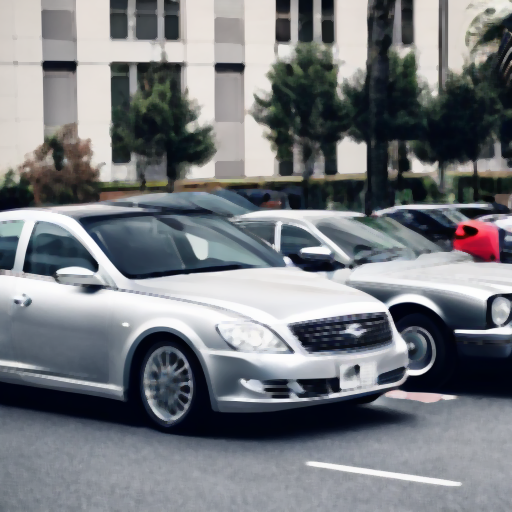}\par\scriptsize $5\times5$\end{minipage} \\[8pt]
\specialrule{0.3pt}{0pt}{6pt}
\parbox[c]{2.1cm}{Salt-and-pepper} &
\begin{minipage}[c]{3cm}\centering\includegraphics[width=3cm]{robustness_example_selected_original.png}\par\vspace{10pt}\end{minipage} &
\begin{minipage}[c]{3cm}\centering\includegraphics[width=3cm]{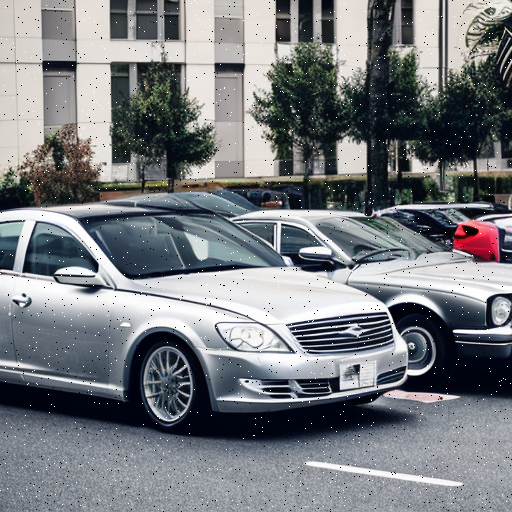}\par\scriptsize $1\%$\end{minipage} &
\begin{minipage}[c]{3cm}\centering\includegraphics[width=3cm]{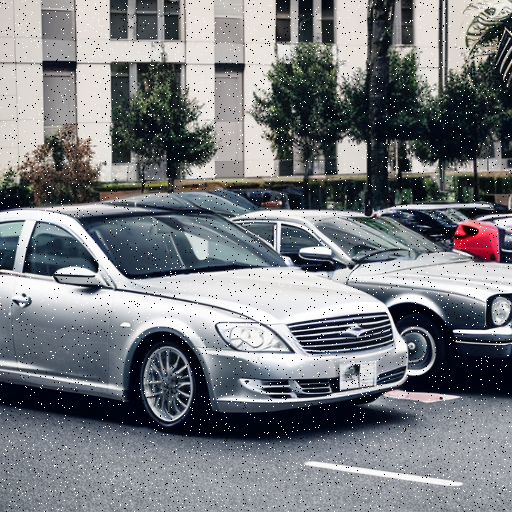}\par\scriptsize $3\%$\end{minipage} \\[8pt]
\specialrule{0.7pt}{0pt}{6pt}
\multicolumn{4}{l}{\textbf{Geometric transformations}} \\[6pt]
\parbox[c]{2.1cm}{Center crop} &
\begin{minipage}[c]{3cm}\centering\includegraphics[width=3cm]{robustness_example_selected_original.png}\par\vspace{10pt}\end{minipage} &
\begin{minipage}[c]{3cm}\centering\includegraphics[width=3cm]{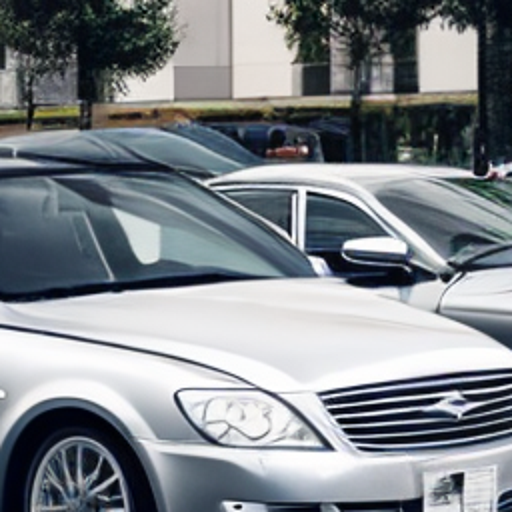}\par\scriptsize $50\%$\end{minipage} &
\begin{minipage}[c]{3cm}\centering\includegraphics[width=3cm]{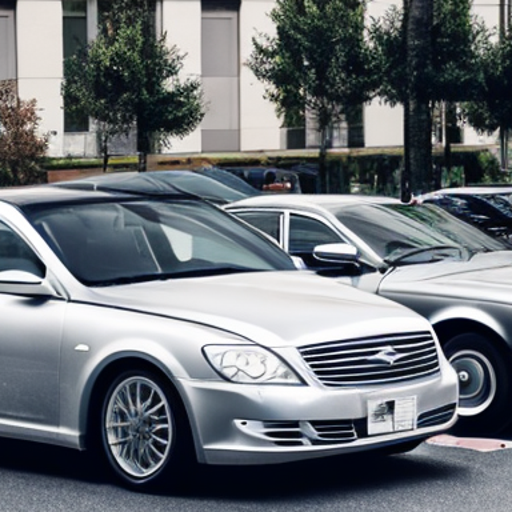}\par\scriptsize $75\%$\end{minipage} \\[8pt]
\specialrule{0.3pt}{0pt}{6pt}
\parbox[c]{2.1cm}{Off-center crop} &
\begin{minipage}[c]{3cm}\centering\includegraphics[width=3cm]{robustness_example_selected_original.png}\par\vspace{10pt}\end{minipage} &
\begin{minipage}[c]{3cm}\centering\includegraphics[width=3cm]{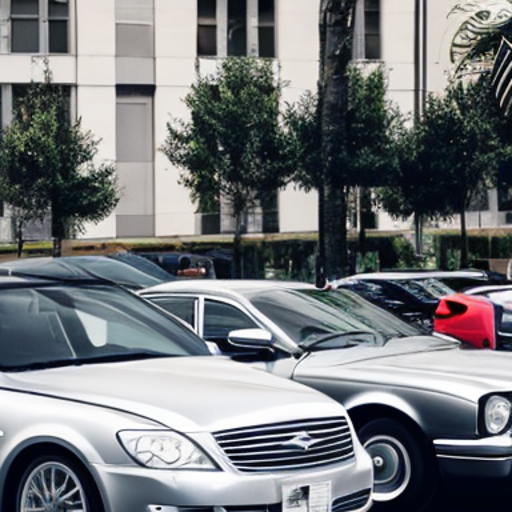}\par\scriptsize $75\%$\end{minipage} &
\begin{minipage}[c]{3cm}\centering\includegraphics[width=3cm]{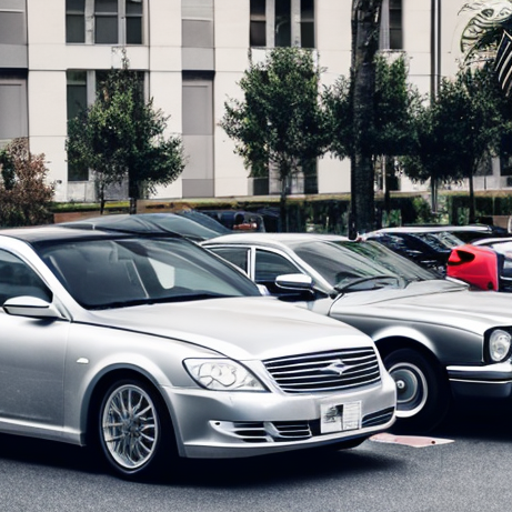}\par\scriptsize $90\%$\end{minipage} \\[8pt]
\specialrule{0.3pt}{0pt}{6pt}
\parbox[c]{2.1cm}{Shift} &
\begin{minipage}[c]{3cm}\centering\includegraphics[width=3cm]{robustness_example_selected_original.png}\par\vspace{10pt}\end{minipage} &
\begin{minipage}[c]{3cm}\centering\includegraphics[width=3cm]{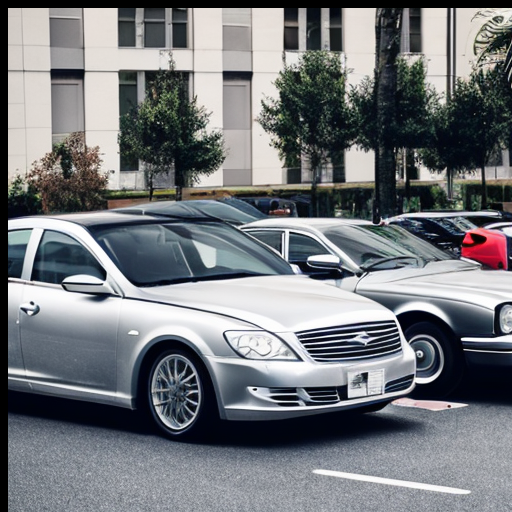}\par\scriptsize 8 px per axis\end{minipage} &
\begin{minipage}[c]{3cm}\centering\includegraphics[width=3cm]{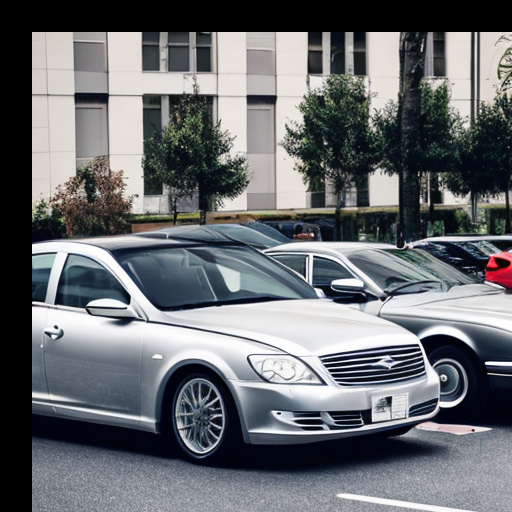}\par\scriptsize 32 px per axis\end{minipage} \\[8pt]
\bottomrule
\end{tabular}
\endgroup

\end{figure}

\begin{figure}[p]
\centering
\textbf{Ordinary-edit examples (3/3)}\par\bigskip
\begingroup
\small
\setlength{\tabcolsep}{5pt}
\begin{tabular}{@{}lcc@{}}
\toprule
Family & Original & Edited\\
\midrule
\multicolumn{3}{l}{\textbf{Edits from an image-editing model}} \\[6pt]
\parbox[c]{2.1cm}{Attribute change} &
\begin{minipage}[c]{4cm}\includegraphics[width=4cm]{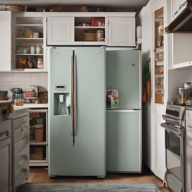}\end{minipage} &
\begin{minipage}[c]{4cm}\includegraphics[width=4cm]{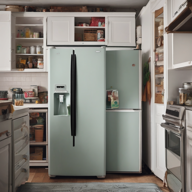}\end{minipage}\\[8pt]
\specialrule{0.3pt}{0pt}{6pt}
\parbox[c]{2.1cm}{Background cleanup} &
\begin{minipage}[c]{4cm}\includegraphics[width=4cm]{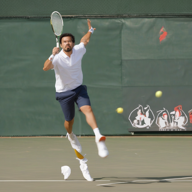}\end{minipage} &
\begin{minipage}[c]{4cm}\includegraphics[width=4cm]{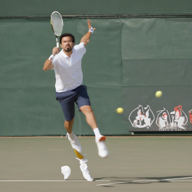}\end{minipage}\\[8pt]
\specialrule{0.3pt}{0pt}{6pt}
\parbox[c]{2.1cm}{Object insertion} &
\begin{minipage}[c]{4cm}\includegraphics[width=4cm]{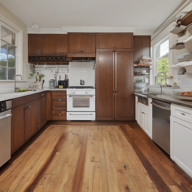}\end{minipage} &
\begin{minipage}[c]{4cm}\includegraphics[width=4cm]{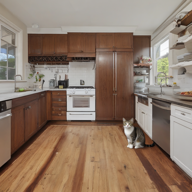}\end{minipage}\\[8pt]
\specialrule{0.3pt}{0pt}{6pt}
\parbox[c]{2.1cm}{Camera-viewpoint change} &
\begin{minipage}[c]{4cm}\includegraphics[width=4cm]{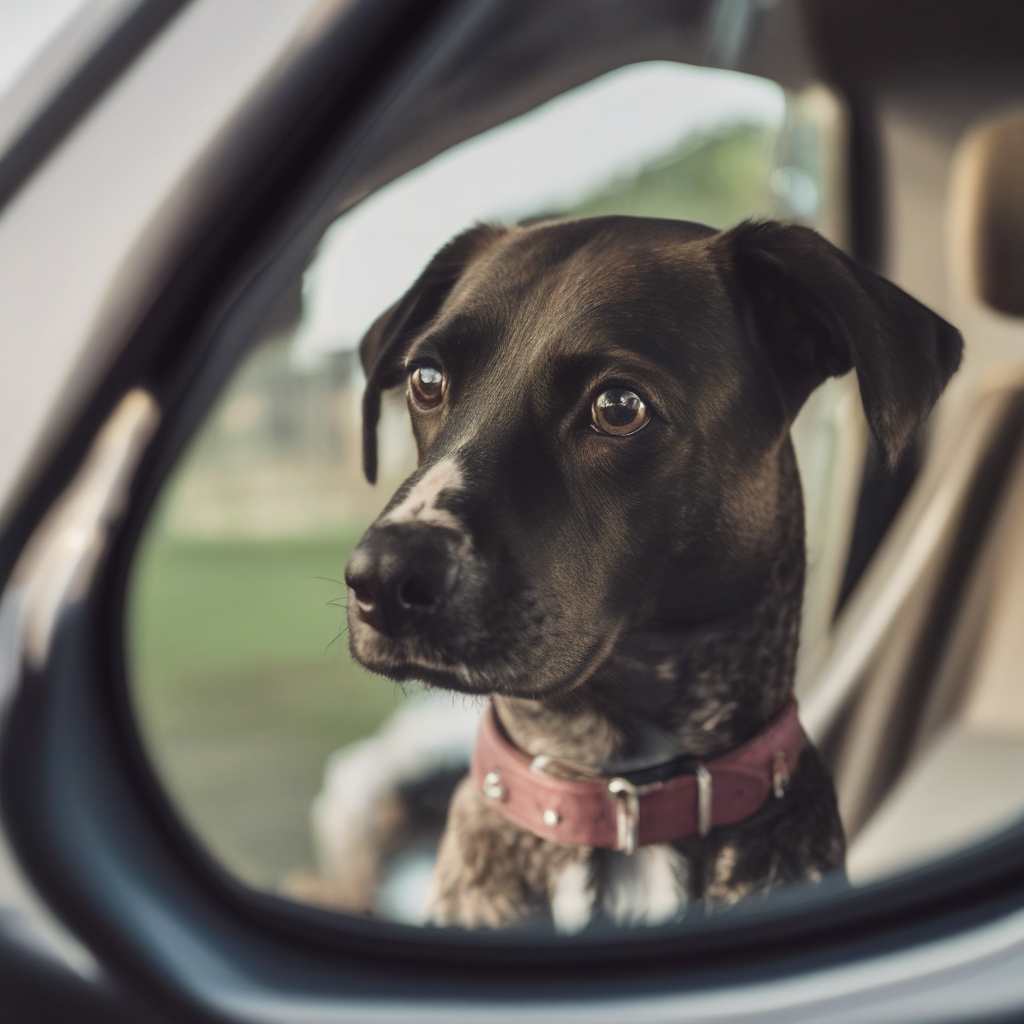}\end{minipage} &
\begin{minipage}[c]{4cm}\includegraphics[width=4cm]{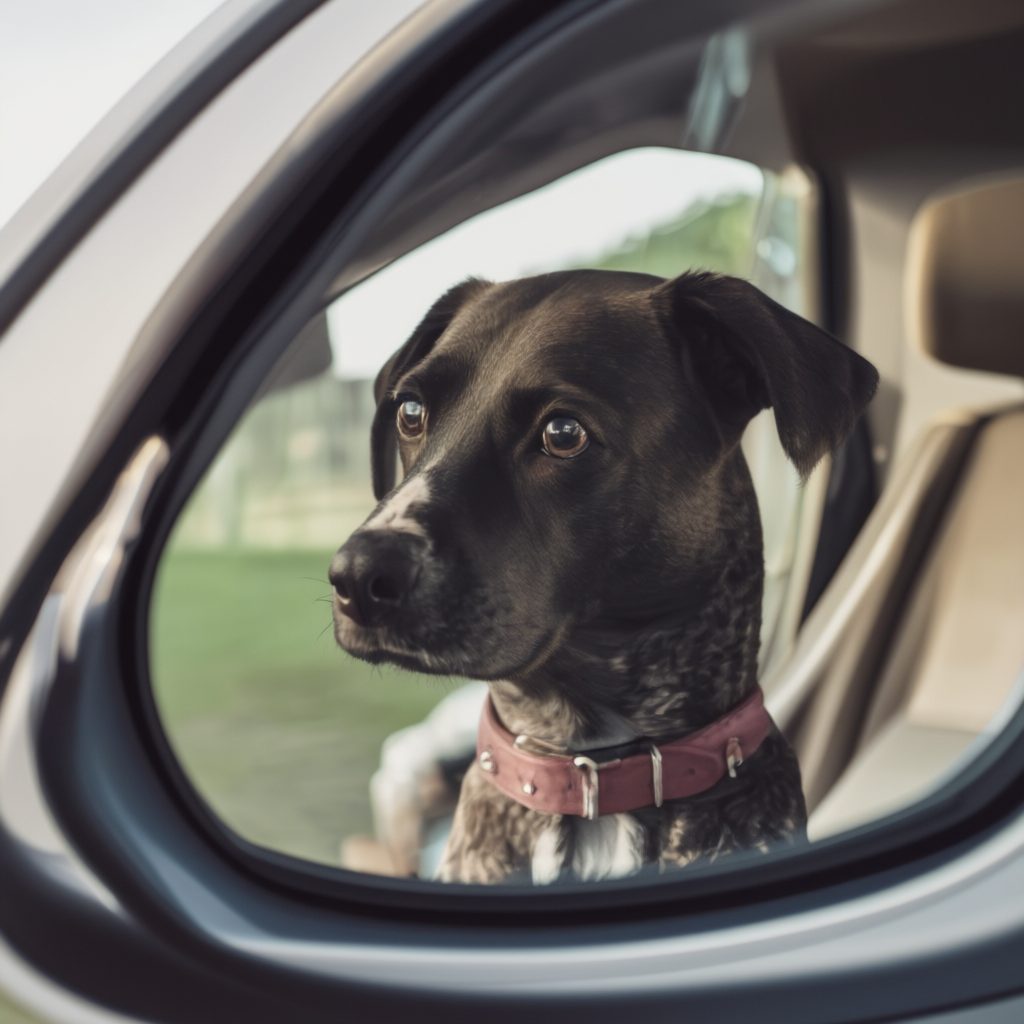}\end{minipage}\\[8pt]
\bottomrule
\end{tabular}
\endgroup

\end{figure}

\clearpage

\clearpage
\section{Additional Results}
\label{app:additional_results}

\subsection{Encoder Similarity and Watermark Retention}
\label{app:encoder_similarity_retention}

We examine whether transformations that remain close to the original image in
our learned representation are also more likely to preserve the watermark. We
use the same 20 edit settings and configuration-specific detector settings as
the robustness evaluation. For each method and edit setting, we pair the mean
original--edited cosine similarity in our embedding with paired detection
retention. Detection thresholds are calibrated on clean images and frozen
across edits.

For our method, the Spearman correlations are $0.9880$, $0.9985$, and $0.9970$
for $(N,M)=(8,4)$, $(16,8)$, and $(32,12)$, respectively. The corresponding
correlations are $0.7471$ for Tree-Ring, $0.7572$ for Gaussian Shading, and
$0.7785$ for SEAL. The stronger association for our method is consistent with
its design: images that remain close in the encoder space are more likely to
retain the same LSH code and clean keyed score. For the baselines, our embedding
serves only as an external diagnostic and does not define their watermarking
mechanisms.

\begin{figure}[H]
\centering
\includegraphics[width=\textwidth]{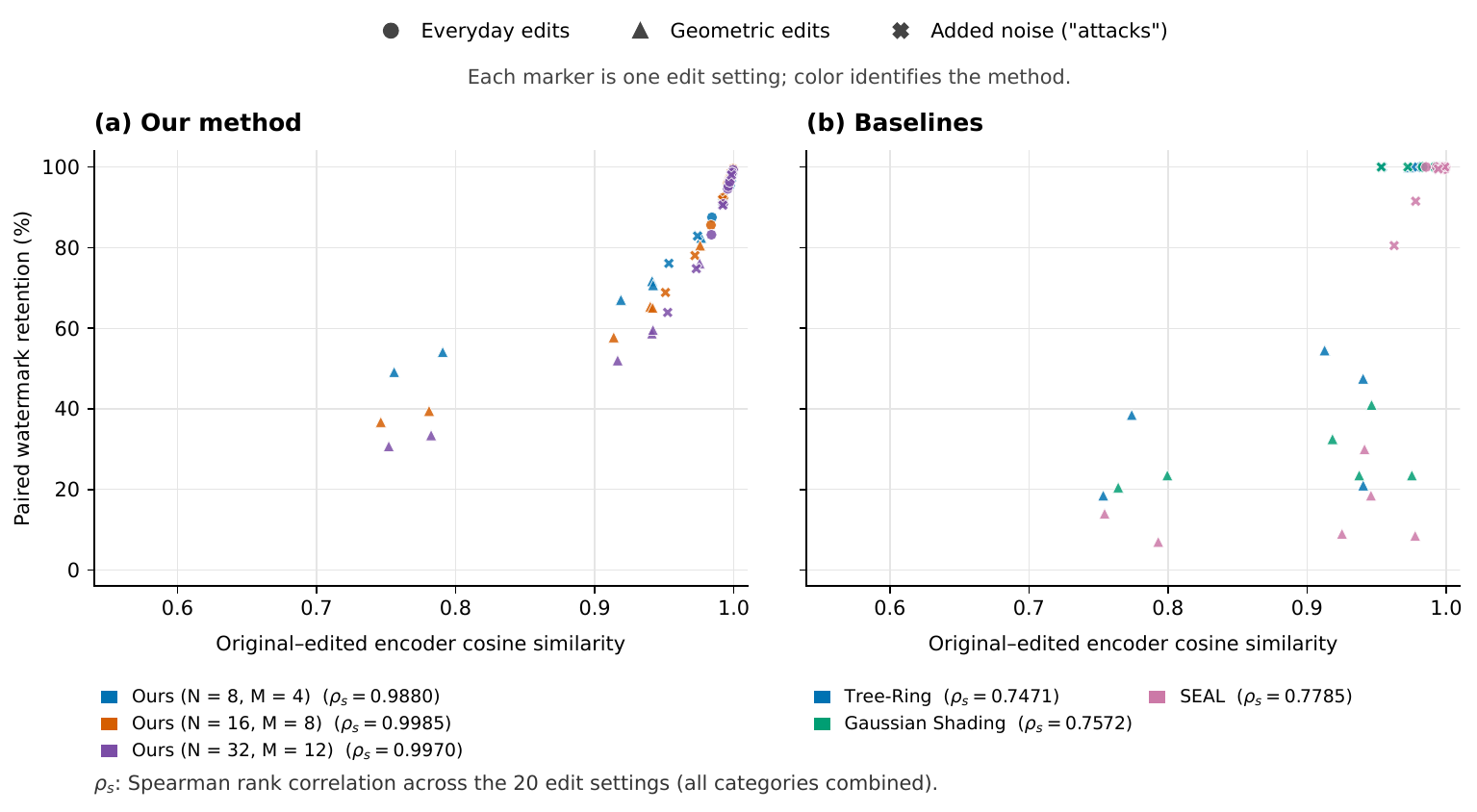}
\caption{\textbf{Encoder similarity and watermark retention.} Each point pairs
mean original--edited cosine similarity in our learned embedding with paired
detection retention for one of 20 edit settings. Marker shape indicates the
edit category, and color indicates the method. Spearman rank correlations are
computed separately for each method. Detection thresholds are calibrated on
clean images and frozen across edits.}
\label{fig:robustness_geometry_diagnostic}
\end{figure}

\clearpage
\subsection{Semantic Encoder Ablation for Clean Detection}
\label{app:encoder_detection_ablation}

To isolate the effect of the semantic encoder, we evaluate our learned encoder,
DINOv2-base, CLIP ViT-B/32, DINOv3 L/16, and TPIPS-Embed (Qwen3VL-8B).
Using the same 1,000 held-out COCO-2017 prompts and Stable Diffusion~2.1, we
rerun our watermarking scheme and clean-image detection separately for each
encoder. We evaluate $(N,M)\in\{(8,4),(16,8),(32,12)\}$ with $B=1$.

Each encoder uses normalized embeddings in its native dimension. We average
over the Cartesian product of five random-hyperplane frames and five PRF keys,
giving 25 replicates per encoder and configuration. We use the same
configuration-specific detector settings as the main clean-detection
evaluation: clean activation targets of $10\%$, $50\%$, and $75\%$ for
$(N,M)=(8,4)$, $(16,8)$, and $(32,12)$, respectively, with Hamming radius
one and five-fold clean-only calibration.

\begin{figure}[H]
\centering
\includegraphics[width=\linewidth]{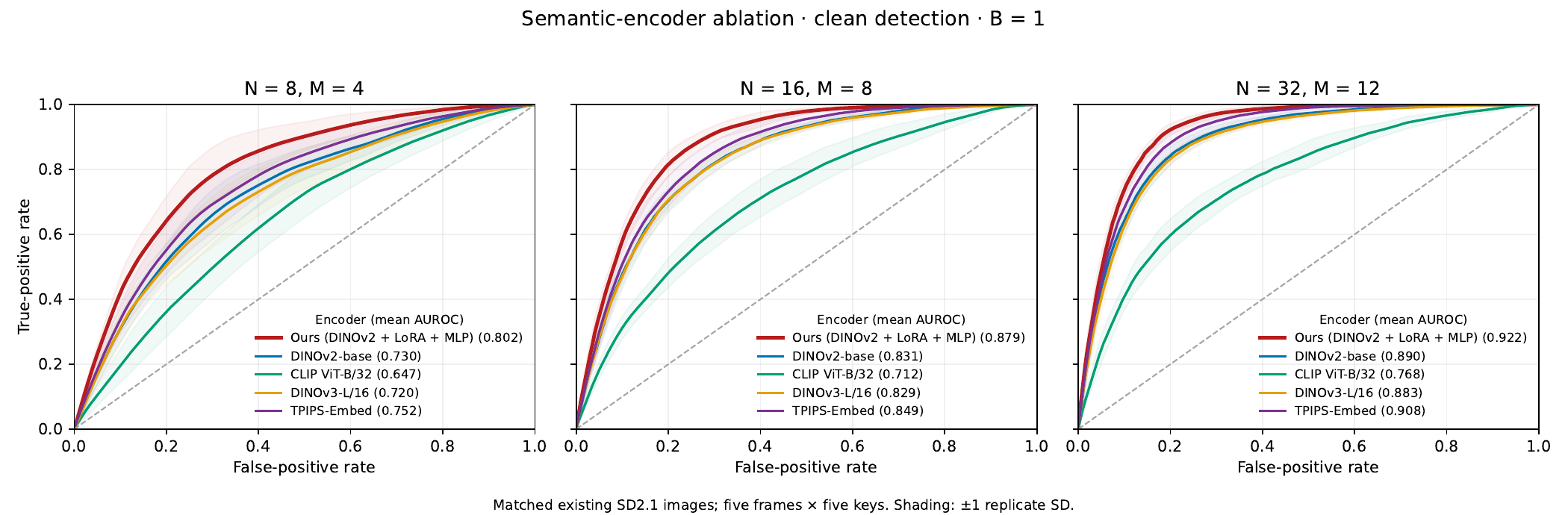}
\caption{\textbf{Semantic-encoder ablation for clean watermark detection.}
Mean ROC curves at $B=1$ for $(N,M)=(8,4)$, $(16,8)$, and $(32,12)$.
Using the same 1,000 held-out COCO-2017 prompts and Stable Diffusion~2.1,
we rerun our watermarking scheme and clean-image detection separately for
each encoder. The dotted diagonal is random guessing (TPR=FPR). Exact
detection settings are given in Appendix~\ref{app:robustness_tables}.}
\label{fig:encoder_detection_ablation}
\end{figure}

Figure~\ref{fig:encoder_detection_ablation} shows that our encoder obtains the
highest mean AUROC in all three configurations: 0.802, 0.879, and 0.922.
TPIPS-Embed is the strongest baseline, with mean AUROCs of 0.752, 0.849,
and 0.908, respectively. At an empirical FPR of at most 0.2, our
corresponding mean TPRs are 0.576, 0.808, and 0.921.

\clearpage
\subsection{Alternative Distortion-Free Sampling Rules}
\label{app:alternative_sampling_rules}

Several distortion-free sampling rules can use the same candidate images and
semantic codes. We compare representative sampling
(Algorithm~\ref{alg:sampling}), our deployed Gumbel rule in
Algorithm~\ref{alg:gumbel_sampling}~\citep{kuditipudi2024robust}, and
non-distortionary Tournament sampling~\citep{dathathri2024scalable}.
All three rules are distortion-free, but they use collisions among candidate
codes differently.

To isolate the effect of the selection rule, we replay all three rules on the
same Stable Diffusion~2.1 candidate pools and cached embeddings, keeping the
candidate images and semantic representations fixed. We evaluate
\((N,M)\in\{(8,4),(16,8),(32,12)\}\) on 1,000 held-out COCO-2017 prompts, with
\(B=1\) for single-image detection. Results average five random-hyperplane
frames and five keys. Each rule uses the same detector configuration and
clean-image threshold calibration, with its corresponding keyed score.

Table~\ref{tab:alternative_sampling_rules} shows that Gumbel sampling improves
AUROC and TPR in all three configurations. At \((N,M)=(32,12)\), for example,
AUROC increases from \(0.872\) to \(0.920\), and TPR@20\% increases from
\(0.852\) to \(0.919\). This advantage also persists after editing: across all
51 evaluated transformations, attacked TPR increases from \(0.679\) to
\(0.726\), while conditional retention remains nearly unchanged
(\(0.787\) versus \(0.792\)). Tournament sampling is weaker than both rules
for \(N\geq16\).

\begin{table}[H]
\centering
\definecolor{samplingbest}{HTML}{007A53}
\definecolor{samplingsecond}{HTML}{C76A00}
\caption{Clean detection under alternative distortion-free sampling rules.
TPR is measured at thresholds calibrated to a 20\% clean FPR target. Results
use \(B=1\) for single-image detection and average five random-hyperplane
frames and five keys. Bold green and orange indicate the best and second-best
values within each row.}
\label{tab:alternative_sampling_rules}
\small
\begin{tabular*}{0.82\linewidth}{@{\extracolsep{\fill}}crrrrrr@{}}
\toprule
& \multicolumn{2}{c}{Representative sampling} &
\multicolumn{2}{c}{Gumbel} & \multicolumn{2}{c}{Tournament} \\
\cmidrule(lr){2-3}\cmidrule(lr){4-5}\cmidrule(l){6-7}
\((N,M)\) & AUROC & TPR & AUROC & TPR & AUROC & TPR \\
\midrule
\((8,4)\) & 0.734 & \textcolor{samplingsecond}{\textbf{0.500}} &
\textcolor{samplingbest}{\textbf{0.802}} & \textcolor{samplingbest}{\textbf{0.576}} &
\textcolor{samplingsecond}{\textbf{0.746}} & 0.454 \\
\((16,8)\) & \textcolor{samplingsecond}{\textbf{0.821}} &
\textcolor{samplingsecond}{\textbf{0.733}} & \textcolor{samplingbest}{\textbf{0.879}} &
\textcolor{samplingbest}{\textbf{0.808}} & 0.762 & 0.545 \\
\((32,12)\) & \textcolor{samplingsecond}{\textbf{0.872}} &
\textcolor{samplingsecond}{\textbf{0.852}} & \textcolor{samplingbest}{\textbf{0.920}} &
\textcolor{samplingbest}{\textbf{0.919}} & 0.756 & 0.532 \\
\bottomrule
\end{tabular*}
\end{table}

These results show that Gumbel selection extracts a stronger detection signal
than representative sampling from the same candidates. We therefore use
Gumbel selection in our experiments, except where sampling rules are explicitly
compared.

\clearpage
\raggedbottom
\subsection{Black-Box Semantic Encoder Access}
\label{app:encoder_dependent_robustness}

One strength of our method is that black-box access to the semantic encoder
allows users to choose a representation suited to their desired notion of
watermark preservation, without changing the watermarking algorithm. We
demonstrate this flexibility using two encoders: CSD~\citep{somepalli2024measuring},
which represents image style, and a depth-based encoder built from Depth
Anything V2 Small~\citep{yang2024depthanythingv2}, which represents spatial
layout.
Both representations are normalized before hashing.

We hypothesize that these representations induce different watermark
invariances. Style edits such as oil painting, watercolor, and pencil sketch
change image appearance while approximately preserving spatial layout. They
should therefore move images farther in CSD's embedding space than in the
depth-based representation, weakening the CSD-based watermark more strongly.
Conversely, translation, cropping, and horizontal flipping alter spatial
arrangement without directly restyling the image. We therefore expect higher
retention for the CSD-based watermark than for the depth-based watermark
under these transformations. These are
empirical predictions: neither encoder guarantees invariance to every
transformation that preserves its intended image property.

To test these predictions, we apply six transformations using the same
watermarking algorithm and evaluation protocol. We use the first 100 prompts
of the held-out COCO-2017 encoder-ablation panel with Stable Diffusion~2.1,
$(N,M)=(16,8)$, $B=1$, Hamming radius one, and a $50\%$ clean activation
target. We average detection results over five random-hyperplane frames and
five keys. The style edits use
InstructPix2Pix~\citep{brooks2023instructpix2pix} with the instructions
``Make it an oil painting'', ``Turn it into a watercolor painting'', and
``Turn it into a pencil sketch''. The spatial transformations are a
32-pixel translation per axis with black padding, a $75\%$ off-center crop
anchored at $(0.65,0.35)$ and resized to the original resolution, and a
horizontal flip. Candidate selection and boundary calibration use only
unedited embeddings. Detection thresholds target $20\%$ FPR on the 100
unedited clean images and remain fixed after editing; these images are also
used to report clean FPR, rather than forming an independent calibration set.
Mean cosine distance is $1-\langle z,z'\rangle$ for normalized original and
edited embeddings, averaged over all candidates and clean controls.

Table~\ref{tab:encoder_dependent_robustness} reports detection, retention, and
representation changes; Figure~\ref{fig:encoder_dependent_edits} illustrates
the six transformations. The results show contrasting
robustness profiles. Across the three style edits, CSD retains
$25.2$--$36.9\%$ of its initial detections, compared with $86.3$--$94.0\%$
for the depth-based encoder. The spatial transformations reverse this pattern:
CSD retains $55.6$--$89.5\%$, whereas depth retains $23.5$--$25.9\%$. The
corresponding embedding changes follow the same pattern.

These results illustrate how users can choose an encoder according to their
desired notion of watermark preservation. Consistent with
Appendix~\ref{app:encoder_similarity_retention}, transformations that induce smaller
cosine distances in the encoder's representation also retain more watermark
detections. A user seeking robustness to style transfer may choose a
layout-based representation, while a user seeking a
watermark tied to image style may choose CSD, preserving detection under
spatial transformations such as horizontal flipping. The evaluated
transformations demonstrate this flexibility without requiring access to
encoder internals or changes to the watermarking algorithm.

\begingroup
\setlength{\intextsep}{4pt}
\begin{table}[H]
\centering
\small
\setlength{\tabcolsep}{6pt}
\caption{Encoder-dependent watermark robustness. TPR uses thresholds
calibrated at $20\%$ FPR on unedited images and frozen after editing.
Retention is conditional on initial detection; smaller cosine distance
indicates greater representation stability.}
\label{tab:encoder_dependent_robustness}
\begin{tabular}{@{}llrrr@{}}
\toprule
Encoder & Transformation & TPR (\%) & Retention (\%) & Cosine distance $\downarrow$ \\
\midrule
CSD style & Original        & 66.8 & ---  & --- \\
\addlinespace
          & \multicolumn{4}{l}{\textit{Style changes}} \\
          & Oil painting & 28.5 & 36.9 & 0.199 \\
          & Watercolor & 20.8 & 25.2 & 0.327 \\
          & Pencil sketch & 23.0 & 25.8 & 0.452 \\
\addlinespace
          & \multicolumn{4}{l}{\textit{Spatial transformations}} \\
          & Translation, 32 px & 44.6 & \textbf{63.6} & 0.041 \\
          & Off-center crop, $75\%$ & 38.9 & \textbf{55.6} & 0.059 \\
          & Horizontal flip & 60.6 & \textbf{89.5} & 0.008 \\
\midrule
Depth layout & Original        & 64.0 & ---  & --- \\
\addlinespace
             & \multicolumn{4}{l}{\textit{Style changes}} \\
          & Oil painting & 60.7 & \textbf{94.0} & 0.009 \\
          & Watercolor & 57.4 & \textbf{88.5} & 0.013 \\
          & Pencil sketch & 56.2 & \textbf{86.3} & 0.021 \\
\addlinespace
             & \multicolumn{4}{l}{\textit{Spatial transformations}} \\
          & Translation, 32 px & 21.3 & 24.4 & 0.504 \\
          & Off-center crop, $75\%$ & 19.3 & 23.5 & 0.402 \\
          & Horizontal flip & 19.9 & 25.9 & 0.418 \\
\bottomrule
\end{tabular}
\end{table}

\begin{figure}[H]
\centering
\includegraphics[width=0.69\linewidth]{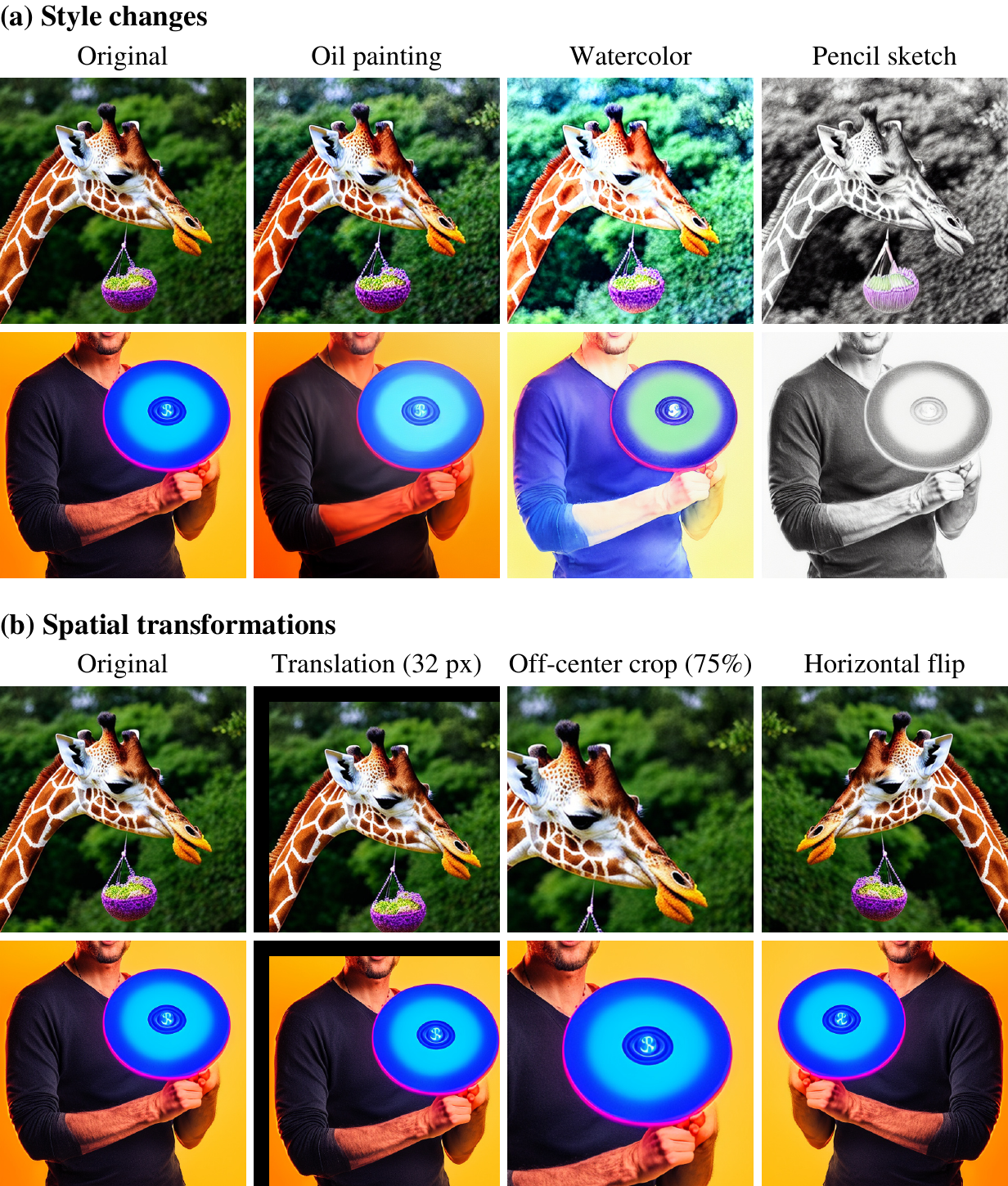}
\caption{\textbf{Transformations used to test encoder-dependent robustness.}
Top: original, oil painting, watercolor, and pencil sketch. Bottom: original,
32-pixel translation, $75\%$ off-center crop, and horizontal flip.}
\label{fig:encoder_dependent_edits}
\end{figure}
\endgroup

\clearpage
\flushbottom
\subsection{Closed-API Black-Box Generator Access}
\label{app:closed_api_generation}

To demonstrate that our sampling-only method can watermark outputs from a
closed-API model without access to its internals, we use OpenAI's GPT Image API
(``\texttt{gpt-image-2.5-sunburst}'') strictly as a black-box generator. We use
the first 300 held-out COCO-2017 prompts.
For each prompt, we independently request eight watermark candidates
and one clean control at low quality and $1024\times1024$ resolution, for 2700 API generations in total.
We use $(N,M)=(8,8)$, exact-code detection ($r_{\mathrm{ham}}=0$), five independently sampled
choices of the random-hyperplane LSH $H_{\mathcal R}$, and five PRF keys. Reported curves and
metrics average the resulting 25 key-hash repetitions.

\begin{figure}[!ht]
\centering
\includegraphics[width=\linewidth]{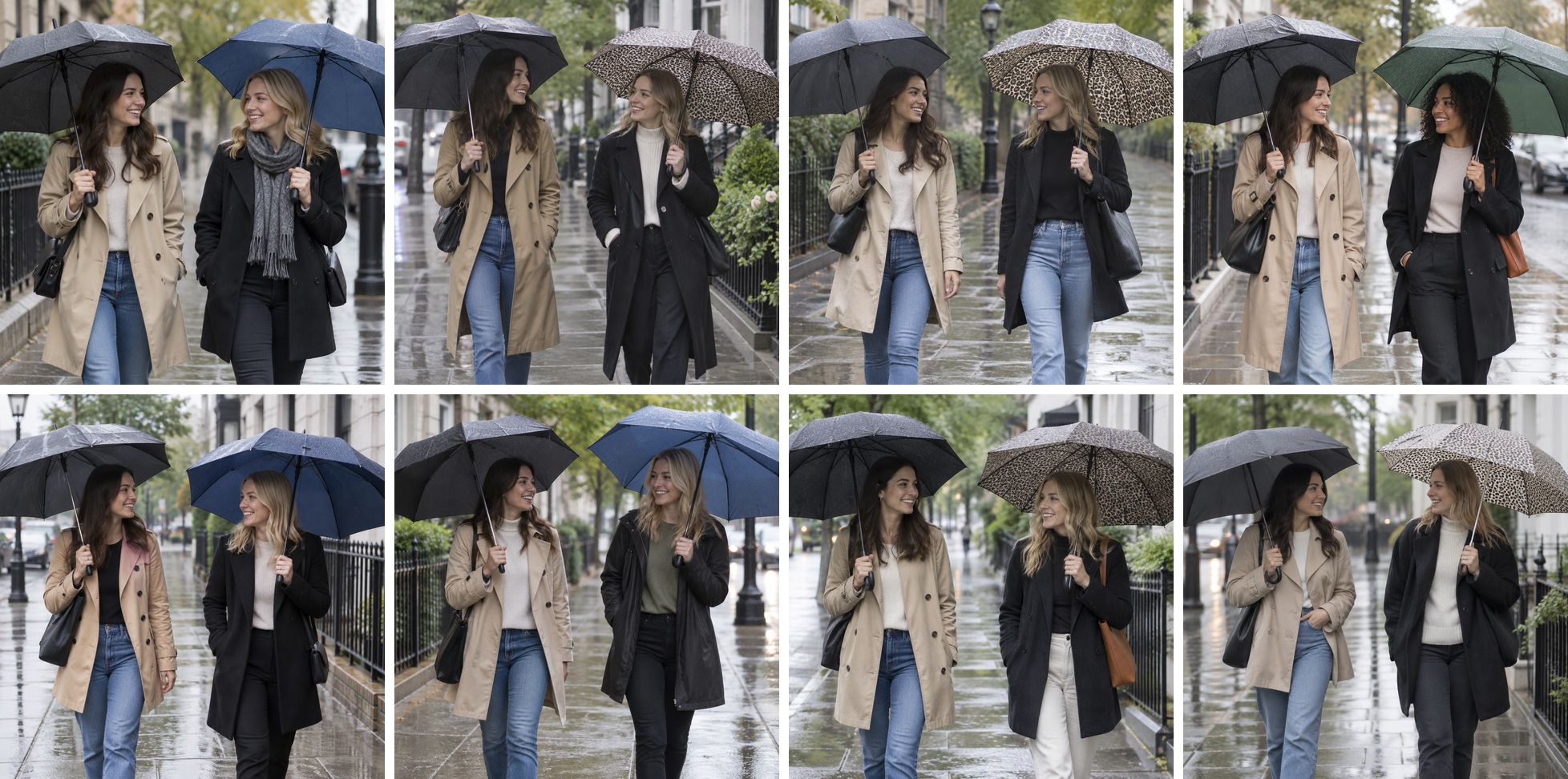}
\caption{\textbf{Eight independent GPT Image generations from one prompt.}
All images were returned by separate API calls for ``Two women walk outside,
both holding up umbrellas.'' Although their layouts are more similar than in
the open-model generation panels, our encoder maps them to a mean of $6.6$
distinct 8-bit codes across five independently sampled random-hyperplane LSH functions.}
\label{fig:closed_api_grid}
\end{figure}

GPT Image's candidates are more clustered under our learned encoder than the
open-model generations in Table~\ref{tab:encoder_comparison}: their mean
within-prompt cosine similarity is $0.573$, compared with $0.222$ for
SD2.1. GPT Image's eight candidates occupy $5.419$ distinct 8-bit
codes on average across the 300 prompts and five choices of $H_{\mathcal R}$, leaving
substantial variation for keyed selection. Figure~\ref{fig:closed_api_grid}
illustrates this behavior for the prompt ``Two women walk outside, both holding
up umbrellas.'' Its eight candidates occupy a mean of $6.6/8$ distinct codes
across the five choices of $H_{\mathcal R}$.

For individual images $B=1$, the mean AUROC is 0.815 and the mean TPR at a nominal 20\% FPR is 64.6\%.
For batches of five $B=5$, we divide 300 prompts into 60 groups.
Each group gives us a clean batch and a watermarked batch generated from the same five prompts.
We average the five detection scores in each batch.
The mean AUROC is 0.978, and the mean TPR at a nominal 20\% FPR is 98.0\%.
The figure shows both ROC curves.

\begin{figure}[!ht]
\centering
\includegraphics[width=0.72\linewidth]{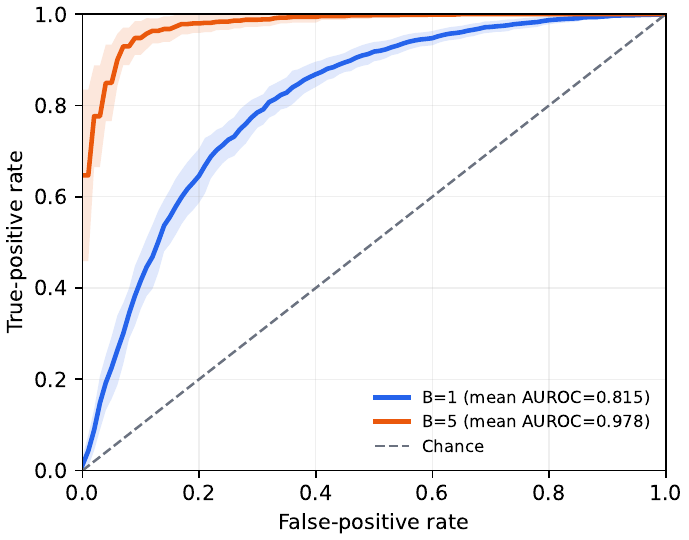}
\caption{\textbf{Closed-API watermark detection.} Mean ROC curves for GPT
Image with $(N,M)=(8,8)$, averaged over five keys and five choices of $H_{\mathcal R}$.
Shading shows one standard deviation across the 25 repetitions. For $B=5$,
each observation averages five watermarked or clean scores with identical
prompt membership.}
\label{fig:closed_api_roc}
\end{figure}

\FloatBarrier

\subsection{Text-to-Audio Watermarking}
\label{app:audio_watermark}

To demonstrate our sampling procedure in another generative domain, we apply
Algorithm~\ref{alg:gumbel_sampling} to text-to-audio generation with AudioLDM~2~\citep{liu2023audioldm2}.
We generate 32 independent 10-second clips for each of 200 prompts from the
AudioCaps test set~\citep{kim2019audiocaps} and encode them with LAION-CLAP~\citep{wu2023laionclap}.
CLAP learns to align audio with text descriptions, rather than to separate
independent clips generated from the same prompt or preserve their hash codes
under audio processing. Nevertheless, its embeddings provide enough code
variation for keyed selection and detection. We evaluate the same three
$(N,M)$ configurations as in our image experiments, using a separately
generated clean clip per prompt, five keys, and five hyperplane choices.
Each prompt provides a single candidate pool, so for $B>1$ the detector
averages the scores of $B$ watermarked clips generated from $B$ different
prompts, and likewise for clean clips.

\begin{table}[H]
\centering
\small
\setlength{\tabcolsep}{6pt}
\caption{\textbf{Audio watermark detection.} Means over 25 key--hyperplane
repetitions on 200 prompts. For $B>1$, each batch contains clips generated
from $B$ different prompts. We calibrate the detector on clean clips and
report the true-positive rate at a nominal $20\%$ clean false-positive rate.}
\label{tab:audio_watermark_detection}
\begin{tabular}{@{}lcccccc@{}}
\toprule
 & \multicolumn{2}{c}{$B=1$} & \multicolumn{2}{c}{$B=5$} & \multicolumn{2}{c}{$B=10$} \\
\cmidrule(lr){2-3}\cmidrule(lr){4-5}\cmidrule(l){6-7}
$(N,M)$ & AUROC & TPR@20\% & AUROC & TPR@20\% & AUROC & TPR@20\% \\
\midrule
$(8,4)$   & 0.711 & 0.397 & 0.903 & 0.846 & 0.964 & 0.956 \\
$(16,8)$  & 0.817 & 0.670 & 0.978 & 0.983 & 0.998 & 1.000 \\
$(32,12)$ & 0.862 & 0.782 & 0.990 & 0.993 & 1.000 & 1.000 \\
\bottomrule
\end{tabular}
\end{table}

\begin{figure}[H]
\centering
\includegraphics[width=\linewidth]{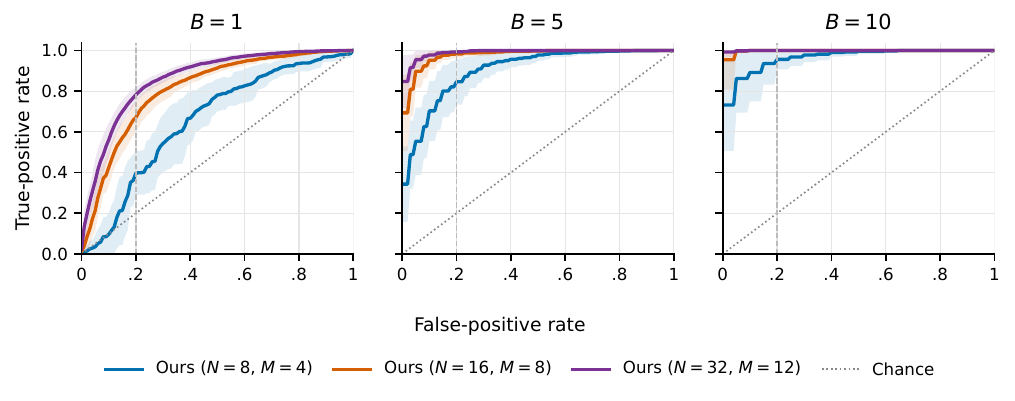}
\caption{\textbf{Audio watermark ROC curves.} Mean clean-detection ROC over
25 key--hyperplane repetitions on 200 prompts for $B=1$, $5$, and $10$ clips
per decision; shaded bands show one standard deviation across repetitions.
The dashed line marks the $20\%$ false-positive rate of
Table~\ref{tab:audio_watermark_detection}. With 200 prompts, $B=10$ yields
20 batches per repetition, so its false-positive rate moves in steps of 0.05.}
\label{fig:audio_watermark_roc}
\end{figure}

Table~\ref{tab:audio_watermark_detection} and Figure~\ref{fig:audio_watermark_roc}
show that single-clip detection improves across the evaluated configurations,
although it remains weaker than in our image experiments. Aggregating evidence
over clips strengthens it substantially: with $B=10$, AUROC reaches $0.964$,
$0.998$, and $1.000$ for the three configurations. We also apply MP3 compression
at 128 kbps, Gaussian noise at 30 dB SNR, a 3 dB reduction in gain, and a 100 ms
shift, keeping the clean-calibrated thresholds fixed. At $(N,M)=(32,12)$, these
transformations retain $56.2\%$, $59.2\%$, $60.7\%$, and $44.5\%$ of initially
detected clips, respectively. Even the gain reduction preserves the complete
12-bit CLAP code for only $36\%$ of selected clips. Thus, the same selection
procedure yields a detectable audio watermark, while this encoder's code
instability limits robustness to benign audio processing.

\FloatBarrier

\clearpage
\subsection{Generation Quality and Prompt Alignment}
\label{app:quality_alignment_table}

To assess generation quality and prompt alignment, we evaluate Stable
Diffusion~2.1 Base at a resolution of \(512 \times 512\). We generate one image
for each of 5,000 COCO-2017 prompts and measure FID~\citep{heusel2017gans} against 5,000 COCO-2017
validation images. We measure OpenCLIP similarity~\citep{cherti2023reproducible} separately on 1,000 prompts.
For every watermarking configuration, we generate an unwatermarked reference set
using the same model, prompts, and sampling settings.

Table~\ref{tab:experiment3_fid} reports generation quality and prompt alignment.
Across our three configurations, FID improves by \(0.38\)--\(1.82\%\) relative to
matched clean generation, while OpenCLIP similarity changes by at most
\(0.03\%\). Tree-Ring and Gaussian Shading change FID by \(-0.16\%\) and
\(+0.05\%\), respectively, while SEAL changes it by \(+1.78\%\). Relative to
their matched clean reference sets, the watermarked populations have comparable
FID to real COCO images and nearly unchanged image-prompt similarity. These
complementary diagnostics are consistent with the direct matched-population FID
results in the main text.

\begin{table}[H]
\centering
\small
\setlength{\tabcolsep}{8pt}
\caption{Distributional quality on 5,000 unique COCO-2017 validation prompts and image--prompt alignment on 1,000 prompts. Each method is compared with its matched clean control. Relative change is $100(s_{\rm wm}/s_{\rm clean}-1)$. This table supports the quality claim in Section~\ref{sec:exp_quality_diversity}.}
\label{tab:experiment3_fid}
\begin{tabular}{@{}lccc@{}}
\toprule
Method & Watermarked & Matched clean & Relative change \\
\midrule
\multicolumn{4}{@{}l}{\textit{FID (lower is better)}} \\
Ours \((N,M)=(8,4)\) & 26.840 & 26.943 & $-0.38\%$ \\
Ours \((N,M)=(16,8)\) & 26.453 & 26.943 & $-1.82\%$ \\
Ours \((N,M)=(32,12)\) & 26.615 & 26.943 & $-1.22\%$ \\
Tree-Ring            & 24.853 & 24.894 & $-0.16\%$ \\
Gaussian Shading     & 24.907 & 24.894 & $+0.05\%$ \\
SEAL                 & 25.649 & 25.201 & $+1.78\%$ \\
\addlinespace
\multicolumn{4}{@{}l}{\textit{OpenCLIP image prompt similarity (higher is better)}} \\
Ours \((N,M)=(8,4)\) & 0.3154 & 0.3154 & $\phantom{+}0.00\%$ \\
Ours \((N,M)=(16,8)\) & 0.3155 & 0.3155 & $-0.01\%$ \\
Ours \((N,M)=(32,12)\) & 0.3156 & 0.3157 & $-0.03\%$ \\
Tree-Ring            & 0.3144 & 0.3145 & $-0.03\%$ \\
Gaussian Shading     & 0.3155 & 0.3145 & $+0.31\%$ \\
SEAL                 & 0.3138 & 0.3140 & $-0.06\%$ \\
\bottomrule
\end{tabular}
\end{table}

\FloatBarrier
\clearpage
\subsection{Detection under Mixed-Source Batches}
\label{app:mixed_source_batches}

We test how our detector performs in real-world batches that mix watermarked
and unwatermarked images or images from different watermarking sources, and
whether it provides a signal for distinguishing those sources.
We start with ten images watermarked under the first key and progressively
replace them, one image at a time, with images watermarked under a second,
independent key. At each mixture, we test the batch under both keys. We also
repeat the replacement procedure with unwatermarked images and evaluate detection
under the first key.

We use Stable Diffusion~2.1 candidates for 1,000 COCO-2017 prompts and apply
Algorithm~\ref{alg:gumbel_sampling} with the original encoder and detector settings.
We evaluate all three configurations over five hyperplane frames and five key
pairs. Both keys share the encoder and hyperplane frame. We calibrate thresholds
on unwatermarked batches at nominal FPRs of 1\%, 5\%, and 20\%, then keep them
fixed across mixtures.

\begin{figure}[p]
\centering
\includegraphics[width=\linewidth,height=0.82\textheight,keepaspectratio]{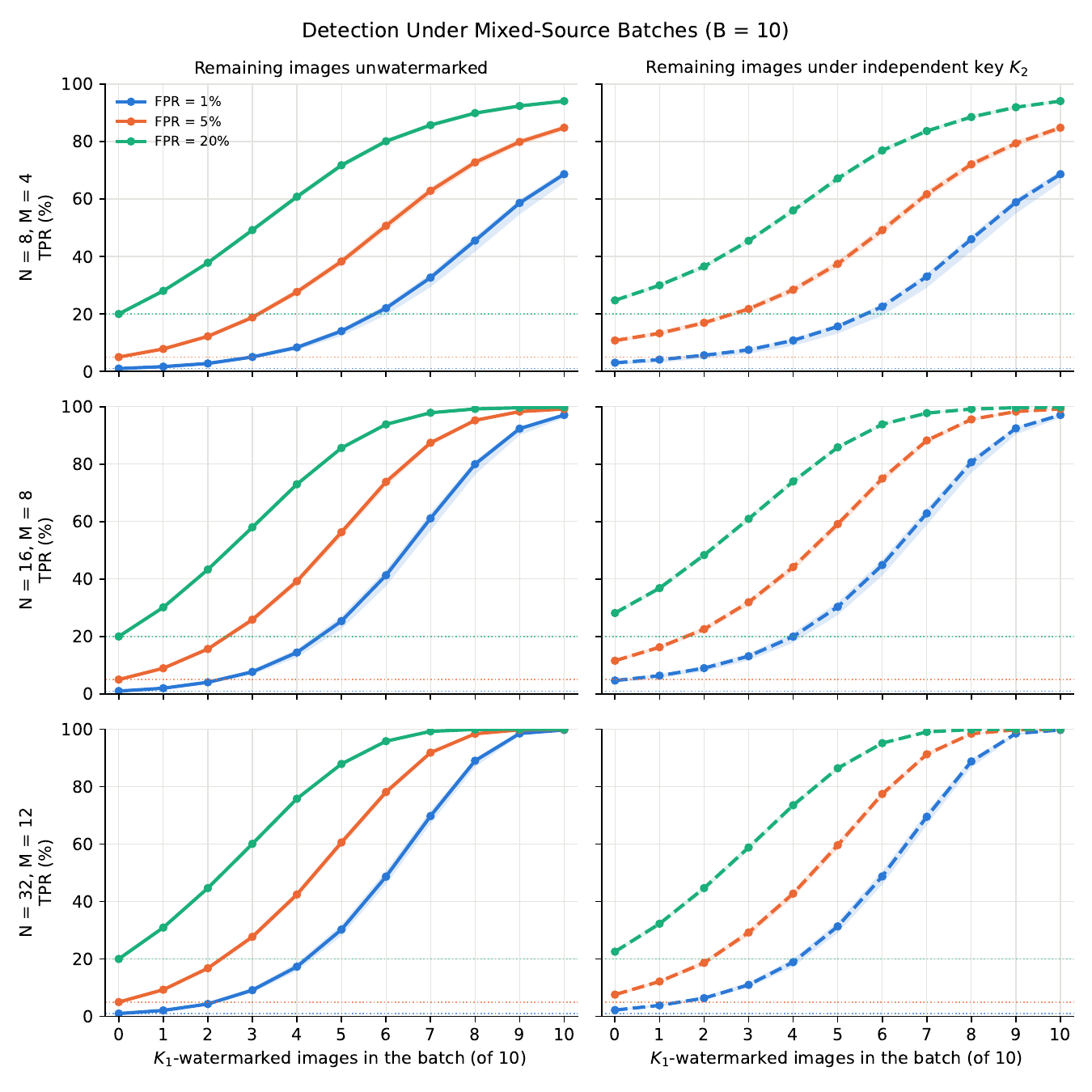}
\caption{\textbf{Detection under mixed-source batches.} We vary the number
of $K_1$ images in a ten-image batch; the rest are unwatermarked (left) or
carry $K_2$ (right). Colors denote nominal clean-calibrated FPR. Shading shows
95\% intervals from 1,000 prompt bootstrap resamples with thresholds recomputed
and keys and frames fixed. Resampling ties can bias calibrated quantities
slightly downward. At zero $K_1$ images, detection rates measure false alarms.}
\label{fig:mixed_source_batches}
\end{figure}

Figure~\ref{fig:mixed_source_batches} shows that detection decreases gradually
as we replace images carrying the tested key with unwatermarked images.
For $(N,M)=(32,12)$, batches containing five watermarked and five unwatermarked
images achieve detection rates of 30.2\%, 60.6\%, and 87.9\% at the three
operating points, respectively. Replacing the five unwatermarked images with
images carrying $K_2$ gives similar rates of 31.3\%, 59.6\%, and 86.4\%.
Detection therefore remains informative even when only half the batch carries
the tested watermark.

\begin{figure}[!ht]
\centering
\includegraphics[width=0.75\linewidth]{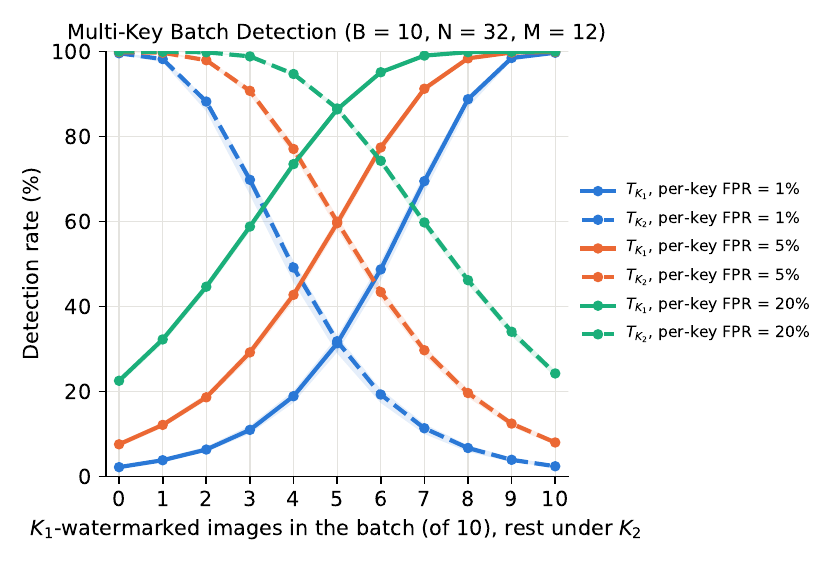}
\caption{\textbf{Detection under two known keys.} We score the same mixed
batches with $K_1$ (solid) and $K_2$ (dashed), using $(N,M)=(32,12)$ and
$B=10$. FPR labels refer to per-key calibration on unwatermarked batches,
not family-wise or cross-key error rates. $T_{K_1}$ and $T_{K_2}$ denote
detection using keys $K_1$ and $K_2$, respectively.}
\label{fig:mixed_source_multikey}
\end{figure}

Figure~\ref{fig:mixed_source_multikey} shows detection under both keys.
At 5\% nominal clean FPR
per key, a batch containing ten $K_1$ images triggers only the $K_1$ detector
in 91.9\% of cases. As we progressively replace $K_1$ images with $K_2$ images,
detection under $K_1$ decreases gradually while detection under $K_2$ increases.
When the batch contains five images from each key, both detectors trigger in
36.6\% of cases. Because both keys select from the same candidates, batches containing only $K_2$ images are also detected under $K_1$ slightly above the nominal rate (2.2\%, 7.6\%, and 22.5\% at the three operating points). We therefore conclude that the detector provides a signal for
distinguishing known watermarking sources at the batch level.

\FloatBarrier

\clearpage
\raggedbottom
\subsection{Choice of Embedding Discretization}
\label{app:binarization_ablation}

Our method discretizes semantic embeddings with random-hyperplane LSH. We
compare LSH with two alternatives that also map each embedding to one of $2^M$
codes. A good discretization should spread independent generations of the same
prompt across many codes, because Lemma~\ref{lem:gumbel-race} ties the expected
keyed score of the selected code to the multiplicities $n_c$ of the occupied
codes: with ideal keyed scores, the expected gap of the selected score above
$\tfrac12$ is $\Delta=\sum_c n_c/(N+n_c)-\tfrac12$, which grows as the
candidates spread over more codes. A good discretization should also keep an
image's code unchanged under content-preserving transformations.

\emph{K-means} assigns each embedding to the nearest of $2^M$ centroids, which
we fit on 1,000 real COCO-2017 validation images. \emph{Coordinate sign} takes
the sign of the first $M$ embedding coordinates; it equals LSH with the random
hyperplanes replaced by the standard basis. We average LSH over five hyperplane
frames and K-means over five clustering seeds.

Only LSH guarantees that two embeddings at angle $\theta$ share a bit with
probability $1-\theta/\pi$, regardless of where they lie. This relation links
semantic entropy to code occupancy (Lemma~\ref{lem:expected-occupied-codes})
and encoder stability to score preservation
(Lemma~\ref{lem:semantic-score-preservation}). K-means cells depend on the
calibration data, and coordinate-sign collisions depend on the position of the
embeddings relative to the coordinate axes.

We measure code occupancy on Stable Diffusion~2.1 generations for the 1,000
held-out COCO-2017 prompts, with up to $N=320$ candidates per prompt, and
compute $\Delta$ from the code multiplicities of each batch. We measure code stability on the images and transformations of the
robustness evaluation (Appendix~\ref{app:robustness_tables}) as the percentage
of images whose full $M$-bit code changes.

\begin{figure}[H]
\centering
\includegraphics[width=\linewidth]{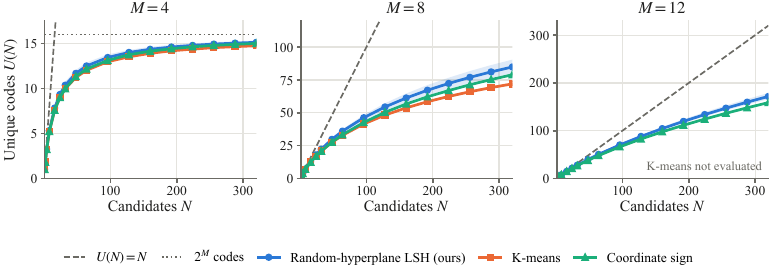}
\caption{\textbf{Code occupancy as the number of candidates grows.} Mean number
of distinct codes among $N$ candidates for the same prompt. Shading spans the
five hyperplane frames (LSH) and five clustering seeds (K-means). Higher is
better.}
\label{fig:binarization_unique_codes}
\end{figure}

\begin{figure}[H]
\centering
\includegraphics[width=\linewidth]{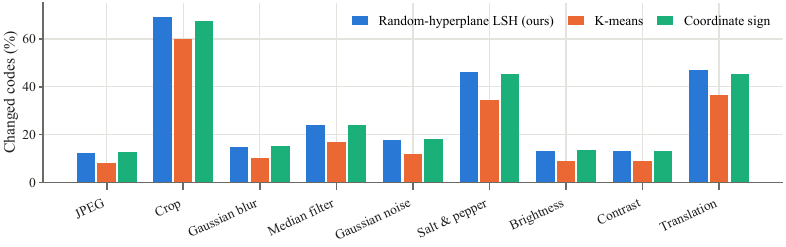}
\caption{\textbf{Code changes under transformations at $M=8$.} Percentage of
images whose full code changes, averaged over the settings in each family.
Lower is better.}
\label{fig:binarization_hash_changes}
\end{figure}

LSH occupies the most codes in every configuration
(Table~\ref{tab:binarization_diversity},
Figure~\ref{fig:binarization_unique_codes}). The margin is small at the paper's
candidate counts: for $(N,M)=(16,8)$, the expected gap is $0.414$ for LSH,
$0.408$ for K-means, and $0.407$ for coordinate sign. The margin grows with
$N$: at $M=8$ and $N=320$, LSH occupies $84.9$ codes, compared with $79.0$ for
coordinate sign and $72.0$ for K-means. In contrast, K-means changes fewer codes
under every transformation family (Figure~\ref{fig:binarization_hash_changes}),
for example $8.1\%$ versus $12.3\%$ for LSH under JPEG compression at $M=8$,
while LSH and coordinate sign behave alike.

No discretization is best on both criteria. We use LSH because it yields the
largest expected keyed-score gap and because it is the only one whose collision
probability depends solely on the angle between embeddings, which our analysis
requires.
Randomization may also make adaptive attacks harder. If we sample the
hyperplanes privately together with the key, public data reveals nothing about
the code boundaries. In contrast, the coordinate axes are fixed, and K-means
centroids depend mostly on the data distribution, so an attacker could
approximate them by clustering similar images. We leave evaluating this to
future work.

\begin{table}[H]
\centering
\small
\caption{Code occupancy at the paper configurations. $U$ is the mean number of
occupied codes among the $N$ candidates of a batch, and $\Delta$ is the expected
keyed-score gap of Gumbel selection (Lemma~\ref{lem:gumbel-race}). We evaluate K-means for
$M\in\{4,8\}$.}
\label{tab:binarization_diversity}
\begin{tabular}{@{}lcccccc@{}}
\toprule
& \multicolumn{2}{c}{LSH (ours)} & \multicolumn{2}{c}{K-means} & \multicolumn{2}{c}{Coordinate sign} \\
\cmidrule(lr){2-3}\cmidrule(lr){4-5}\cmidrule(lr){6-7}
$(N,M)$ & $U$ & $\Delta$ & $U$ & $\Delta$ & $U$ & $\Delta$ \\
\midrule
$(8,4)$   & \textbf{5.29}  & \textbf{0.309} & 5.17  & 0.304 & 5.18  & 0.304 \\
$(16,8)$  & \textbf{12.94} & \textbf{0.414} & 12.41 & 0.408 & 12.36 & 0.407 \\
$(32,12)$ & \textbf{27.71} & \textbf{0.459} & ---   & ---   & 26.88 & 0.456 \\
\bottomrule
\end{tabular}
\end{table}

\clearpage
\flushbottom

\clearpage
\subsection{Detection after Content-Changing Edits}
\label{app:content_change_detection}

Our encoder is trained to preserve detection under edits that preserve the
main content. We evaluate what happens when an edit replaces the main subject,
scene, or action by comparing edited watermarked images with edited
unwatermarked images.

We use 180 COCO-2017 prompts with Stable Diffusion~2.1, with 60 prompts per
edit category. Qwen3-VL-8B-Instruct~\citep{bai2025qwen3vl} writes one editing
instruction per prompt from the prompt text. For each prompt,
Qwen-Image-Edit~\citep{wu2025qwenimage} applies the same instruction and edit
seed to the Gumbel-selected images and to an independently generated clean
image. We include all renders without screening individual edits for success.
We use our single-image detector settings from
Table~\ref{tab:detection_settings}, with a fixed key and hyperplane frame.
Tree-Ring, Gaussian Shading, and SEAL receive the same instructions and edit
seeds and use their own clean-calibrated detectors.

\begin{figure}[H]
\centering
\includegraphics[width=\linewidth]{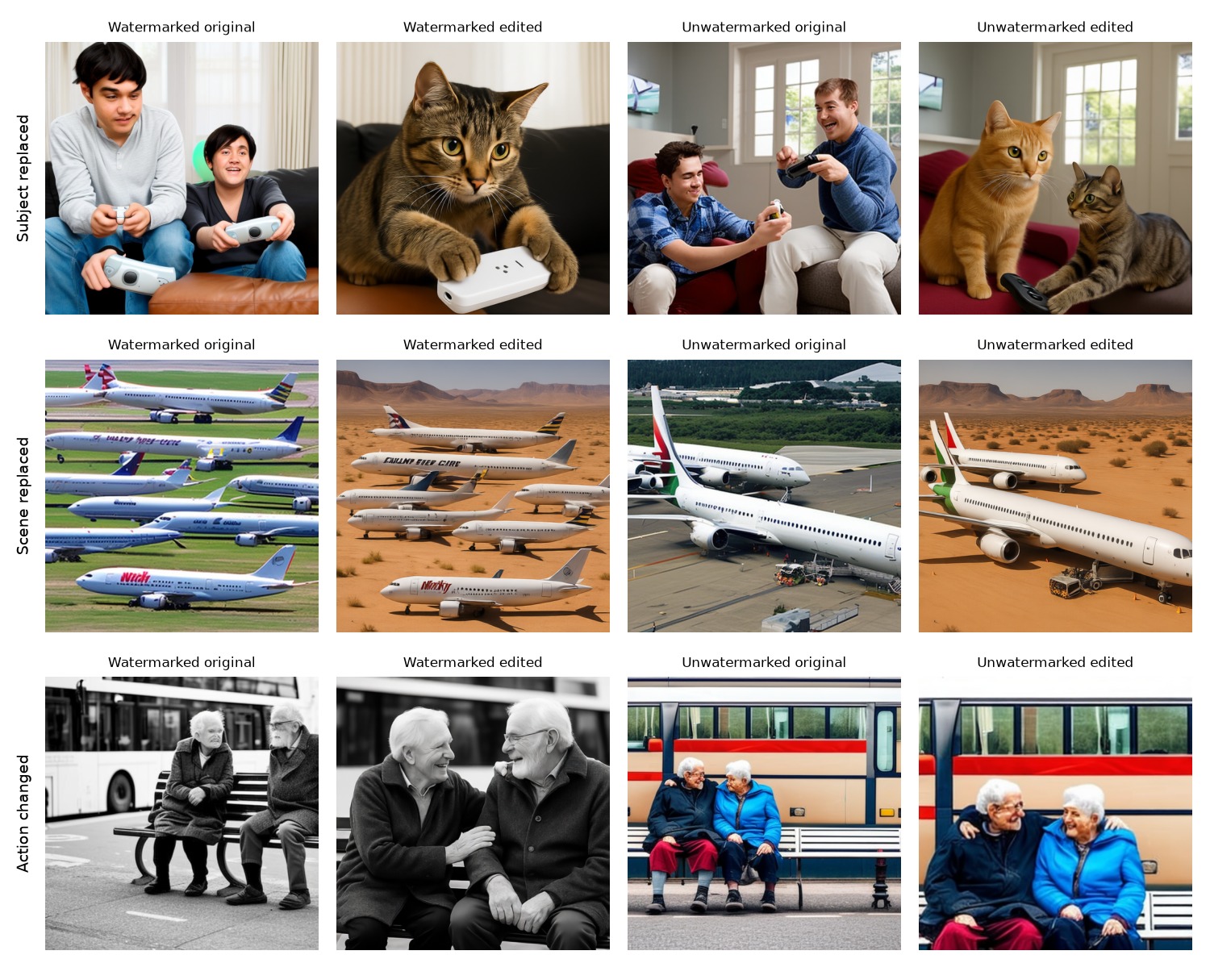}
\caption{\textbf{Content-changing edit examples.} The first prompt by
panel rank in each edit category for $(N,M)=(32,12)$. Each row shows
the same instruction and seed applied to the watermarked and clean
generations for that prompt.}
\label{fig:content_change_examples}
\end{figure}

Table~\ref{tab:content_change_detection} reports detection on both edited
populations. For $(N,M)=(32,12)$, edited TPR is $21.7$--$23.3\%$, compared with
$16.7$--$21.7\%$ on edited clean images. AUROC for this configuration is
$0.536$ for subject replacement, $0.463$ for scene replacement, and $0.541$
for action changes. Paired retention is $22.4$--$25.9\%$ across these edits.
For comparison, the separate content-preserving panel in
Figure~\ref{fig:robustness_per_edit_diagnostic} reports $94.7\%$ retention
after background cleanup for the same configuration.
Figure~\ref{fig:content_change_examples} shows an example from each category.

\begin{table}[H]
\centering
\small
\setlength{\tabcolsep}{4pt}
\caption{\textbf{Detection after content-changing edits.} All values are percentages at
$20\%$ nominal clean FPR. Each category has 60 prompts.
Paired retention is the fraction of originally detected watermarked images
that remain detected after editing. SEAL achieves $8.1\%$ on its
unedited calibration set.}
\label{tab:content_change_detection}
\begin{tabular}{@{}lrrrrrr@{}}
\toprule
Edit & \shortstack{Ours\\$(8,4)$} & \shortstack{Ours\\$(16,8)$} & \shortstack{Ours\\$(32,12)$} & Tree-Ring & \shortstack{Gaussian\\Shading} & SEAL \\
\midrule
\multicolumn{7}{@{}l}{\textit{Edited TPR}} \\
Subject replacement & 20.0 & 15.0 & 23.3 & 50.0 & 38.3 & 3.3 \\
Scene replacement & 25.0 & 23.3 & 21.7 & 40.0 & 25.0 & 11.7 \\
Action change & 38.3 & 28.3 & 23.3 & 31.7 & 25.0 & 10.0 \\
\addlinespace
\multicolumn{7}{@{}l}{\textit{Edited clean FPR}} \\
Subject replacement & 16.7 & 28.3 & 16.7 & 31.7 & 23.3 & 8.3 \\
Scene replacement & 15.0 & 16.7 & 16.7 & 28.3 & 18.3 & 13.3 \\
Action change & 25.0 & 15.0 & 21.7 & 13.3 & 15.0 & 6.7 \\
\addlinespace
\multicolumn{7}{@{}l}{\textit{Paired retention}} \\
Subject replacement & 31.8 & 14.6 & 25.9 & 50.0 & 38.3 & 3.3 \\
Scene replacement & 42.3 & 26.5 & 22.4 & 40.0 & 25.0 & 11.7 \\
Action change & 54.8 & 37.0 & 25.0 & 31.7 & 25.0 & 10.0 \\
\bottomrule
\end{tabular}
\end{table}

\clearpage
\FloatBarrier

\section{Extended Limitations and Future Directions}
\label{app:extended_limitations}

This appendix expands on the limitations summarized in the main paper. We
discuss single-image detection under limited black-box sampling, computational
cost, fixed-key reuse, adaptive attacks, and extension beyond images.

\subsection{Single-Image Detection under Limited Black-Box Sampling}
\label{app:single-image-sampling-limit}

Our image implementation has weaker clean single-image detection than methods
that insert a signal directly into the diffusion process
(Table~\ref{tab:experiment3_detection}). Aggregating scores across independently
generated outputs substantially improves detection, in agreement with
Lemma~\ref{lem:batch-detection-concentration}. However, aggregation is
unavailable when each image must be judged independently.

The following theorem shows that any black-box sampling method selecting one
of \(N\) independent, unmodified generator outputs can amplify the single-image
detection probability over the false-positive rate by at most a factor of
\(N\). This \(O(N)\) ceiling belongs to black-box sampling, not our construction.

\begin{resulttheorem}[Single-image detection under limited sampling]
\label{thm:single-image-sampling-limit}
Fix a prompt \(p\), secret key \(\kappa\), and detector
\(D_\kappa:\mathcal X\to\{0,1\}\). Let
\(X_1,\ldots,X_N\overset{\mathrm{i.i.d.}}{\sim}P_G(\cdot\mid p)\).
Consider any possibly randomized watermarking algorithm that returns
\(X_{I^\star}\), where \(I^\star\in[N]\) may depend on the candidates and key.
For a clean image \(X\sim P_G(\cdot\mid p)\), define
\[
\alpha=\operatorname{FPR}_{\kappa,p}
:=\Pr[D_\kappa(X)=1],
\qquad
\operatorname{TPR}_{\kappa,p}
:=\Pr[D_\kappa(X_{I^\star})=1].
\]
Then
\[
\operatorname{TPR}_{\kappa,p}
\leq N\alpha
=N\operatorname{FPR}_{\kappa,p}.
\]
\end{resulttheorem}

\begin{proof}
The algorithm can return a detected image only if at least one candidate is
detected. Therefore,
\[
\{D_\kappa(X_{I^\star})=1\}
\subseteq\bigcup_{i=1}^N\{D_\kappa(X_i)=1\}.
\]
\[
\operatorname{TPR}_{\kappa,p}
\leq\Pr\!\left[\bigcup_{i=1}^N\{D_\kappa(X_i)=1\}\right]
\leq\sum_{i=1}^N\Pr[D_\kappa(X_i)=1]=N\alpha.
\]
\end{proof}

This bound applies to any method selecting among \(N\) independent, unmodified
generator outputs, regardless of distortion-freeness, encoder, or scoring rule.
For \(\alpha>0\),
achieving \(\operatorname{TPR}_{\kappa,p}\geq c>0\) requires
\(N\geq c/\alpha\). Thus, targeting \(\alpha\leq e^{-t}\) requires
\(N\geq ce^t\): exponentially small FPR at constant TPR requires
exponentially many generator queries.

This does not contradict the exponential improvement from aggregating
independently generated images in
Lemma~\ref{lem:batch-detection-concentration}: the query limit concerns one
sampled output, whereas batch detection accumulates evidence across multiple
outputs.

\subsection{Computational Cost and Early Candidate Selection}

Our strongest configuration generates 32 complete candidates for every
released image. This cost may be prohibitive in large-scale or latency-sensitive
applications.

Recent work suggests that earlier candidate selection may be feasible for image
diffusion models. \citet{guo2026earlyquality} show that intermediate denoiser
activations can predict final-image quality and allow unpromising trajectories
to be terminated early. \citet{habibian2024clockwork} show that low-resolution
diffusion features carry semantic layout and can be reused across denoising
steps. These results do not establish that our final semantic code can be
predicted before generation is complete, but they show that relevant structural
and semantic information is available at intermediate stages.

A promising direction is to predict the final semantic code or keyed score from
intermediate features, discard unlikely candidates, and complete only promising
trajectories. Any such optimization must preserve the selection probabilities
required by Theorem~\ref{thm:distortion-free}. Prediction errors may discard
trajectories that would have received high final scores, weakening detection
even when image quality is unchanged. A useful evaluation must therefore report
expected completed denoising steps and wall-clock latency together with
detection, image quality, and distribution shift. This would expose the full
computation--detection tradeoff rather than treating early termination as a
speedup in isolation.

\subsection{Distribution Preservation under Fixed-Key Reuse}

Theorem~\ref{thm:distortion-free} establishes equality with the original
generator distribution after averaging over the secret key and the sampler's
internal randomness. It does not establish distribution preservation separately
for every fixed key. A fixed key deliberately favors particular semantic
codes. Theorem~\ref{thm:semantic-entropy-tpr} characterizes the resulting
detection probability only after taking expectations over the key and random
hyperplanes; it does not imply the same guarantee for every fixed deployed key.

The matched direct-FID experiment shows that our method produces the smallest
measured distribution shift among the evaluated methods, while the
same-prompt diversity experiment shows that it stays within \(1.3\%\) of the
diversity of matched clean generations
(Table~\ref{tab:experiment3_quality_diversity}).
These experiments provide evidence that fixed-key reuse preserves quality and
diversity in the evaluated setting, but they do not constitute a general
per-key guarantee. Establishing fixed-key bounds in terms of the number of
candidates, code occupancy, and encoder geometry remains an important
theoretical direction.

\subsection{Adaptive Removal and Forgery Attacks}

Our robustness evaluation considers non-adaptive appearance, compression,
geometric, and generative transformations. Figure~\ref{fig:experiment3_robustness_categories}
summarizes the resulting detection and retention rates, while
Appendix~\ref{app:robustness_protocols} reports the individual settings, detector
configurations, and achieved false-positive rates.

These experiments do not establish security against an attacker who
deliberately targets the encoder or detector. Because our score is bound to the
semantic representation of the final image rather than to diffusion-specific
latent noise, the method may be less susceptible to reprompting attacks that
transfer a watermark to semantically different content, as studied by
\citet{muller2025forgery} and addressed by SemBind~\citep{zhang2026sembind}.
This possibility has not been evaluated directly. An adaptive attacker may
instead attempt to cross encoder or hash boundaries while preserving apparent
content, remove a high-scoring code, or produce altered content that retains
the original score. Lemma~\ref{lem:semantic-score-preservation} characterizes
non-adversarial behavior under representation similarity and should not be
interpreted as a security guarantee against such attacks.

\subsection{Extension beyond Images}

The framework treats the generator as a sampling oracle and is not tied to a
particular architecture or output modality.

Each new domain requires a semantic encoder that separates independent outputs
while remaining stable under the relevant content-preserving transformations.
Future work should determine whether such encoders provide sufficient code
occupancy for detection, whether their geometry supports robustness to
realistic transformations, and whether sampling enough candidates remains
computationally practical.

\end{document}